%% file: main.tex
\documentclass[journal]{IEEEtran}
\usepackage{xcolor}
\usepackage{array}
\usepackage{colortbl}
\usepackage{booktabs}
\usepackage{multirow}
\usepackage{fancyhdr}
\usepackage{amsmath}
\usepackage{customcode}
\usepackage{caption}

\usepackage{tocloft}
\usepackage{etoc}
\usepackage{titletoc}
\usepackage{tcolorbox}
\usepackage{graphicx}
\usepackage{pifont}
\usepackage[T1]{fontenc}
\usepackage[utf8]{inputenc}
\usepackage{csquotes}
\usepackage{url}
\usepackage{hyperref}
\usepackage{enumitem}

\newcommand\DoToC{%
  \startcontents
  \printcontents{}{1}{\noindent \textbf{\Large{Table of Contents}}\vskip3pt\vskip5pt}
  \vskip3pt\vskip5pt
}

\usepackage{tocloft}

\begin{document}

\title{\raisebox{-0.35cm}{\includegraphics[width=1.2cm]{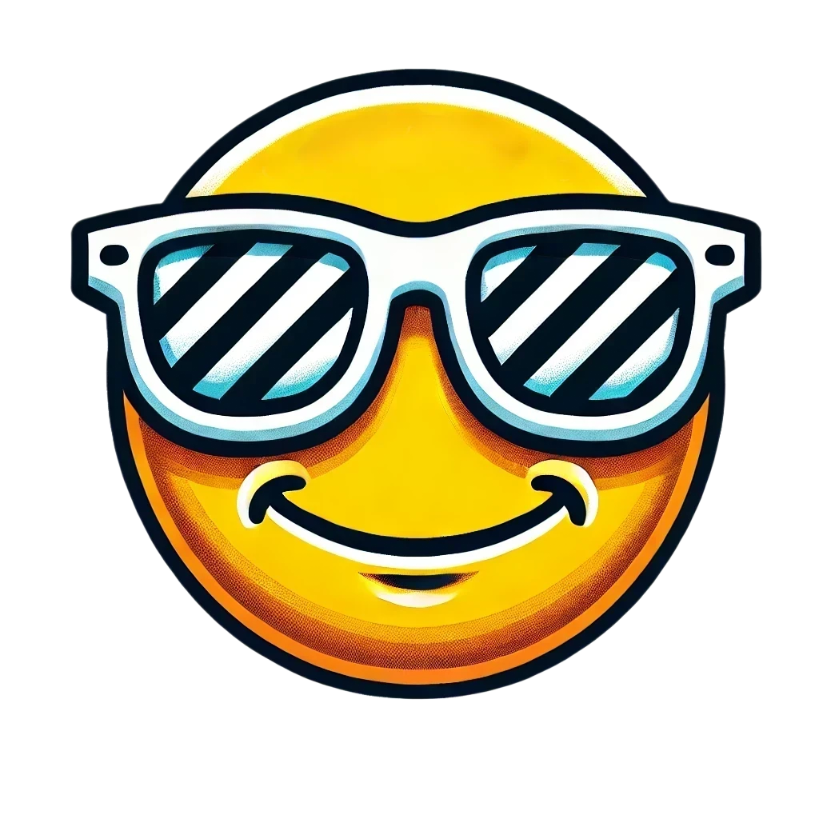}} FaceXBench: Evaluating Multimodal LLMs \\ on Face Understanding}

\author{Kartik Narayan,
        Vibashan VS,
        and~Vishal M. Patel.
\thanks{K. Narayan is with the Department
of Computer Science, Johns Hopkins University, Baltimore,
MD, 21218 USA. e-mail: knaraya4@jhu.edu.}
\thanks{VS Vibashan and V M. Patel. are with the Department of Electrical and Computer Engineering, Johns Hopkins University, Baltimore,
MD, 21218 USA. emal: \{vvishnu2, vpatel36\}@jhu.edu}
\\
\textcolor{magenta}{\url{https://kartik-3004.github.io/facexformer/}
}
}

\markboth{Accepted to IEEE T-BIOM}%
{\MakeLowercase{\textit{Narayan et al.}}: FaceXBench}

\maketitle
\input{sec/0_abstract}


\IEEEpeerreviewmaketitle

\input{sec/1_intro}
\input{sec/2_related}

\input{sec/3_proposed}

\input{sec/4_experiments}

\input{sec/5_results}

\input{sec/6_discussion}
\input{sec/8_conclusion}

\bibliographystyle{IEEEtran}
\bibliography{references}

\counterwithin{figure}{section}
\counterwithin{table}{section}

\fancypagestyle{appendixfooter}{
  \fancyhf{} 
  \renewcommand{\headrulewidth}{0pt}
  \renewcommand{\footrulewidth}{0pt}
  \fancyfoot[L]{\hyperlink{appendix-start}{\textbf{\textit{Go to Supplementary Index}}}}
  \fancyfoot[C]{\thepage} 
}

\onecolumn
\makebox[\textwidth][c]{%
    \parbox{ \textwidth }{%
        \centering
        {\Large \textbf{\raisebox{-0.32cm}{\includegraphics[width=1cm]{uploads/logo.png}} FaceXBench: Evaluating Multimodal LLMs on Face Understanding}}\\
        {\Large Supplementary Material}
    }
}
\vspace{10pt}

\hypertarget{appendix-start}{}
\pagestyle{appendixfooter}

In the supplementary material, we discuss additional questions that frame the motivation for our work and provide a discussion of its limitations. Along with this, we present detailed information regarding the benchmark, including the broad categories, the tasks, the source datasets used, and the dataset statistics.
Additionally, we focus on its implementation and provide extensive details about the prompts used for dataset collection and the evaluation strategy. Furthermore, we expand on the results presented in the main paper, providing more details about the baselines, showcasing failure cases of the models, and concluding with an ethics statement.
\\

\DoToC
\clearpage
\onecolumn
\input{appendix/X_motivation}
\input{appendix/X_limitations}
\input{appendix/X_facexbench}
\input{appendix/X_dataset}

\input{appendix/X_results}
\input{appendix/X_ethics}

\ifCLASSOPTIONcaptionsoff
  \newpage
\fi

\end{document}

%% file: sec/0_abstract.tex
\begin{abstract}
\input{figures/visual_abstract}
Multimodal Large Language Models (MLLMs) demonstrate impressive problem-solving abilities across a wide range of tasks and domains. However, their capacity for face understanding has not been systematically studied. To address this gap, we introduce FaceXBench, a comprehensive benchmark designed to evaluate MLLMs on complex face understanding tasks. FaceXBench includes $5,000$ multimodal multiple-choice questions derived from $25$ public datasets and a newly created dataset, FaceXAPI. These questions cover $14$ tasks across $6$ broad categories, assessing MLLMs' face understanding abilities in bias and fairness, face authentication, recognition, analysis, localization and tool retrieval.
Using FaceXBench, we conduct an extensive evaluation of $26$ open-source MLLMs alongside $2$ proprietary models, revealing the unique challenges in complex face understanding tasks. We analyze the models across three evaluation settings: zero-shot, in-context task description, and chain-of-thought prompting. Our detailed analysis reveals that current MLLMs, including advanced models like GPT-4o, and GeminiPro 1.5, show significant room for improvement. We believe FaceXBench will be a crucial resource for developing MLLMs equipped to perform sophisticated face understanding.
\end{abstract}

%% file: figures/visual_abstract.tex
\begin{figure*}[!ht]
    \centering
    \includegraphics[width=\linewidth]{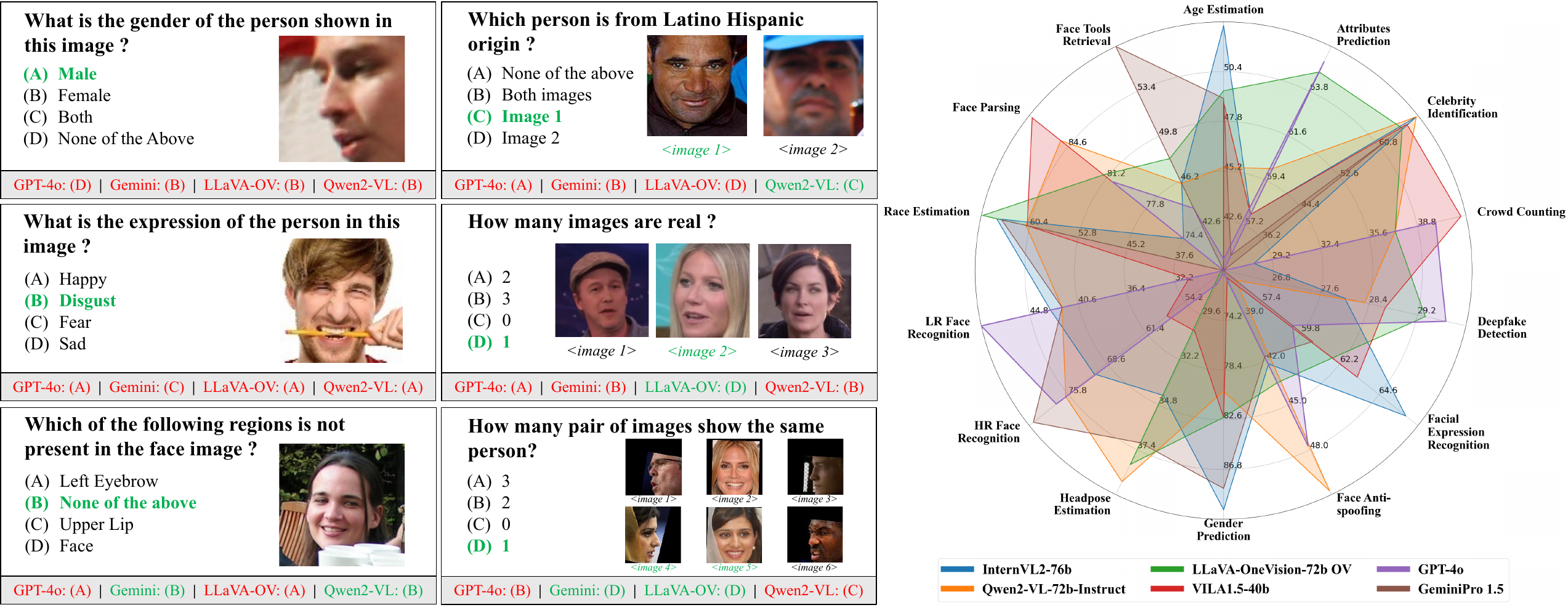}
    \caption{\textbf{Left:} Failure cases of leading MLLMs, such as LLaVA-OV~\cite{li2024llava}, Qwen2-VL~\cite{Qwen2VL}, GPT-4o~\cite{hurst2024gpt} and GeminiPro1.5~\cite{team5gemini}, on \textit{basic} questions related to face understanding. \textbf{Right:} Performance comparison of top models across the $14$ tasks included in the benchmark.}
    \label{fig:visual_abstract}
\end{figure*}

%% file: sec/1_intro.tex
\section{Introduction}
\label{introduction}
Recent advances in large language models (LLMs)\cite{achiam2023gpt, anthropic2024claude, touvron2023llama} have demonstrated strong capabilities in understanding, reasoning, and generating text across diverse open-ended tasks. Building on this progress, multimodal LLMs (MLLMs)\cite{li2023blip, zhu2023minigpt, liu2024visual, ye2023mplug, li2023videochat, maaz2023video} have emerged, leveraging LLMs to process multimodal inputs such as images, videos, and audio. These models~\cite{li2024llava, chen2024far, abdin2024phi, lin2024vila, hong2024cogvlm2} excel in visual-question-answering (VQA) benchmarks like MMMU~\cite{yue2024mmmu}, MME~\cite{yin2023survey}, MM-Bench~\cite{liu2025mmbench}, SEED-Bench~\cite{li2024seed}, MMAU~\cite{yin2024mmau}, MMIE~\cite{xia2024mmie}, and MathVista~\cite{lu2024mathvista}, which test a range of skills including subject knowledge, mathematical and commonsense reasoning, planning, coding, and spatio-temporal understanding. Despite this progress, leading MLLMs such as GPT-4o~\cite{hurst2024gpt} and GeminiPro 1.5~\cite{team5gemini} still struggle with basic face understanding tasks (see Figure~\ref{fig:visual_abstract}).

Before exploring MLLMs for face understanding capabilities, it’s essential to address the fundamental question: \textquotedblleft \textit{\textbf{Why should MLLMs be proficient in face understanding?}}\textquotedblright~MLLMs are increasingly deployed as central processors in various advanced applications, including virtual-reality headsets~\cite{konenkov2024vr}, embodied AI~\cite{huang2023modality, driess2023palm}, driving safety~\cite{hwang2024emma, sreeram2024probing}, authentication~\cite{deandres2024good}, human-computer interaction~\cite{wang2024large}, and sports analysis~\cite{xia2024sportu}. In these applications, accurate face understanding is crucial, as face images appear frequently and require accurate face understanding for appropriate responses. However, the face understanding capabilities of existing MLLMs are limited; they often fail to answer basic questions such as \textquotedblleft \textit{What is the expression of the person in this image?}\textquotedblright~or \textquotedblleft \textit{What is the age of the person in the face image?}\textquotedblright~These shortcomings indicate a significant scope for improvement. Recent works, such as EMO-LLaMA~\cite{xing2024emo}, develop an instruction-tuning set for supervised fine-tuning to enhance MLLMs' capabilities in understanding facial expressions. Face-MLLM~\cite{sun2024face} proposed a three-stage training pipeline to equip MLLMs' with face perception capabilities. Nonetheless, the research community currently lacks a standardized benchmark to quantify and compare MLLMs' performance in face understanding. We believe that a comprehensive benchmark incorporating various aspects of face understanding is a \textit{crucial and essential foundational step} for monitoring progress and advancing MLLMs' performance in this domain. A more detailed discussion, addressing additional questions that frame our motivation, are provided in the Appendix.

To this end, we propose \textbf{FaceXBench}, a comprehensive benchmark designed for complex face understanding and related tasks. We identify six broad categories, each encompassing distinct tasks essential for complete face understanding within the context of MLLMs:

\begin{enumerate}
\item \textbf{Bias and Fairness}: \textit{Age Estimation} (Age), \textit{Gender Prediction} (Gender), and \textit{Race Estimation} (Race);
\item \textbf{Face Recognition}: \textit{High-Resolution Face Recognition} (HR-FR), \textit{Low-Resolution Face Recognition} (LR-FR), and \textit{Celebrity Identification} (Celeb);
\item \textbf{Face Authentication}: \textit{Face Anti-Spoofing} (FAS) and \textit{Deepfake Detection} (Deepfake);
\item \textbf{Face Analysis}: \textit{Attributes Prediction} (Attr) and \textit{Facial Expression Recognition} (FER);
\item \textbf{Face Localization}: \textit{Head Pose Estimation} (HPE), \textit{Face Parsing} (Seg), and \textit{Crowd Counting} (Crowd);
\item \textbf{Face Tools Use}: \textit{Face Tools Retrieval} (Tools).
\end{enumerate}

Recent works~\cite{qin10toolllm, tang2023toolalpaca} emphasize that, although existing open-source MLLMs' have achieved advanced capabilities, they still lack the sophistication to perform complex tasks. To address this, these approaches enable MLLMs with tool use, extending their capabilities by leveraging external tools to fulfill complex human instructions that require task-specific processing. Motivated by this, we introduce FaceXAPI, a new dataset that forms part of FaceXBench. It is designed to evaluate MLLMs' ability to select the appropriate API and function calls for handling complex tasks in face understanding. FaceXBench comprises $5,000$ carefully and manually filtered VQA-type questions derived from $25$ public datasets, covering $14$ tasks. Following established benchmarks such as MMBench~\cite{liu2024mmbench} and SEEDBench~\cite{li2023seed}, we curate multiple-choice questions derived from off-the-shelf datasets to maintain consistency and standardization, ensuring ease of use. In contrast, open-ended facial VQA relies on external models for answer matching, making the evaluation non-deterministic. Additionally, FaceXBench includes $10,441$ unique face images representing a diverse range of age groups, genders, races, varied resolutions, head poses and expressions, reflecting the diversity of faces encountered in real-world scenarios. 

We conduct extensive experiments and benchmark $26$ open-source MLLMs' and $2$ advanced proprietary MLLMs, GPT-4o~\cite{hurst2024gpt} and GeminiPro-1.5~\cite{team5gemini}. We analyze the models across three evaluation settings: (1) \textit{zero-shot}, (2) \textit{in-context task description} and (3) \textit{chain-of-thought (CoT)} prompting. Our findings reveals \underline
{two main takeaways}: First, existing MLLMs struggle with tasks like deepfake detection and crowd counting, which require fine-grained visual analysis to detect subtle inconsistencies and the ability to recognize and adapt to faces at varying scales. The performance of top models across various tasks is illustrated in Figure~\ref{fig:visual_abstract}. Second, attempts to leverage the face knowledge of MLLMs through in-context prompting lead to performance drops, indicating a struggle to utilize contextual information effectively. Chain-of-thought prompting similarly fails to improve performance, suggesting that while these models have reasoning capabilities, they cannot apply them to face-related tasks, limiting their ability to make nuanced interpretations or adjustments. Finally, we present experiments highlighting the potential of diverse supervised fine-tuning and tool use as promising directions for enhancing MLLMs' face understanding capabilities. 

\noindent Our key contributions are as follows:
\begin{itemize}
    \item \textbf{Introducing FaceXBench:} A comprehensive benchmark for evaluating MLLMs' face understanding across $14$ tasks in $6$ key categories. It includes $5,000$ VQA questions derived from $25$ public datasets and a newly developed dataset, FaceXAPI.
    \item \textbf{Extensive Evaluation:} We evaluate $26$ open-source MLLMs and two proprietary models, GPT-4o and GeminiPro1.5, which achieve accuracies of $50.24$\% and $54.40$\%, respectively, highlighting the significant challenge posed by FaceXBench and the substantial room for improvement. 
    \item \textbf{Analysis and Discussion:} We provide a detailed analysis of MLLMs performance across various aspects of face understanding, identifying areas where current MLLMs' fall short. Additionally, we suggest potential research directions that could enhance MLLMs' face understanding.
\end{itemize}

%% file: sec/2_related.tex
\section{Related Work}
\noindent \textbf{Face Understanding:}
Face understanding concerns the analysis and interpretation of human facial features, encompassing tasks such as age estimation~\cite{sun2021deep, kjaerran2021facial, levi2015age, cao2020rank}, gender prediction~\cite{kumar2019gender, al_dujaili2023gender, lin2020face}, race estimation~\cite{ahmed2022race, ahmed2024hybrid}, face recognition~\cite{narayan2024petalface, wang2018cosface, deng2019arcface}, face anti-spoofing~\cite{yang2019face, liu2023towards, narayan2024hyp}, deepfake detection~\cite{zhao2021multi, thakral2024deephynet, narayan2023df, narayan2022deephy}, facial attributes prediction~\cite{miyato2018virtual, zhuang2018multi, zhang2024distributionally}, facial expression recognition~\cite{wang2023rethinking, Lee_2023_CVPR, zeng2022face2exp}, headpose estimation~\cite{zhang2023tokenhpe, hempel20226d, ruiz2018fine}, crowd counting~\cite{zhang2015cross, ranjan2018iterative, ranasinghe2023diffuse} and face parsing~\cite{zheng2022decoupled, sarkar2023parameter, narayan2024segface, te2021agrnet}. Early methods focused on developing task-specific models for each facial understanding task, achieving promising results. With the advent of transformers, works such as FaceXFormer~\cite{narayan2024facexformer}, Q-Face~\cite{sun2024task} and Faceptor~\cite{qin2025faceptor} aim to unify multiple tasks within a single architecture to improve generalization performance. The recent rise in MLLMs' enabled new avenue of research in face understanding by integrating information across text, visual and other modalities. Recent works~\cite{chettaoui2024froundation, xu2024fakeshield, zhao2024enhancing, shi2024shield, xing2024emo, sun2024face}, leverage the reasoning and zero-shot capabilities of MLLMs' to approach traditional face-related tasks. However, the field still lacks a standardized benchmark to monitor and regulate the development of these models. Our proposed work addresses this gap by introducing a comprehensive benchmark, FaceXBench, which covers a diverse range of face understanding tasks.  

\noindent \textbf{Multimodal Large Language Models and Benchmarks:}
Following the success of Large Language Models~\cite{achiam2023gpt, touvron2023llama, bai2023qwen}, recent research has focused on MLLMs to enhance multimodal comprehension and generation by leveraging the strong generalization capabilities of LLMs. With the rapid development of MLLMs~\cite{li2023blip, zhu2023minigpt, liu2024visual, ye2023mplug, jiang2024mantis, beyer2024paligemma, shi2024eagle, lin2024vila, tong2024cambrian, wang2023cogvlm, chen2024internvl, bai2023qwen}, several works propose benchmarks to MLLMs across different aspects. MMMU~\cite{yue2024mmmu} meticulously collects multimodal questions from college exams, quizzes, and textbooks to assess college-level subject knowledge. MathVista~\cite{lu2024mathvista} combines challenges from diverse mathematical and visual tasks, focusing on evaluating models' fine-grained, deep visual understanding and compositional reasoning. MMBench~\cite{liu2025mmbench} proposes a bilingual objective benchmark and a CircularEval strategy for models with limited instruction-following capabilities. SEED-Bench~\cite{li2023seed}, MMAU~\cite{sakshi2024mmau} and MMIE~\cite{xia2024mmie} introduce benchmarks for generative comprehension and generation, audio understanding and interleaved multimodal comprehension and generation, respectively. Some benchmarks, such as SWE-Bench~\cite{jimenez2023swe} and OSWorld~\cite{xie2024osworld} focus on agentic reasoning and multimodal agents for open-ended tasks. SHIELD~\cite{shi2024shield} is an early attempt to benchmark MLLMs on face anti-spoofing and face forgery detection but overlooks several other aspects of face understanding. Recent works, such as EMO-LLaMA~\cite{xing2024emo} and Face-MLLM~\cite{sun2024face}, aim to enhance MLLMs' capabilities in facial expression recognition and face perception; however the field lacks a comprehensive benchmark to objectively evaluate these tasks. In this work, we introduce FaceXBench, a comprehensive benchmark, which contains $5,000$ MCQ that evaluates $6$ broad categories and $14$ different tasks.

%% file: sec/3_proposed.tex
\section{The FaceXBench Benchmark}
\subsection{Overview of FaceXBench}
We introduce FaceXBench, a novel and comprehensive benchmark encompassing multiple aspects of face understanding. Additionally, we propose FaceXAPI, a manually curated dataset within FaceXBench that aims to evaluate tool-use capabilities. FaceXBench is the first benchmark specifically designed to evaluate MLLMs' performance on face-related tasks. It assess MLLMs' across six broad categories: bias and fairness, face authentication, face recognition, face analysis, face localization, and facial tool use. Figure~\ref{fig:pie} presents a detailed taxonomy of these categories, outlining their respective tasks and corresponding number of questions. The designed questions test a MLLMs' capabilities in areas such as visual grounding, fine-grained feature extraction, anomaly detection, emotion recognition, fairness, contextual understanding, spatial understanding and agentic reasoning. 
\input{figures/statistics}

\noindent \textbf{Benchmark Statistics.}
FaceXBench consists of $5,000$ questions, divided into $6$ categories covering $14$ tasks, and is derived from $25$ public datasets along with one proposed dataset (i.e. FaceXAPI). There are $2,900$ questions with multiple images, $2,000$ single-image questions, and $100$ text-only questions. The questions are generated using $573$ unique question templates and contain $10,735$ unique images. The correct answer options are approximately equally distributed between A, B, C and D to avoid bias. The key statistics of the benchmark are summarized in Table~\ref{tab:stats}.

\subsection{Data Collection}
Our data collection pipeline consists of three steps. \textbf{Step 1.} In the first step, we iterate through the identified tasks in each category and select the datasets corresponding to each task. We collect the test sets of existing standard datasets for each task to avoid \textit{data leakage}. We also created a new dataset, FaceXAPI, specifically for the tool-retrieval task. In total, our collection includes $25$ public datasets along with this newly proposed dataset. 
\textbf{Step 2.} In this step, we manually created question templates for each task involving single and multiple images. While creating these templates, we carefully framed questions to encourage the model to reason, compare, and think critically to find the correct answer. We included a mix of easy and hard questions to maintain a balanced distribution. For example, a hard question is, ``\textit{\textless image1\textgreater\textless image2\textgreater\textless image3\textgreater\ How many images show a person in the age range of $30$ to $39$?},'' while an easy question is, ``\textit{\textless image1\textgreater\ What is the age range of the person shown in the image?}''. We prompt GPT-4o with the manually curated question templates as in-context examples and generate additional templates that address similar questions but with varied phrasing, perspective, and/or complexity. For example, ``\textit{\textless image1\textgreater\textless image2\textgreater\ Which among these two appears to be older?}''. We manually filter the question templates for each task to ensure diversity and task relevance. Additional details about the question templates and their generation can be found in the supplementary material. 
\textbf{Step 3.} The final step involves generating answer options. Each question includes four options, with one correct answer. We ensure that distractor options are close to the correct answer but distinct enough to require critical thinking and reasoning when selecting the correct response. For example, for the question ``\textit{\textless image1\textgreater\ What is the age range of the person shown in the image?}'', the options designed are (A) $20$ to $29$, (B) $10$ to $19$ (C) $30$ to $39$ and (D) $40$ to $49$, with the correct answer being (A) $20$ to $29$. The thoughtfully chosen distractor options close to the actual answer, encourages the model to reason, compare and carefully select the correct choice. This approach raises the difficulty for the model, making the questions more challenging. To validate this, we conduct a mini experiment using LLaVA-OV for age estimation on FairFace dataset questions. The model achieved $88$\% accuracy with randomly chosen options, compared to $53$\% with strategically designed options. This result highlights the importance of identifying effective distractor options. Further details on generating distractor options for each task are provided in the supplementary material. Overall, we observed that the model's performance significantly varies with the complexity of the questions and distractor options.
\input{figures/samples}
\input{figures/piechart}

\subsection{FaceXAPI} 
We believe that face understanding is an application domain that can be better addressed by equipping MLLMs with tools rather than relying solely on supervised fine-tuning. We provide a detailed discussion, supported by experimental validation, in the Discussion and Future Directions section (Section~\ref{sec:discussion}). To test MLLMs' ability to select the correct sequence of APIs and function calls for successful task completion, we created FaceXAPI, a dataset of $100$ text-only questions referencing 13 APIs and 32 function calls. The questions were designed to ensure diversity in scenarios, reflecting a broad range of real-world applications and requiring a sequence of $3$ to $5$ API calls to solve. The dataset includes a total of 88 unique function call sequences. The questions were generated using GPT-4o, followed by manual filtering. A sample from the dataset is presented in Figure~\ref{fig:samples}, with the complete prompt provided in the supplementary material.

\subsection{Quality Control}
We implement quality control checks at each step of the data collection process to prevent the error propagation. In the first stage of data collection, we convert established, manually annotated datasets for each task into VQA format. This approach ensures a high level of initial accuracy, compared to benchmarks with AI generated ground truths. In the second stage, we focus on achieving diversity and variety in question generation while maintaining appropriate difficulty levels, which ensures that the questions generated are of mixed difficulty and relevant to the task. We manually verify the correctness and relevance of the $757$ unique question templates in the benchmark. Finally, we systematically design options that are of high quality and are difficult.

%% file: figures/statistics.tex
\begin{table}[t!]
    \centering
    \fontsize{9pt}{\baselineskip}\selectfont
    \renewcommand\tabcolsep{3.0pt}
    \renewcommand\arraystretch{1.2}
    \caption{Key statistics of questions in FaceXBench.}
    \begin{tabular}{l@{\hspace{20pt}}l@{\hspace{2pt}}}
    \hline
    \textbf{Statistic} & \textbf{Number} \\
    \hline
    Total questions & 5000 \\
    Total categories & 6 \\
    Total tasks & 14 \\
    Public datasets used & 25 \\
    New dataset proposed (FaceXAPI) & 1 \\
    \hline
    Questions with multiple images & 2900 (58\%) \\
    Questions with single image & 2000 (40\%) \\
    Questions with only text & 100 (2\%) \\
    \hline
    Total images in all questions & 11694 \\
    Unique number of images & 10735 \\
    Unique question templates & 573 \\
    \hline
    Maximum question length & 676 \\
    Maximum option length & 207 \\
    Average question length & 67.13 \\
    Average option length & 11.55 \\
    \hline
    Total options in each question & 4 \\
    Frequency of A as correct option & 1256 (25.12\%) \\
    Frequency of B as correct option & 1266 (25.32\%) \\
    Frequency of C as correct option & 1246 (24.92\%) \\
    Frequency of D as correct option & 1232 (24.64\%) \\
    \hline
    \end{tabular}
    \label{tab:stats}
\end{table}

%% file: figures/samples.tex
\begin{figure*}[!t]
    \centering
    \includegraphics[width=0.97\linewidth]{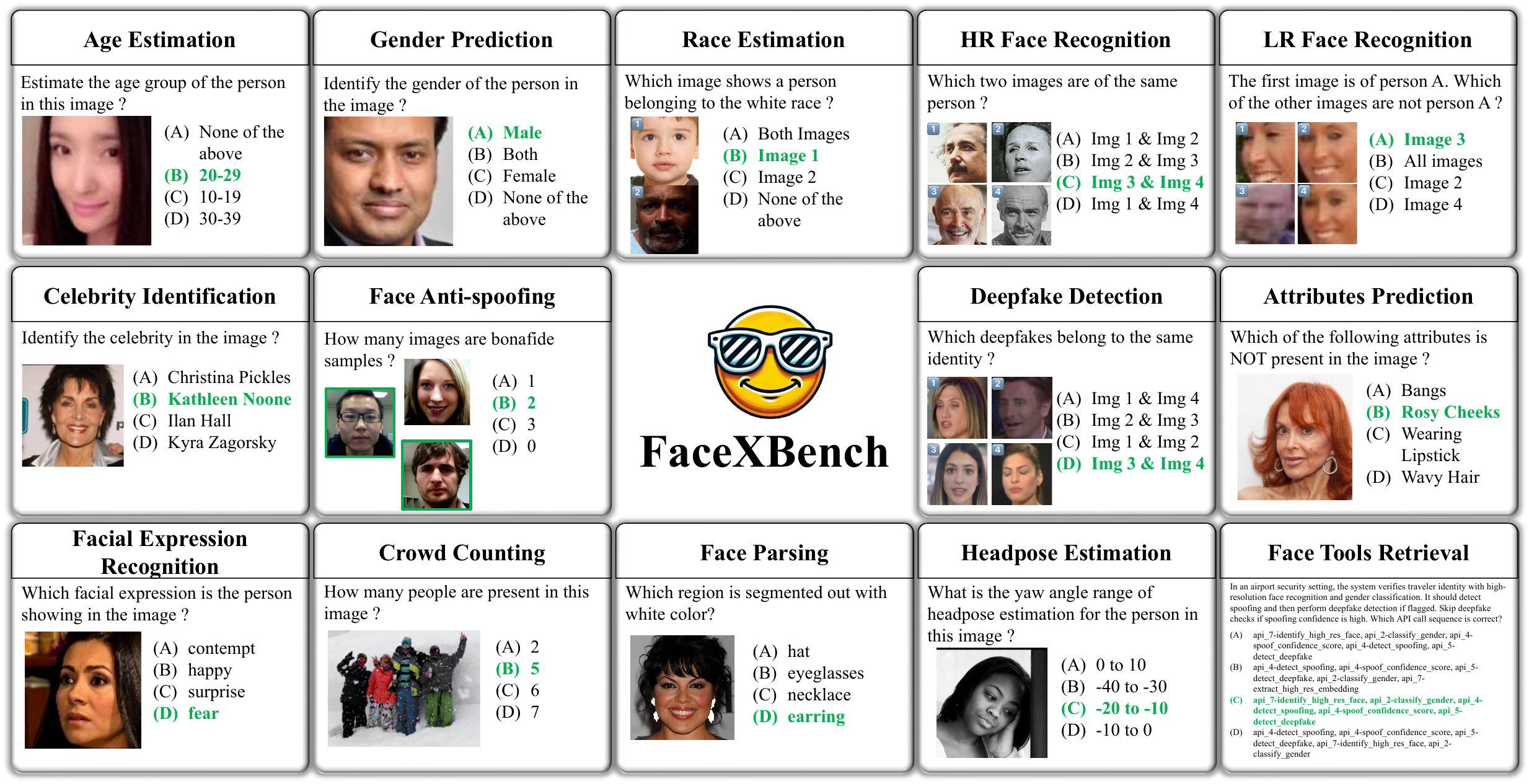}
    \caption{\textbf{FaceXBench examples} cover a total of 14 tasks, addressing various aspects of face understanding. Each question may consist of single or multiple images. Every question includes four options, with only one correct answer. The options are strategically designed to prompt the model to analyze carefully before selecting an option.}
    \label{fig:samples}
\end{figure*}

%% file: figures/piechart.tex
\begin{figure}[t!]
    \centering
    \includegraphics[width=0.8\linewidth]{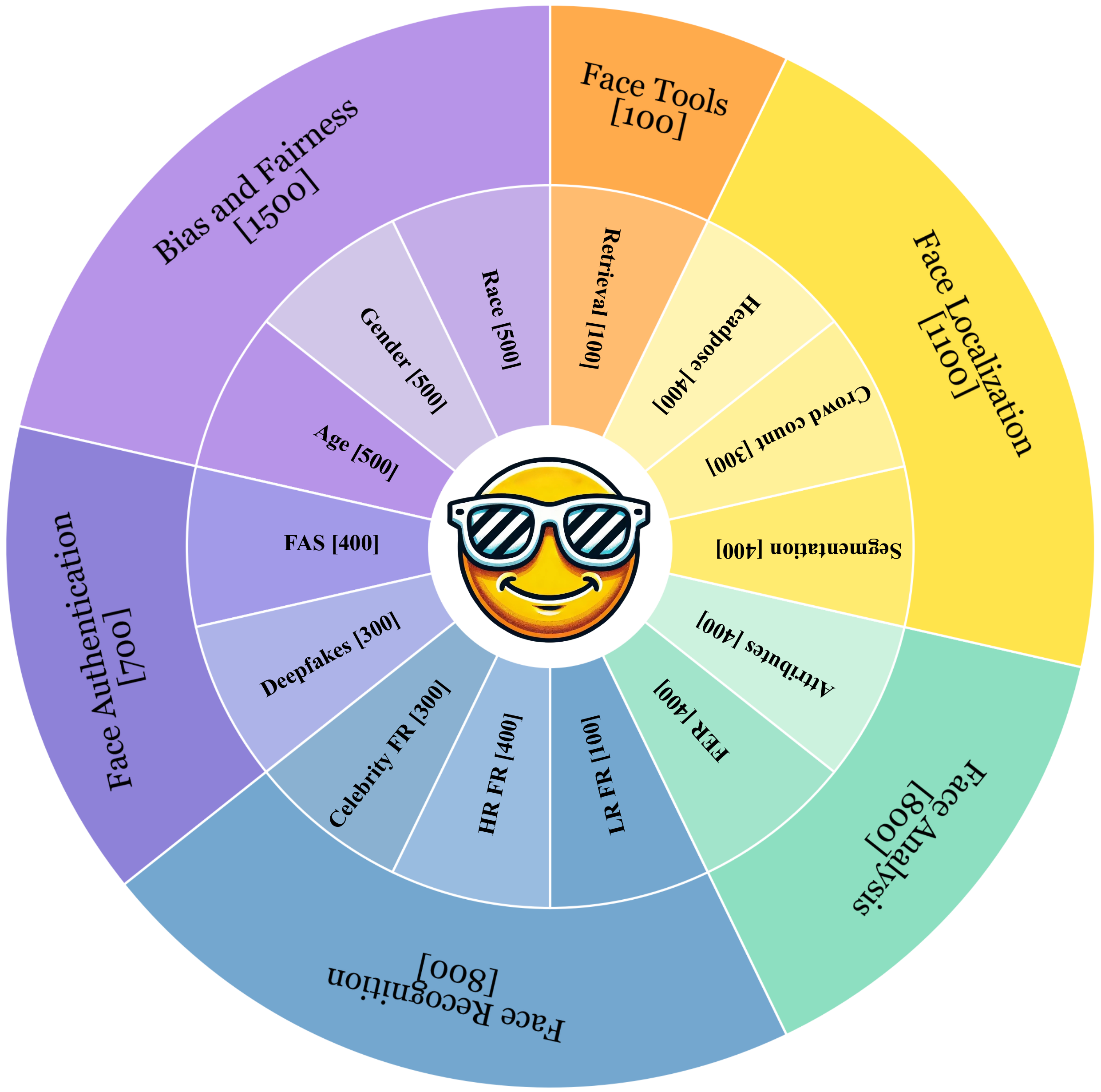}
    \caption{Distribution of questions across different categories and tasks in FaceXBench.}
    \label{fig:pie}
\end{figure}

%% file: sec/4_experiments.tex
\section{Experiments}
In this section, we detail the experiments conducted to benchmark and analyze various MLLMs on face understanding. We evaluate $2$ proprietary and $26$ open-source models, listed in Section~\ref{exp:model}. For a fair comparison, all models are evaluated in a zero-shot setting using the same base prompt. We analyze selected MLLMs under in-context and chain-of-thought settings. The evaluation settings and results are described in Sections~\ref{exp:evaluation} and~\ref{exp:main_results}, respectively.

\subsection{Models}
\label{exp:model}
The $2$ proprietary models used are GPT-4o~\cite{hurst2024gpt} and GeminiPro 1.5~\cite{team5gemini}. We divide the $26$ open-source models into three major categories based on parameter size: \textbf{(a) Open Source MLLMs (\textless4B parameters)}: PaliGemma~\cite{beyer2024paligemma}, LLaVA-OneVision-0.5b-OV~\cite{li2024llava}, and VILA 1.5-3b~\cite{lin2024vila}; \textbf{(b) Open Source MLLMs (4B-13B parameters)}: Chameleon-7b~\cite{chameleon}, Eagle-X4-8B-Plus~\cite{shi2024eagle}, Idefics2-8b~\cite{laurençon2024matters}, Idefics-9b-Instruct~\cite{laurencon2023obelics}, LLaVA-v1.5-7b~\cite{liu2024visual}, Monkey-Chat~\cite{li2024monkey}, MiniCPM-Llama3-v2.5~\cite{yao2024minicpmv}, LLaVA-OneVision-7b-SI~\cite{li2024llava}, LLaVA-NeXT-Interleave-7b~\cite{li2024llava_next}, Mantis-SIGLIP-8b~\cite{jiang2024mantis}, Phi-3.5-Vision~\cite{abdin2024phi}, LLaVA-OneVision-7b-OV, Qwen2-VL-7b-Instruct~\cite{Qwen2VL}, and InternVL2-8b~\cite{chen2024far}; \textbf{(c) Open Source MLLMs (\textgreater13B parameters)}: CogVLM2-19b~\cite{hong2024cogvlm2}, Idefics-80b-Instruct~\cite{laurencon2023obelics}, LLaVA-v1.5-13b~\cite{liu2024visual}, VILA 1.5-13b~\cite{lin2024vila}, InternVL-Chat-v1.5~\cite{chen2024far}, VILA 1.5-40b~\cite{lin2024vila}, LLaVA-OneVision-72b-OV~\cite{li2024llava}, Qwen2-VL-72b-Instruct~\cite{Qwen2VL}, and InternVL2-76b~\cite{chen2024far}. In the supplementary material, we provide detailed information regarding the architecture and the parameter size for all open-source MLLMs' evaluated, along with additional results under different settings.

\subsection{Evaluation Settings}
\label{exp:evaluation}
We evaluate the models under three settings: (a) \textit{zero-shot}, (b) \textit{in-context task description}, and (c) \textit{chain-of-thought} prompting. In the zero-shot setting, we pass the input with only the base prompt. In the in-context task description setting, we prepend the prompt with a brief description of the specific task required by the question. In the chain-of-thought setting, we prompt the model to reason step-by-step before selecting the correct option. The task-specific prepended text and additional details of the evaluation settings are provided in the supplementary material.

The proposed benchmark consists of multiple-choice questions (MCQs), which standardize the evaluation across different models. We empirically construct a diverse set of regular expressions and employ a three-step evaluation strategy to extract the correct option in cases where intermediate reasoning or calculations are present in the model's output.
Our evaluation strategy is as follows: First, we use a regular expression to match the correct option (A, B, C, or D) at the beginning of the model’s prediction. If this attempt fails, we then search for the correct option within the entire prediction text. As a final fallback, we compare parts of the predicted output with the option values to find a match. If none of these steps succeed, we categorize the prediction as incorrect. The complete function is provided in the supplementary material. Alternatively, we could implement a random-choice or frequent-choice strategy instead of labeling responses as incorrect. In the random-choice approach, a random option is selected as the prediction, while in the frequent-choice approach, we select the most common correct option in the dataset. However, these strategies can lead to misleading results, especially when the model fails to respond due to content moderation measures or poor instruction-following.

\subsection{Implementation Details}
To ensure reproducibility and efficiency, we utilized the open-source \texttt{VLMEvalKit}~\cite{duan2024vlmevalkit} for all experiments. The \texttt{VLMEvalKit} provides a comprehensive framework for evaluating vision-language models, streamlining the experimental workflow. Below, we outline the key libraries and parameter settings used in our implementation:

\begin{itemize}
    \item \textbf{VLMEvalKit:} We utilized the official open-source implementation shared and maintained on GitHub, available at  \url{https://github.com/open-compass/VLMEvalKit}.
    \item \textbf{PyTorch:} All experiments were conducted using PyTorch version \texttt{2.1.2}.
    \item \textbf{Hugging Face Transformers:} We leveraged open-source MLLMs hosted on Hugging Face. Following the \texttt{VLMEvalKit} documentation, we used the corresponding Transformer packages for the respective models.
    \item \textbf{NumPy:} Version \texttt{1.21.2} was employed for efficient numerical computations, particularly for array and matrix operations.
    \item \textbf{Matplotlib} and \textbf{Seaborn:} For data visualization, we used Matplotlib version \texttt{3.4.3} and Seaborn version \texttt{0.11.2}.
\end{itemize}

%% file: sec/5_results.tex
\section{Results}
\label{exp:main_results}
\input{tables/results}
The performance of various models on FaceXBench is shown in Table~\ref{tab:results}. The results emphasize the challenging nature of the benchmark, with no model achieving more than $60$\% accuracy. We observe that Qwen2-VL-72b-Instruct~\cite{Qwen2VL} achieves the best overall performance of 57.86\%. InternVL2-76b~\cite{chen2024far}, GeminiPro 1.5~\cite{team5gemini}, Qwen2-VL-72b-Instruct~\cite{Qwen2VL}, LLaVA-OneVision-72b-OV~\cite{li2024llava}, Qwen2-VL-72b-Instruct~\cite{Qwen2VL} and GeminiPro 1.5~\cite{team5gemini} achieve the highest performances in the categories of \textit{Bias \& Fairness}, \textit{Face Recognition}, \textit{Face Authentication}, \textit{Face Analysis}, \textit{Face Localization} and \textit{Face Tools Use} with accuracies of $69.53$\%, $70.00$\%, $41.14$\%, $63.25$\%, $55.45$\% and $57.00$\%, respectively. Human performance at $70.28$\% and SOTA vision models at $84.50$\%, serve as the upper bounds for the benchmark. The significant gap between the current state-of-the-art MLLM and this upper bound highlights the substantial room for improvement.
\input{tables/results_tasks}
\input{figures/ablation}
\input{tables/ablation}

\noindent \textbf{Results across different evaluation categories.}
The performance of various models on FaceXBench, as shown in Table~\ref{tab:results} highlights the limited face understanding capabilities of current models. Specifically, they perform poorly on tasks such as face authentication and face localization, which require fine-grained facial feature extraction and spatial understanding of facial structure. The low performance in the \textit{``Bias \& Fairness''} category suggests that existing MLLMs exhibit biases toward certain age, gender, and racial groups, which need to be mitigated prior to deployment. Surprisingly, GPT-4o performs poorly in the \textit{``Bias \& Fairness category''}, but on examination, it was found that the model often chose not to answer due to its safety alignment. The top-performing models achieve around $70$\% performance in face recognition tasks; however, their performance drops significantly on low-resolution face recognition. This decline can be attributed to training data predominantly comprising of high-resolution images, which limits models' effectiveness on low-resolution inputs. Some models may have been trained on image-text datasets with face images, such as FFHQ-Text~\cite{zhou2021generative}, CelebA-Dialog~\cite{jiang2021talkedit}, LAION-Face~\cite{zheng2022general}, FaceCaption-15M~\cite{dai202415m}, as part of large-scale pretraining. Although these datasets often contain attribute and expression information in text, the models still perform poorly in the \textit{``Face Analysis''} section, indicating substantial room for improvement.

\noindent  \textbf{Open Source vs Closed Source Models.}
We observe that open-source models, such as InternVL2 and Qwen2-VL, outperform proprietary models like GPT-4o and GeminiPro 1.5, achieving accuracies of $57.80$\% and $57.86$\%, compared to $50.50$\% and $56.96$\%, respectively. Generally, MLLMs have shown a trend where closed-source models outperform open-source ones. However, in the sensitive domain of face analysis, proprietary models are safety-aligned before deployment. The relatively poor performance of proprietary models in \textit{``Bias \& Fairness''} and \textit{``Face Analysis''} is primarily due to content moderation constraints. Notably, GeminiPro 1.5 demonstrates comparatively better performance than other models on \textit{``Face Tools Use''}, showcasing its ability to leverage specialized tools for complex scenarios involving multiple face-related tasks.

\noindent \textbf{Performance across various tasks.}
To gauge the difficulty of various tasks, we plot the performance of the top-5 models ($4$B-$13$B parameters) across each task, as shown in Figure~\ref{fig:ablation}(a). To provide task-specific insights, we summarize the task-wise performance of various models in Table~\ref{tab:results_tasks}. We observe that, on average, models struggle with tasks such as crowd counting, deepfake detection, head pose estimation, and low-resolution face recognition, highlighting areas where MLLMs' need improvement. In contrast, gender prediction is one of the easier tasks, with an average performance of approximately $80$\%. Additionally, we plot the average performance of these top-5 models on multiple-image and single-image question. Figure~\ref{fig:ablation}(c) shows that models generally perform poorly on questions with multiple images as inputs compared to single-image questions. This is primarily because the models are required to process more visual information and compare and contrast facial features across multiple images, making the questions more challenging. Furthermore, models such as Qwen2-VL~\cite{Qwen2VL} and LLaVA-OneVision~\cite{li2024llava}, which use dynamic resolution to handle arbitrary image resolutions, perform better than models on segmentation task. This mapping of images to a dynamic number of visual tokens more closely resembles human-like visual processing and leads to improved performance in face-understanding tasks. 

\noindent \textbf{Effect of LLM and it's size on performance.}
To analyze the impact of LLM on model performance, we plot a performance curve for models that share the same SigLIP SO400M/14@384 vision encoder but differ in their LLM backbone. From Figure~\ref{fig:ablation}(b), we observe that performance improves with an increase in LLM size. Furthermore, when comparing LLMs of different sizes within the same family, such as Qwen2, we see that the curve shape remains consistent, with performance values translating upward across all dimensions. This indicates that as LLM size increases, model capabilities improve proportionately across various dimensions, maintaining a similar pattern of performance enhancement.

\noindent \textbf{Performance under different evaluation settings.}
We evaluate three selected models under the \textit{in-context task description} and \textit{chain-of-thought} settings, with results summarized in Table.~\ref{tab:ablation}. In the in-context setting, we observe some performance improvements in \textit{face authentication} and \textit{face tools use}. However, overall performance drops, indicating that the models struggle to utilize in-context information effectively. In the chain-of-thought setting, we observe a substantial decline in performance, suggesting that, although these models exhibit reasoning capabilities, they do not transfer effectively to face understanding.

%% file: tables/results.tex
\begin{table*}[t]
    \centering
    \resizebox{\textwidth}{!}{
    \begin{tabular}{@{}l>{\columncolor[HTML]{EFEFEF}}c cccccc@{}}
        \toprule
        \textbf{\begin{tabular}[c]{@{}c@{}}Models \\  \end{tabular}} & \textbf{\begin{tabular}[c]{@{}c@{}}Overall \\  \end{tabular}} & \textbf{\begin{tabular}[c]{@{}c@{}}Bias \& \\  Fairness\end{tabular} } & \textbf{\begin{tabular}[c]{@{}c@{}}Face \\  Recognition\end{tabular}} & \textbf{\begin{tabular}[c]{@{}c@{}}Face \\  Authentication\end{tabular}} & \textbf{\begin{tabular}[c]{@{}c@{}}Face \\  Analysis\end{tabular}} & \textbf{\begin{tabular}[c]{@{}c@{}}Face \\ Localization\end{tabular}} & \textbf{\begin{tabular}[c]{@{}c@{}}Face \\  Tools Use\end{tabular}} \\
        (28) & (5,000) & (1,500) & (8,00) & (700) & (800) & (1,100) & (100) \\ \midrule
        \textcolor{gray}{Random Choice} & \textcolor{gray}{25.10} & \textcolor{gray}{24.73} & \textcolor{gray}{26.88} & \textcolor{gray}{22.71} & \textcolor{gray}{24.75} & \textcolor{gray}{25.64} & \textcolor{gray}{30.00} \\
        \textcolor{gray}{Frequent Choice} & \textcolor{gray}{26.68} & \textcolor{gray}{25.60} & \textcolor{gray}{26.88} & \textcolor{gray}{35.14} & \textcolor{gray}{28.38} & \textcolor{gray}{26.55} & \textcolor{gray}{40.00} \\
        \textcolor{gray}{Human} & \textcolor{gray}{70.28} & \textcolor{gray}{72.33} & \textcolor{gray}{65.50} & \textcolor{gray}{66.00} & \textcolor{gray}{76.12} & \textcolor{gray}{67.27} & \textcolor{gray}{94.00} \\
        \textcolor{gray}{Vision SOTA models} & \textcolor{gray}{84.50} & \textcolor{gray}{84.33} & \textcolor{gray}{81.87} & \textcolor{gray}{89.57} & \textcolor{gray}{91.37} & \textcolor{gray}{80.90} & \textcolor{gray}{57.00} \\
        \midrule
        \rowcolor[HTML]{FFF2CC}
        \multicolumn{8}{c}{\textbf{Open source MLLMs ($\mathbf{<4}\text{B}$ parameters)}} \\
        PaliGemma~\cite{beyer2024paligemma} & 32.22 & 35.67 & 26.50 & 28.00 & 37.62 & 32.27 & 12.00 \\
        LLaVA-OneVision-0.5b-OV~\cite{li2024llava} & 34.00 & 34.93 & 28.12 & 30.29 & 44.62 & 32.91 & 20.00 \\
        VILA 1.5-3b~\cite{lin2024vila} & 35.80 & 38.27 & 33.25 & 30.86 & 44.50 & 31.82 & 28.00 \\
        \midrule
        \rowcolor[HTML]{D9EAD3}
        \multicolumn{8}{c}{\textbf{Open source MLLMs ($\mathbf{4}\text{B}$ - $\mathbf{13}\text{B}$ parameters)}} \\
        Chameleon-7b~\cite{chameleon} & 17.04 & 10.27 & 17.12 & 6.86 & 20.25 & 28.91 & 33.00 \\
        Eagle-X4-8B-Plus~\cite{shi2024eagle} & 31.44 & 25.00 & 23.12 & 30.00 & 35.62 & 43.64 & 37.00 \\
        Idefics-9b-Instruct~\cite{laurencon2023obelics} & 34.58 & 37.93 & 28.62 & 34.43 & 37.38 & 34.18 & 15.00 \\
        LLaVA-v1.5-7b~\cite{liu2024visual} & 36.22 & 41.20 & 33.12 & 30.14 & 43.50 & 32.18 & 15.00 \\
        Monkey-Chat~\cite{li2024monkey} & 37.40 & 39.00 & 31.50 & 26.00 & 44.00 & 41.73 & 40.00 \\
        MiniCPM-Llama3-v2.5~\cite{yao2024minicpmv} & 40.70 & 45.80 & 29.88 & 32.86 & 52.38 & 40.45 & 15.00 \\
        LLaVA-NeXT-Interleave-7b~\cite{li2024llava_next} & 43.80 & 52.53 & 38.00 & 38.57 & 55.88 & 32.27 & 26.00 \\
        LLaVA-OneVision-7b-SI~\cite{li2024llava} & 44.32 & 50.73 & 32.75 & 29.86 & 52.25 & 47.27 & 46.00 \\
        Idefics2-8b~\cite{laurençon2024matters} & 44.52 & 52.67 & 31.25 & 33.57 & 53.25 & 43.91 & 42.00 \\
        Mantis-SIGLIP-8b~\cite{jiang2024mantis} & 44.60 & 56.13 & 45.12 & 36.86 & 48.00 & 31.64 & 37.00 \\
        Phi-3.5-Vision~\cite{abdin2024phi} & 45.16 & 52.47 & 50.12 & 40.00 & 51.00 & 31.64 & 34.00 \\
        LLaVA-OneVision-7b-OV~\cite{li2024llava} & 48.98 & 61.40 & 38.38 & 35.57 & 55.12 & 44.82 & 38.00 \\
        Qwen2-VL-7b-Instruct~\cite{Qwen2VL} & 51.58 & 57.47 & 57.88 & 34.00 & 57.50 & 47.09 & 38.00 \\
        InternVL2-8b~\cite{chen2024far} & 53.24 & 62.40 & 61.75 & 35.43 & 55.38 & 45.09 & 45.00 \\
        \midrule
        \rowcolor[HTML]{F4CCCC}
        \multicolumn{8}{c}{\textbf{Open source MLLMs ($\mathbf{>13}\text{B}$ parameters)}} \\
        Idefics-80b-Instruct~\cite{laurencon2023obelics} & 35.86 & 39.87 & 35.12 & 27.71 & 35.12 & 38.55 & 15.00\\
        LLaVA-v1.5-13b~\cite{liu2024visual} & 39.88 & 44.60 & 34.88 & 34.14 & 44.75 & 37.27 & 39.00 \\
        VILA 1.5-13b~\cite{lin2024vila} & 40.00 & 45.07 & 40.00 & 28.43 & 49.25 & 34.18 & 35.00 \\
        CogVLM2-19b~\cite{hong2024cogvlm2} & 40.46 & 43.13 & 33.88 & 35.71 & 45.62 & 41.91 & 29.00 \\
        InternVL-Chat-v1.5~\cite{chen2024far} & 49.18 & 59.73 & 41.38 & 33.00 & 55.12 & 46.73 & 46.00 \\
        VILA 1.5-40b~\cite{lin2024vila} & 55.48 & 64.00 & 57.63 & 33.14 & 60.50 & 54.36 & 39.00 \\
        LLaVA-OneVision-72b-OV~\cite{li2024llava} & 56.42 & 66.53 & 52.00 & 37.43 & \textbf{\textcolor{blue}{63.25}} & 53.73 & 48.00 \\
        InternVL2-76b~\cite{chen2024far} & 57.80 & \textbf{\textcolor{blue}{69.53}} & 66.62 & 36.14 & 62.00 & 47.18 & 46.00 \\
        Qwen2-VL-72b-Instruct~\cite{Qwen2VL} & \textbf{\textcolor{blue}{57.86}} & 62.20 & 69.12 & \textbf{\textcolor{blue}{41.14}} & 57.88 & \textbf{\textcolor{blue}{55.45}} & 46.00 \\
        \midrule
        \rowcolor[HTML]{D9D2E9}
        \multicolumn{8}{c}{\textbf{Proprietary MLLMs}} \\
        GPT-4o~\cite{hurst2024gpt} & 50.50 & 46.93 & 55.62 & 40.00 & 62.25 & 50.36 & 44.00 \\
        GeminiPro 1.5~\cite{team5gemini} & 56.96 & 67.40 & \textbf{\textcolor{blue}{70.00}} & 35.00 & 58.13 & 46.36 & \textbf{\textcolor{blue}{57.00}} \\
        \bottomrule
    \end{tabular}
    }
    \caption{Results of different models on the FaceXBench. We categorize the open-source models in three categories based on parameter size: (a) Open source MLLMs (\textless4B parameters), (b) Open source MLLMs (4B-13B parameters), (c) Open source MLLMs (\textgreater13B parameters). Additionally, we evaluate (d) proprietary models. The best-performing model in each category is highlighted in \textbf{\textcolor{blue}{bold}}.}
    \label{tab:results}
\end{table*}

%% file: tables/results_tasks.tex
\begin{table*}[p]
    \centering
    \rotatebox{90}{
    \resizebox{0.95\textheight}{!}{
    \begin{tabular}{@{}l>{\columncolor[HTML]{EFEFEF}}c cccccccccccccc@{}}
        \toprule
        \textbf{Models} & \textbf{Overall} & \textbf{\begin{tabular}[c]{@{}c@{}} Expression \\ Recognition\end{tabular}} & \textbf{\begin{tabular}[c]{@{}c@{}}Age \\ Estimation\end{tabular}} & \textbf{\begin{tabular}[c]{@{}c@{}}Race \\ Estimation\end{tabular}} & \textbf{\begin{tabular}[c]{@{}c@{}}Crowd \\ Counting\end{tabular}} & \textbf{\begin{tabular}[c]{@{}c@{}}Celebrity \\ Identification\end{tabular}} & \textbf{HR FR} & \textbf{LR FR} & \textbf{\begin{tabular}[c]{@{}c@{}}Face \\ Anti-spoofing\end{tabular}} & \textbf{\begin{tabular}[c]{@{}c@{}}Tools \\ Retrieval\end{tabular}} & \textbf{\begin{tabular}[c]{@{}c@{}}Attributes \\ Prediction\end{tabular}} & \textbf{\begin{tabular}[c]{@{}c@{}}Gender \\ Prediction\end{tabular}} & \textbf{\begin{tabular}[c]{@{}c@{}}Headpose \\ Estimation\end{tabular}} & \textbf{\begin{tabular}[c]{@{}c@{}}Face \\ Parsing\end{tabular}} & \textbf{\begin{tabular}[c]{@{}c@{}}Deepfake \\ Detection\end{tabular}} \\
        \midrule
        \textcolor{gray}{Random Choice} & \textcolor{gray}{25.10} & \textcolor{gray}{21.75} & \textcolor{gray}{24.20} & \textcolor{gray}{24.40} & \textcolor{gray}{26.00} & \textcolor{gray}{29.00} & \textcolor{gray}{25.25} & \textcolor{gray}{27.00} & \textcolor{gray}{20.75} & \textcolor{gray}{30.00} & \textcolor{gray}{27.75} & \textcolor{gray}{25.60} & \textcolor{gray}{23.25} & \textcolor{gray}{27.75} & \textcolor{gray}{25.33} \\
        \textcolor{gray}{Frequent Choice} & \textcolor{gray}{26.68} & \textcolor{gray}{27.50} & \textcolor{gray}{27.40} & \textcolor{gray}{27.20} & \textcolor{gray}{28.00} & \textcolor{gray}{28.67} & \textcolor{gray}{26.00} & \textcolor{gray}{35.00} & \textcolor{gray}{41.25} & \textcolor{gray}{40.00} & \textcolor{gray}{33.50} & \textcolor{gray}{28.20} & \textcolor{gray}{26.25} & \textcolor{gray}{26.50} & \textcolor{gray}{28.00} \\
        \textcolor{gray}{Human} & \textcolor{gray}{70.28} & \textcolor{gray}{77.75} & \textcolor{gray}{47.00} & \textcolor{gray}{75.20} & \textcolor{gray}{40.33} & \textcolor{gray}{32.67} & \textcolor{gray}{91.75} & \textcolor{gray}{59.00} & \textcolor{gray}{71.75} & \textcolor{gray}{94.00} & \textcolor{gray}{74.50} & \textcolor{gray}{94.80} & \textcolor{gray}{60.75} & \textcolor{gray}{94.00} & \textcolor{gray}{58.33} \\
        \textcolor{gray}{Vision SOTA models} & \textcolor{gray}{84.50} & \textcolor{gray}{91.00} & \textcolor{gray}{68.00} & \textcolor{gray}{88.60} & \textcolor{gray}{59.33} & \textcolor{gray}{63.00} & \textcolor{gray}{98.00} & \textcolor{gray}{74.00} & \textcolor{gray}{90.25} & \textcolor{gray}{57.00} & \textcolor{gray}{91.75} & \textcolor{gray}{96.40} & \textcolor{gray}{81.25} & \textcolor{gray}{96.75} & \textcolor{gray}{88.67} \\
        \midrule
        \rowcolor[HTML]{FFF2CC}
        \multicolumn{16}{c}{\textbf{Open source MLLMs ($\mathbf{<4}\text{B}$ parameters)}} \\
        PaliGemma-3b & 32.22 & 46.50 & 28.20 & 38.40 & 25.67 & 31.00 & 23.75 & 24.00 & 28.50 & 12.00 & 28.75 & 40.40 & 24.00 & 45.50 & 27.33 \\
        LLaVA-OneVision-Qwen2-0.5b & 34.00 & 50.00 & 23.00 & 34.40 & 35.33 & 36.67 & 23.50 & 21.00 & 36.25 & 20.00 & 39.25 & 47.40 & 32.25 & 31.75 & 22.33 \\
        VILA1.5-3b & 35.80 & 53.25 & 32.00 & 35.00 & 23.67 & 44.67 & 26.00 & 28.00 & 38.75 & 28.00 & 35.75 & 47.80 & 31.50 & 38.25 & 20.33 \\
        \midrule
        \rowcolor[HTML]{D9EAD3}
        \multicolumn{16}{c}{\textbf{Open source MLLMs ($\mathbf{4}\text{B}$ - $\mathbf{13}\text{B}$ parameters)}} \\
        Chameleon-7b & 17.04 & 15.00 & 15.80 & 5.80 & 23.00 & 28.00 & 11.25 & 8.00 & 8.50 & 33.00 & 25.50 & 9.20 & 20.00 & 42.25 & 4.67 \\
        Eagle-X4-8B-Plus & 31.44 & 39.00 & 26.20 & 24.20 & 24.33 & 22.33 & 23.75 & 23.00 & 38.00 & 37.00 & 32.25 & 24.60 & 27.25 & 74.50 & 19.33 \\
        Idefics-9b-Instruct & 34.58 & 45.00 & 27.60 & 40.60 & 35.33 & 41.67 & 20.50 & 22.00 & 42.25 & 15.00 & 29.75 & 45.60 & 30.50 & 37.00 & 24.00 \\
        LLaVA-v1.5-7b & 36.22 & 50.75 & 32.60 & 46.80 & 31.00 & 46.67 & 26.50 & 19.00 & 32.50 & 15.00 & 36.25 & 44.20 & 28.75 & 36.50 & 27.00 \\
        Monkey-Chat & 37.40 & 51.00 & 31.20 & 36.40 & 29.33 & 48.67 & 19.25 & 29.00 & 26.25 & 40.00 & 37.00 & 49.40 & 26.75 & 66.00 & 25.67 \\
        MiniCPM-Llama3-v2.5 & 40.70 & 54.75 & 34.60 & 45.60 & 27.67 & 49.67 & 17.00 & 22.00 & 37.00 & 15.00 & 50.00 & 57.20 & 30.50 & 60.00 & 27.33 \\
        LLaVA-Next-Interleave-7b & 43.80 & 55.88 & 42.40 & 44.00 & 19.00 & 43.33 & 37.50 & 24.00 & 45.00 & 26.00 & 54.25 & 71.20 & 28.00 & 46.50 & 30.00 \\
        LLaVA-OneVision-Qwen2-7b-SI & 44.32 & 52.00 & 35.00 & 52.60 & 24.67 & 48.67 & 23.75 & 21.00 & 39.25 & 46.00 & 52.50 & 64.60 & 35.00 & 76.50 & 17.33 \\
        Idefics2-8b & 44.52 & 57.50 & 42.60 & 48.00 & 31.00 & 49.67 & 21.00 & 17.00 & 38.50 & 42.00 & 49.00 & 67.40 & 33.00 & 64.50 & 27.00 \\
        Mantis-SIGLIP-8b & 44.60 & 50.25 & 41.00 & 46.60 & 21.00 & 56.00 & 42.25 & 24.00 & 44.25 & 37.00 & 45.75 & 80.80 & 27.50 & 43.75 & 27.00 \\
        Phi-3.5-Vision & 45.16 & 51.00 & 39.40 & 45.00 & 29.00 & 53.67 & 49.25 & 43.00 & 49.75 & 34.00 & 51.00 & 73.00 & 25.50 & 39.75 & 27.00 \\
        LLaVA-OneVision-Qwen2-7b-OV & 48.98 & 56.25 & 42.20 & 61.00 & 23.67 & 49.67 & 34.25 & 21.00 & 43.00 & 38.00 & 54.00 & 81.00 & 31.00 & 74.50 & 25.67 \\
        Qwen2-VL-7b-Instruct & 51.58 & 56.75 & 43.80 & 49.00 & 20.67 & 56.33 & 65.50 & 32.00 & 39.00 & 38.00 & 58.25 & 79.60 & 32.00 & 82.00 & 27.33 \\
        InternVL2-8b & 53.24 & 60.75 & 44.40 & 61.80 & 31.67 & 57.33 & 69.50 & 44.00 & 44.75 & 45.00 & 50.00 & 81.00 & 33.50 & 66.75 & 23.00 \\
        \midrule
        \rowcolor[HTML]{F4CCCC}
        \multicolumn{16}{c}{\textbf{Open source MLLMs ($\mathbf{>13}\text{B}$ parameters)}} \\
        Idefics-80b-Instruct & 35.86 & 36.50 & 36.00 & 34.00 & 32.67 & 53.33 & 23.75 & 26.00 & 30.75 & 15.00 & 33.75 & 49.60 & 31.00 & 50.50 & 23.67 \\
        LLaVA-v1.5-13b & 39.88 & 48.75 & 34.80 & 51.00 & 35.00 & 54.33 & 23.75 & 21.00 & 40.75 & 39.00 & 40.75 & 48.00 & 24.25 & 52.00 & 25.33 \\
        VILA1.5-13b & 40.00 & 57.50 & 35.00 & 41.80 & 25.00 & 57.33 & 30.25 & 27.00 & 35.50 & 35.00 & 41.00 & 58.40 & 28.75 & 46.50 & 19.00 \\
        CogVLM2-Llama3-19b & 40.46 & 45.75 & 31.20 & 49.80 & 31.00 & 53.33 & 19.50 & 33.00 & 44.25 & 29.00 & 45.50 & 48.40 & 28.50 & 63.50 & 24.33 \\
        InternVL-Chat-v1.5 & 49.18 & 63.50 & 42.60 & 56.80 & 24.67 & 56.00 & 35.25 & 22.00 & 40.00 & 46.00 & 46.75 & 79.80 & 27.25 & 82.75 & 23.67 \\
        VILA1.5-40b & 55.48 & 63.25 & 48.80 & 60.80 & \textbf{\textcolor{blue}{41.67}} & 66.67 & 57.50 & 31.00 & 36.50 & 39.00 & 57.75 & 82.40 & 30.50 & \textbf{\textcolor{blue}{87.75}} & 28.67 \\
        LLaVA-OneVision-Qwen2-72b-OV & 56.42 & 61.75 & 49.40 & \textbf{\textcolor{blue}{67.80}} & 37.33 & 65.67 & 47.75 & 28.00 & 43.50 & 48.00 & 64.75 & 82.40 & 38.25 & 81.50 & 29.33 \\
        InternVL2-76b & 57.80 & \textbf{\textcolor{blue}{66.25}} & \textbf{\textcolor{blue}{52.80}} & 65.60 & 28.00 & 68.67 & 71.00 & 43.00 & 42.25 & 46.00 & 57.75 & \textbf{\textcolor{blue}{90.20}} & 34.25 & 74.50 & 28.00 \\
        Qwen2-VL-72b-Instruct & \textbf{\textcolor{blue}{57.86}} & 55.75 & 45.40 & 61.00 & 37.33 & \textbf{\textcolor{blue}{68.67}} & 76.25 & 42.00 & \textbf{\textcolor{blue}{50.75}} & 46.00 & 60.00 & 80.20 & \textbf{\textcolor{blue}{39.25}} & 85.25 & 28.33 \\
        \midrule
        \rowcolor[HTML]{D9D2E9}
        \multicolumn{16}{c}{\textbf{Proprietary MLLMs}} \\
        GPT-4o & 50.50 & 59.25 & 40.60 & 30.00 & 40.00 & 28.00 & 78.00 & \textbf{\textcolor{blue}{49.00}} & 47.75 & 44.00 & \textbf{\textcolor{blue}{65.25}} & 70.20 & 27.75 & 80.75 & \textbf{\textcolor{blue}{29.67}} \\
        GeminiPro 1.5 & 56.96 & 60.50 & 49.00 & 64.80 & 26.00 & 63.00 & \textbf{\textcolor{blue}{82.25}} & 42.00 & 41.75 & \textbf{\textcolor{blue}{57.00}} & 55.75 & 88.40 & 37.00 & 71.00 & 26.00 \\
        \bottomrule
    \end{tabular}
    }
    }
    \caption{Performance of various models across all evaluation tasks of FaceXBench.}
    \label{tab:results_tasks}
\end{table*}

%% file: figures/ablation.tex
\begin{figure*}[!t]
    \centering
    \includegraphics[width=\linewidth]{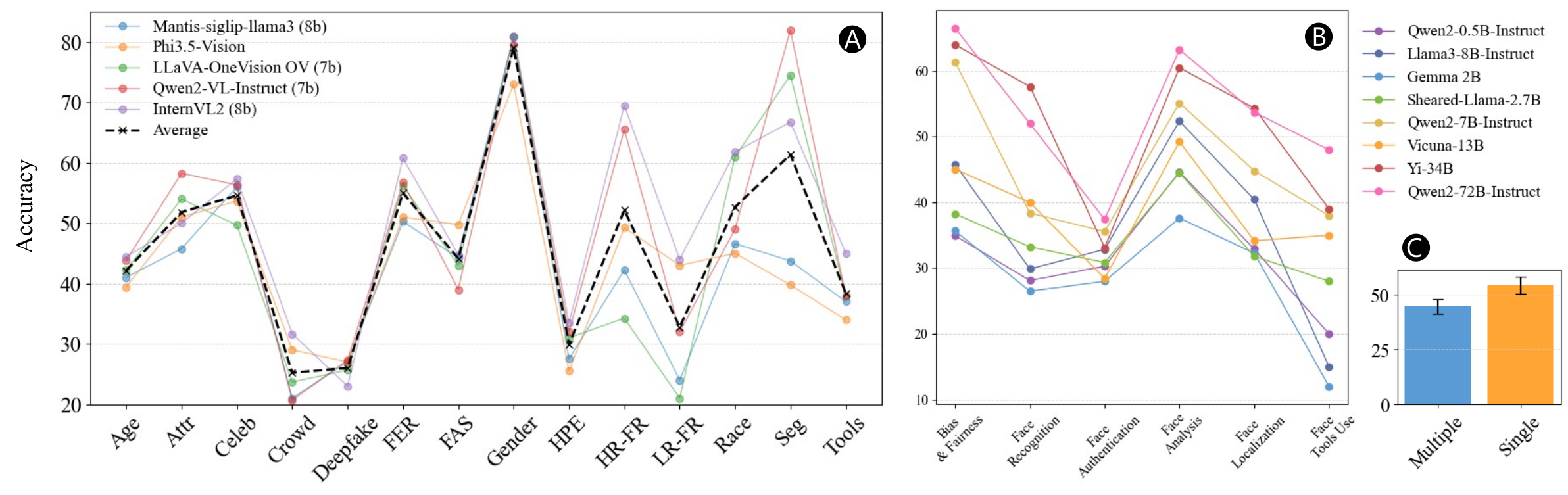}
    \caption{(a) Performance of top-5 models ($4$B-$13$B parameters) across various tasks. (b) Effect of LLM and it's size on model performance. (c) Average performance of the top-5 models ($4$B-$13$B parameters) on multiple-image and single-image questions.}
    \label{fig:ablation}
\end{figure*}

%% file: tables/ablation.tex
\begin{table}[t]
    \centering
    \resizebox{\columnwidth}{!}{
    \begin{tabular}{@{}l>{\columncolor[HTML]{EFEFEF}}c cccccc@{}}
        \toprule
        \textbf{Models} & \textbf{Overall} & \textbf{B \& F} & \textbf{FR} & \textbf{F Auth.} & \textbf{F Anlys.} & \textbf{FL} & \textbf{Tools} \\
        \midrule
        \rowcolor[HTML]{FFF2CC}
        \multicolumn{8}{c}{\textbf{In-context Description}} \\
        Phi-3.5-Vision~\cite{abdin2024phi} & -3.68 & -5.00 & -7.74 & -3.86 & -3.12 & +0.00 & +5.00 \\
        Qwen2-VL-7B~\cite{Qwen2VL} & -1.26 & -2.40 & -2.50 & +5.00 & -2.38 & -2.45 & +4.00 \\
        InternVL2-8B~\cite{chen2024far} & -1.10 & -3.13 & -0.50 & +6.00 & -2.01 & -3.00 & +03.00 \\
        \midrule
        \rowcolor[HTML]{FFF2CC}
        \multicolumn{8}{c}{\textbf{Chain of Thought}} \\
        Phi-3.5-Vision~\cite{abdin2024phi} & -11.80 & -20.20 & -19.87 & -4.71 & -13.50 & 1.72 & -6.00 \\
        Qwen2-VL-7B~\cite{Qwen2VL} & -11.96 & -13.74 & -16.76 & -6.86 & -12.12 & -10.27 & +0.00 \\
        InternVL2-8B~\cite{chen2024far} & -4.36 & -3.00 & -3.75 & -4.86 & -5.50 & -5.91 & +0.00 \\
        \bottomrule
    \end{tabular}
    }
    \\ \small (a)
    \\\textcolor{white}{a}\\
    \begin{minipage}{0.26\textwidth}
        \centering
        \resizebox{\textwidth}{!}{
        \begin{tabular}{lccc}
            \toprule
            \textbf{\begin{tabular}[c]{@{}c@{}}Models \\  \end{tabular}} & \textbf{\begin{tabular}[c]{@{}c@{}}Seg \\ (CelebMask)\end{tabular}} & \textbf{\begin{tabular}[c]{@{}c@{}}HPE \\ (BIWI)\end{tabular}} & \textbf{\begin{tabular}[c]{@{}c@{}}FER \\ (AffectNet)\end{tabular}} \\
            \midrule
            Phi-3.5-Vision~\cite{abdin2024phi} & +7.00 & +12.00 & +13.00 \\
            Qwen2-VL-7B~\cite{Qwen2VL} & +5.00 & +6.67 & +30.00 \\
            InternVL2-8B~\cite{chen2024far} & +4.00 & +7.33 & +23.00 \\
            \bottomrule
        \end{tabular}
        }
        \\ \small (b)
    \end{minipage}%
    \hspace{0.01\textwidth} 
    \begin{minipage}{.2\textwidth}
        \centering
        \resizebox{\textwidth}{!}{
        \begin{tabular}{lc}
            \toprule
            \textbf{Data} & \textbf{Acc.} \\
            \midrule
            FaceSFT & 33.58 \\
            LLaVA SFT (665k) + FaceSFT & 35.18 \\
            LLaVA SFT (200k) + FaceSFT & 35.24 \\
            \bottomrule
        \end{tabular}
        }
        \\ \small (c)
    \end{minipage}
    \caption{(a) Change in performance of select models under different evaluation settings. (b) Performance improvement seen by leveraging tool-use. (c) Results of finetune experiments on LLaVA1.5~\cite{li2024llava} using different data composition.}
    \label{tab:ablation}
\end{table}

%% file: sec/6_discussion.tex
\section{Discussion and Future Directions}
\label{sec:discussion}
In this section, we discuss possible future directions to improve the face understanding capabilities of MLLMs. The MLLM community has developed numerous supervised fine-tuning (SFT) datasets, such as COCO Caption~\cite{chen2015microsoft}, ScienceQA~\cite{saikh2022scienceqa}, Vision-FLAN~\cite{xu2024vision}, ChartQA~\cite{masry2022chartqa}, FigureQA~\cite{kahou2017figureqa}, Geometry3K~\cite{gao2023g}, MAVIS MCollect~\cite{zhang2024mavis}, MATHQA~\cite{amini2019mathqa}, TextOCR~\cite{singh2021textocr}, OCR-VQA~\cite{mishra2019ocr}, and MagpiePro~\cite{xu2024magpie}. These supervised fine-tuning sets target various skills, including document and chart understanding, mathematics, fine-grained perception, grounding, reasoning, and general OCR. Developing these general skills helps models perform better on existing tasks and enhances their reasoning and visual processing capabilities for improved zero-shot task performance. 

However, to answer the question, \textbf{\textit{``Will supervised fine-tuning improve the performance of MLLMs on face understanding?''}} we conducted multiple experiments by finetuning LLaVA1.5~\cite{liu2024improved} on a random $70$k subset of the FaceCaption15M~\cite{dai202415m}, which comprises image-text pairs containing attribute, age, and gender information. We fine-tuned the vision projector and the LLM backbone using LoRA~\cite{hu2021lora} and summarized the results for various data compositions in Table~\ref{tab:ablation}(c). Our findings reveal that naive fine-tuning on MLLM with only Face SFT data results in poor performance (Table~\ref{tab:ablation}(c), row 1), as models tend to lose their general reasoning and perception capabilities. We incorporated the complete LLaVA SFT dataset ($665$k samples) alongside the Face SFT, and the model achieved a score of 35.18 (Table~\ref{tab:ablation}(c), row 2). We randomly sampled $200$k examples from the full $665$k LLaVA 1.5 dataset, combining them with the 70k Face SFT data for fine-tuning, resulting in an improved score of $35.24$ (Table~\ref{tab:ablation}(c), row 3). This experiment demonstrates the benefit of integrating FaceSFT data in the right proportion with existing reasoning and instruction-tuning datasets.

We explored an alternative research direction to improve MLLMs’ performance on face understanding using specialized tools. To investigate the benefits of tool use, we selected specific datasets and converted predictions from state-of-the-art models into text, which we then provided as context to the MLLM. As shown in Table~\ref{tab:ablation}(b), this approach resulted in a significant performance boost. We believe equipping MLLMs with agentic behavior through specialized tools is a promising way to enhance their face understanding capabilities.
\noindent \textbf{For Future Work}: We advocate the development of diverse supervised fine-tuning sets covering various aspects of face understanding and training MLLMs to leverage specialized tools. Following this direction, we believe future MLLMs will be capable of advanced face understanding, with FaceXBench serving as a critical resource for monitoring progress by benchmarking models across multiple dimensions of face understanding.

%% file: sec/8_conclusion.tex
\section{Conclusion}
We propose FaceXBench, a comprehensive benchmark for face understanding that consists of 5,000 multiple-choice questions. It covers 14 tasks across six broad categories and is derived from 26 datasets. We employed a three-step data collection process to convert existing datasets into a VQA format, implementing quality control checks at each step. We conducted a thorough evaluation of 26 open-source models and two proprietary models, GPT-4o and GeminiPro 1.5, revealing the limitations of these models in face understanding. Our analysis examines performance across various dimensions and tasks, identifying factors that impact model performance and providing valuable insights. Finally, we discuss possible future directions, advocating for the development of instruction-tuning datasets, with FaceXBench serving as a catalyst for further research.

\section*{Acknowledgment}
This research is based upon work supported in part by the Office of the Director of National Intelligence (ODNI), Intelligence Advanced Research Projects Activity (IARPA), via [2022-21102100005]. The views and conclusions contained herein are those of the authors and should not be interpreted as necessarily representing the official policies, either expressed or implied, of ODNI, IARPA, or the U.S. Government. The US Government is authorized to reproduce and distribute reprints for governmental purposes notwithstanding any copyright annotation therein.

%% file: appendix/X_motivation.tex
\section{Motivation}
\label{motivation}
We already discussed \textquotedblleft \textit{\textbf{Why should MLLMs be proficient in face understanding?}}\textquotedblright in Introduction~\ref{introduction}. To reiterate, MLLMs are increasingly deployed as central processors in various advanced applications, including virtual-reality headsets~\cite{konenkov2024vr}, embodied AI~\cite{huang2023modality, driess2023palm}, driving safety~\cite{hwang2024emma, sreeram2024probing}, authentication~\cite{deandres2024good}, human-computer interaction~\cite{wang2024large}, and sports analysis~\cite{xia2024sportu}. In these applications, accurate face understanding is crucial, as face images appear frequently and require accurate face understanding for appropriate responses. Below we answer some additional questions that frame our motivation and benchmark design.

\textquotedblleft \textit{\textbf{Why do we need MLLMs to accomplish face analysis task,  which are mostly perception-based, when MLLMs are not explicitly trained for these tasks?}}\textquotedblright. To clarify, our objective is not to solve face analysis tasks using MLLMs, nor do we aim to outperform specialized face analysis models on established benchmarks. Instead, our focus is on evaluating MLLMs' ability to understand faces, which is a fundamental aspect of general visual and multimodal reasoning. Strong face understanding is essential for a variety of real-world applications, as discussed in the previous question. Given that MLLMs are expected to serve as general-purpose vision-language models, assessing their performance in this domain provides valuable insights into their limitations and potential areas for improvement. FaceXBench serves as a standardized benchmark to systematically evaluate and monitor the progress of MLLMs in face understanding. By identifying gaps and shortcomings in their capabilities, our work can help guide future advancements in MLLMs, ensuring that they develop stronger and more reliable face understanding skills.

\textquotedblleft \textit{\textbf{Face analysis tasks are primarily fine-grained and vision-centric, requiring little to no language reasoning. Why is it necessary to evaluate MLLMs on face understanding rather than using specialized vision models for face analysis and then passing the results to LLMs?}}\textquotedblright. We acknowledge that using an external tool or agent leveraging specialized vision models and then passing the results to LLMs would likely yield better performance in face analysis tasks. To validate this, we have included tool-use results in Table 3(b) of the main paper. However, the primary goal of FaceXBench is not to optimize performance through external assistance but to assess the inherent face-understanding capabilities of existing MLLMs. As discussed in the introduction (lines 46–72) of main paper, we argue that MLLMs should possess strong face-understanding skills intrinsically, as this ability is crucial for a wide range of real-world multimodal applications. Many general tasks such as identity-aware reasoning, social interaction analysis, and facial emotion recognition require a model to process and reason about facial features without relying on a separate vision module. Evaluating MLLMs independently allows us to benchmark their progress, identify limitations, and drive improvements in their multimodal reasoning capabilities, ultimately making them more robust for downstream applications where face understanding is essential.

\textquotedblleft \textit{\textbf{What is the rationale behind our selected tasks? For example, why do we not include face detection or prompt MLLMs to generate face parsing maps?}}\textquotedblright. We select tasks that are most relevant to the capabilities and real-world applications of MLLMs. While face detection and direct face segmentation are important computer vision tasks, we exclude them from our benchmark because current MLLMs generally fail to handle them effectively and falls outside the intended use of MLLMs. Using a large MLLM to carry out face detection or parsing is thus impractical , instead, we include tasks such as crowd counting and evaluating segmented attributes because these tests still probe an MLLM’s ability to localize and reason about faces, which is essential for many downstream applications that MLLMs are designed to solve. Again, we would like to emphasize that our goal is not to solve face analysis tasks outright; rather, we select and adapt tasks that suit MLLMs and align with their purpose and intended use.

\textquotedblleft \textit{\textbf{Why we not include open-ended visual question answering in the benchmark ?}}\textquotedblright. 
We chose not to include open-ended visual question answering (VQA) in our benchmark, following the precedent set by leading benchmarks such as MMBench~\cite{liu2025mmbench} [ECCV 2024], SEEDBench~\cite{li2024seed} [CVPR 2024], MMMU~\cite{yue2024mmmu} [CVPR 2024], and MathVista~\cite{lu2024mathvista} [ICLR 2024], which rely exclusively on multiple-choice questions (MCQs). The MCQ format offers key advantages: it enables standardized, reproducible, and scalable evaluation, while eliminating the ambiguity and subjectivity inherent in open-ended answers, especially critical in the facial domain, where expressions and attributes can be nuanced and difficult to judge consistently. In contrast, open-ended VQA often depends on human evaluation or external matching models, introducing non-determinism and inconsistency. By using MCQs, we ensure deterministic ground truth and fair comparisons across models. Importantly, MCQs do not simplify the reasoning task; models must still perform complex interpretation and decision-making. The format merely standardizes the output, making the evaluation process more reliable and accessible, and helping drive community-wide adoption and progress in facial understanding tasks.

%% file: appendix/X_limitations.tex
\section{Limitations}
Our benchmark has a limited number of questions that address scenarios where multiple faces appear within a single image, limiting its applicability in such contexts. Additionally, generative models, such as diffusion models, cannot be evaluated using the current benchmark, restricting its applicability for assessing performance in image generation tasks. Future work will address these limitations by extending the benchmark to include questions designed to evaluate image generation capabilities.

%% file: appendix/X_facexbench.tex
\section{FaceXBench}
In this section of the supplementary material, we will provide more information on the broad categories and tasks included in the FaceXBench. Additionally, we will provide details on the source datasets and the dataset statistics, with examples of images used in the benchmark. 

\subsection{FaceXBench Categories}
\noindent \textbf{Bias and Fairness:} In this category, we evaluate the model's ability to estimate age, predict gender, and identify race as key indicators of its understanding and analysis of demographic attributes. The focus extends beyond the accuracy of these predictions to ensuring fairness and inclusivity across diverse groups. By analyzing the model’s performance, we aim to uncover potential biases and ensure that predictions remain consistent and unbiased regardless of age group, gender, or race. This assessment is crucial for promoting fairness, mitigating the amplification of societal biases, and enhancing the model’s generalization capabilities for real-world applications.
\\

\noindent \textbf{Face Recognition:}
In this category, we evaluate the model's ability to perform accurate face recognition across various contexts, including high-resolution face recognition, low-resolution face recognition, and celebrity identification. These tasks test the model's proficiency in feature extraction, spatial awareness, and handling variations in image quality, lighting, and pose. High-resolution face recognition assesses the model's ability to leverage fine-grained details for precise identification, while low-resolution face recognition challenges its capability to generalize from limited information. Celebrity identification evaluates its knowledge base and contextual understanding of well-known individuals. 
\\

\noindent \textbf{Face Authentication:}
In this category, we evaluate the model's ability to perform robust face authentication, with a focus on critical tasks such as face anti-spoofing and deepfake detection. These tasks assess the model's capability to distinguish bonafide facial data from spoofing attempts, thereby ensuring the security and reliability of authentication systems. They are essential for safeguarding sensitive applications like identity verification, access control, and fraud prevention. By analyzing the model's performance, we aim to confirm that it demonstrates high sensitivity to subtle cues, generalizes effectively across diverse attack methods, and minimizes both false positives and false negatives. This assessment is vital for building trust in face authentication systems and addressing emerging threats in a rapidly evolving technological landscape.
\\

\noindent \textbf{Face Analysis:}
In this category, we evaluate the model's ability to analyze and interpret facial features through tasks such as facial attribute prediction and facial expression recognition. This evaluation focuses on the model’s ability to accurately identify static attributes, such as physical traits (e.g., glasses, hair color, or beard), which are essential for applications like targeted content delivery and user profiling. It also emphasizes the dynamic understanding of emotions and subtle expressions, including micro-expressions, which play a key role in applications related to human-computer interaction, mental health assessment, and sentiment analysis. These skills are critical for capturing nuanced characteristics and ensuring effective interaction across diverse real-world scenarios.
\\

\noindent \textbf{Face Localization:}
In this category, we evaluate the model's ability to accurately locate and analyze facial regions through tasks such as head pose estimation, face parsing, and crowd counting. Head pose estimation assesses the model's spatial awareness and its ability to interpret the orientation of faces in three-dimensional space, which is crucial for applications like gaze tracking, augmented reality, and driver monitoring systems. Face parsing focuses on the precise segmentation and labeling of facial regions, such as eyes, nose, and mouth, enabling fine-grained analysis for tasks like virtual makeup, medical diagnosis, and personalized user interfaces. Crowd counting evaluates the model's proficiency in detecting and quantifying multiple faces in dense or cluttered environments, ensuring robust performance in scenarios such as public safety monitoring, event analysis, and resource planning. Together, these tasks test the model’s ability to generalize across varying scales, perspectives, and levels of complexity. This assessment is vital for enhancing real-world applications that rely on accurate face localization and ensuring consistent performance across diverse conditions.
\\

\noindent \textbf{Face Tools Use:}
In this category, we evaluate the model's ability to leverage external tools for face understanding tasks, reflecting a shift from traditional supervised fine-tuning to tool-based problem-solving. Using the FaceXAPI dataset, we assess the model's proficiency in selecting and sequencing the correct APIs and function calls to solve complex tasks. This evaluation emphasizes the model's ability to interpret detailed task requirements, identify relevant tools, and construct accurate operational workflows. The significance of this task lies in its alignment with real-world applications, where equipping MLLMs with tools enhances both scalability and adaptability.

\subsection{Tasks}
\noindent \textbf{Age Estimation:} This task involves determining an individual's age or age range based on their facial features.\\
\noindent \textbf{Gender Prediction:} Gender prediction is the process of identifying a person’s gender from facial images by analyzing visual characteristics of the face.\\
\noindent \textbf{Race Estimation:} Race estimation involves predicting an individual's racial background by analyzing their facial features.\\
\noindent \textbf{High-resolution Face Recognition:} High-resolution face recognition is the identification or verification of individuals using detailed facial images captured at high resolutions, which enhances the accuracy in distinguishing fine facial features.\\
\noindent \textbf{Low-resolution Face Recognition:} Low-resolution Face Recognition refers to performing face recognition tasks on images with limited detail, such as surveillance footage, which poses challenges due to reduced image clarity.\\
\noindent \textbf{Celebrity Identification:} Celebrity identification involves recognizing and naming well-known individuals in images or videos by comparing their facial features against a database of celebrity faces.\\
\noindent \textbf{Face Anti-spoofing:} This task focuses on detecting attempts to deceive facial recognition systems using methods such as photos, videos, or masks, ensuring that the face presented is genuine.\\
\noindent \textbf{Deepfake Detection:} Deepfake detection is the process of identifying synthetic media where a person's likeness has been digitally altered or replaced, aiming to detect manipulated or fabricated content.\\
\noindent \textbf{Attributes Prediction:} Attributes prediction involves inferring various facial characteristics, such as the presence of glasses, facial hair, or specific expressions, from images.\\
\noindent \textbf{Facial Expression Recognition:} This task involves analyzing facial movements to determine a person’s emotional state, such as happiness, sadness, or anger.\\
\noindent \textbf{Headpose Estimation:} Head pose estimation is the process of determining the orientation of a person's head (e.g., pitch, yaw, roll) relative to the camera, and is useful in applications like gaze tracking.\\
\noindent \textbf{Face parsing} This task refers to segmenting a facial image into distinct regions (e.g., eyes, nose, mouth) to facilitate detailed analysis or manipulation of facial components.\\
\noindent \textbf{Crowd Counting:} Crowd counting involves estimating the number of individuals present in an image or video frame, often used in surveillance and event monitoring.\\
\noindent \textbf{Face Tools Retrieval:} It refers to predicting the correct sequence of API calls that MLLMs need to execute to complete complex face-related scenarios requiring multiple tasks. 

\subsection{Source Datasets}
\noindent \textbf{FairFace~\cite{karkkainen2021fairface}:} FairFace is a face image dataset designed to address racial bias in facial recognition systems. It contains 108,501 images with annotations for race, gender, and age. The dataset includes seven race categories: White, Black, East Asian, Southeast Asian, Indian, Middle Eastern, and Latino. The images are sourced primarily from the YFCC-100M Flickr dataset and are balanced across different demographic groups.\\
\noindent \textbf{UTKFace~\cite{zhifei2017cvpr}:} UTKFace is a large-scale face dataset comprising over 20,000 images with annotations for age, gender, and ethnicity. The age range spans from 0 to 116 years. The images exhibit variations in pose, facial expression, illumination, occlusion, and resolution, making it suitable for tasks like face detection, age estimation, and landmark localization.\\
\noindent \textbf{WMCA (Wide Multi-Channel Presentation Attack)~\cite{george2019biometric}:} The WMCA dataset consists of 1,941 short video recordings from 72 different identities, including both bona fide and presentation attacks. Data is recorded across multiple channels: color, depth, infrared, and thermal. The dataset is designed for research in face anti-spoofing and presentation attack detection.\\
\noindent \textbf{MSU MFSD~\cite{wen2015face}:} It contains 280 video recordings of genuine and attack faces from 35 individuals. Each individual has two real-access videos captured with laptop cameras and Android devices. Attack videos include high-definition replays and photo attacks. The dataset is divided into 120 training videos from 15 subjects and 160 testing videos from 20 subjects.\\
\noindent \textbf{CASIA MFSD~\cite{zhang2012face}:} It is a face anti-spoofing dataset containing 600 video recordings from 50 subjects. Each subject has 12 videos under different resolutions and lighting conditions. The dataset includes three spoof attack types: replay, warp print, and cut print attacks. It is divided into 240 training videos from 20 subjects and 360 testing videos from 30 subjects.\\
\noindent \textbf{Replay Attack~\cite{chingovska2012effectiveness}:} The Replay Attack dataset is designed for evaluating face anti-spoofing systems. It includes videos of both genuine access attempts and various spoofing attacks, such as printed photos and video replays. The dataset provides a diverse set of conditions to test the robustness of anti-spoofing algorithms.\\
\noindent \textbf{CelebDF~\cite{li2020celeb}:} CelebDF is a large-scale dataset for deepfake detection, containing 590 real videos of celebrities and 5,639 corresponding deepfake videos. The dataset is designed to evaluate the performance of deepfake detection algorithms under real-world conditions. \\
\noindent \textbf{FF++~\cite{rossler2019faceforensics++}:} FaceForensics++ is a dataset for evaluating facial manipulation detection methods. It consists of over 1,000 original video sequences and more than 4,000 manipulated videos using four different face manipulation techniques. Annotations include manipulation methods and compression levels.\\
\noindent \textbf{TinyFace~\cite{cheng2019low}:} TinyFace is a dataset focused on detecting small faces in images. It contains images with a wide range of face sizes, particularly emphasizing faces that occupy a small number of pixels. Annotations include bounding boxes for each face.\\
\noindent \textbf{LFW~\cite{huang2008labeled}:} LFW is a dataset of 13,000 labeled images of faces from the wild, collected from the internet. It includes annotations for the identity of the person in each image, with 1,680 individuals having two or more images. The dataset is widely used for studying face recognition in unconstrained environments.\\
\noindent \textbf{AgeDB~\cite{moschoglou2017agedb}:} AgeDB is a manually collected dataset containing 16,488 images of 568 distinct subjects. It provides age annotations and is used for evaluating age-invariant face verification and recognition algorithms.\\
\noindent \textbf{CFP-FF and CFP-FP~\cite{sengupta2016frontal}:} The CFP dataset consists of two subsets: CFP-FF (Frontal-Frontal) and CFP-FP (Frontal-Profile). Each contains 7,000 images of 500 subjects. CFP-FF includes frontal face pairs, while CFP-FP includes frontal and profile face pairs, facilitating the study of face recognition across different poses.\\
\noindent \textbf{CALFW~\cite{zheng2017cross}:} CALFW is a dataset derived from LFW, focusing on cross-age face verification. It contains 4,025 image pairs with age differences, aiming to evaluate the performance of face recognition systems under age variation.\\
\noindent \textbf{CPLFW~\cite{zheng2018cross}:} CPLFW is another extension of LFW, emphasizing cross-pose face verification. It includes 3,000 image pairs with pose variations, challenging face recognition models to handle different facial orientations.\\
\noindent \textbf{IMDB~\cite{Rothe-IJCV-2018}:} The IMDB dataset comprises 460,723 face images of 20,284 celebrities, collected from the Internet Movie Database (IMDb). Annotations include age, gender, and name, making it suitable for age estimation and gender classification tasks.\\
\noindent \textbf{CelebA~\cite{liu2015faceattributes}:} CelebA is a large-scale face attributes dataset with more than 200,000 celebrity images, each annotated with 40 attribute labels. The dataset covers a wide range of poses and backgrounds, supporting tasks like attribute prediction and face detection.\\
\noindent \textbf{RAF-DB~\cite{li2017reliable}:} RAF-DB contains 29,672 facial images with annotations for basic and compound emotions. The dataset is used for studying facial expression recognition in real-world scenarios.\\
\noindent \textbf{AffectNet~\cite{mollahosseini2017affectnet}:} AffectNet is a comprehensive facial expression dataset with over 1 million images collected from the internet. Annotations include seven discrete facial expressions (anger, contempt, disgust, fear, happiness, sadness, and surprise) and the intensity of valence and arousal. It is widely used for emotion recognition and affective computing research.\\
\noindent \textbf{AFLW2000~\cite{yin2017towards}:} AFLW2000 contains 2,000 face images annotated with 68 facial landmarks. The images are diverse, covering various poses, expressions, and occlusions. It is often used for face alignment and landmark localization tasks. \\
\noindent \textbf{BIWI~\cite{fanelli2011real}:} The BIWI dataset includes 15,678 images of 20 subjects, captured using a Kinect camera. Annotations consist of 6D head poses (yaw, pitch, roll, and translation vectors) and 3D face models. It is designed for head pose estimation research.\\
\noindent \textbf{JHUCrowd++~\cite{sindagi2020jhu}:} JHUCrowd++ is a large-scale dataset for crowd counting, containing 4,372 images with over 1.5 million annotated heads. The annotations include head locations, crowd density maps, and visibility levels, making it suitable for crowd analysis and density estimation.\\
\noindent \textbf{ShanghaiTech~\cite{zhang2016single}:} The ShanghaiTech dataset contains two parts: Part A, with 482 images captured in crowded scenes, and Part B, with 716 images from less dense environments. It includes over 330,000 annotated individuals' head locations, making it a benchmark for crowd counting and density estimation.\\
\noindent \textbf{CelebAMask-HQ~\cite{lee2020maskgan}:} CelebAMask-HQ is an extension of the CelebA dataset with 30,000 high-resolution face images and fine-grained segmentation masks for 19 facial attributes (e.g., eyes, nose, hair, and skin). It supports tasks like face parsing and image editing.\\
\noindent \textbf{LaPa~\cite{liu2020new}:} LaPa includes 22,000 facial images with high-quality annotations for 11 facial regions. It offers various attributes like pose, expression, and occlusion, making it suitable for face parsing and semantic segmentation tasks.\\
\noindent \textbf{FaceXAPI} It is a dataset consisting of 100 text-only questions, each with four options and one correct answer. It is designed to assess the capabilities of MLLMs (Multimodal Large Language Models) in predicting the correct sequence of API calls needed to accomplish complex scenarios involving multiple face-related tasks.

\subsection{Dataset Statistics}
The FaceXBench dataset is derived from 25 public datasets and one newly created dataset. The number of questions sourced from each dataset, along with the type of questions (multiple images, single images, or text-only), as well as the associated tasks and categories, are summarized in Table~\ref{tab:dataset_stats}.

\begin{table*}[ht]
\centering
\resizebox{\textwidth}{!}{
\begin{tabular}{@{}lcccccc@{}}
\toprule
\textbf{\begin{tabular}[c]{@{}c@{}}Dataset\\ \end{tabular}} & 
\textbf{\begin{tabular}[c]{@{}c@{}}Number of \\ Questions\end{tabular}} & 
\textbf{\begin{tabular}[c]{@{}c@{}}Multiple \\ Images\end{tabular}} & 
\textbf{\begin{tabular}[c]{@{}c@{}}Single \\ Images\end{tabular}} & 
\textbf{\begin{tabular}[c]{@{}c@{}}Text \\ Only\end{tabular}} & 
\textbf{\begin{tabular}[c]{@{}c@{}}Task\\ \end{tabular}} & 
\textbf{\begin{tabular}[c]{@{}c@{}}Category\\ \end{tabular}} \\ 
\midrule
FairFace~\cite{karkkainen2021fairface} & 300 & 200 & 100 & 0 & Age Estimation & Bias \& Fairness \\
UTKFace~\cite{zhifei2017cvpr} & 200 & 150 & 50 & 0 & Age Estimation & Bias \& Fairness \\
FairFace~\cite{karkkainen2021fairface} & 300   & 200  & 100 & 0 & Gender Prediction  & Bias \& Fairness \\
UTKFace~\cite{zhifei2017cvpr} & 200 & 150 & 50  & 0  & Gender Prediction  & Bias \& Fairness \\
FairFace~\cite{karkkainen2021fairface} & 300 & 200 & 100 & 0 & Race Estimation & Bias \& Fairness\\
UTKFace~\cite{zhifei2017cvpr} & 200 & 150 & 50 & 0 & Race Estimation & Bias \& Fairness\\
LFW~\cite{huang2008labeled} & 60 & 60 & 0 & 0 & HR Face Recognition & Face Recognition\\
AgeDB~\cite{moschoglou2017agedb} & 100 & 100 & 0 & 0 & HR Face Recognition & Face Recognition\\
CFP-FF~\cite{sengupta2016frontal} & 60 & 60 & 0 & 0 & HR Face Recognition & Face Recognition\\
CFP-FP~\cite{sengupta2016frontal} & 60 & 60 & 0 & 0 & HR Face Recognition & Face Recognition\\
CALFW~\cite{zheng2017cross} & 60 & 60 & 0 & 0 & HR Face Recognition & Face Recognition\\
CPLFW~\cite{zheng2018cross} & 60 & 60 & 0 & 0 & HR Face Recognition & Face Recognition\\
TinyFace~\cite{cheng2019low} & 100 & 100 & 0 & 0 & LR Face Recognition & Face Recognition\\
IMDB~\cite{Rothe-IJCV-2018} & 300 & 150 & 150 & 0 & Celebrity Identification & Face Recognition\\
WMCA~\cite{george2019biometric} & 250 & 250 & 0 & 0 & Face Anti-spoofing & Face Authentication \\
MSU-MFSD~\cite{wen2015face} & 50 & 50 & 0 & 0 & Face Anti-spoofing & Face Authentication\\
CASIA-MFSD~\cite{zhang2012face} & 50 & 50 & 0 & 0 & Face Anti-spoofing & Face Authentication\\
ReplayAttack~\cite{chingovska2012effectiveness} & 50 & 50 & 0 & 0 & Face Anti-spoofing & Face Authentication\\
CelebDF~\cite{li2020celeb} & 150 & 150 & 0 & 0 & Deepfake Detection & Face Authentication\\
FF++~\cite{rossler2019faceforensics++} & 150 & 150 & 0 & 0 & Deepfake Detection & Face Authentication\\
CelebA~\cite{liu2015faceattributes} & 400 & 200 & 200 & 0 & Attributes Prediction & Face Analysis\\
RAF-DB~\cite{li2017reliable} & 200 & 100 & 100 & 0 & Facial Expression Recognition & Face Analysis\\
AffectNet~\cite{mollahosseini2017affectnet} & 200 & 100 & 100 & 0 & Facial Expression Recognition & Face Analysis\\
AFLW2000~\cite{yin2017towards} & 200 & 50 & 150 & 0 & Headpose Estimation & Face Analysis \\
BIWI~\cite{fanelli2011real} & 200 & 50 & 150 & 0 & Headpose Estimation & Face Analysis\\
JHUCrowd++~\cite{sindagi2020jhu} & 200 & 0 & 200 & 0 & Crowd Counting & Face Localization\\
ShanghaiTech~\cite{zhang2016single} & 100 & 0 & 100 & 0 & Crowd Counting & Face Localization \\
CelebAMask-HQ~\cite{lee2020maskgan} & 200 & 0 & 200 & 0 & Face Parsing & Face Localization\\
LaPa~\cite{liu2020new} & 200 & 0 & 200 & 0 & Face Parsing & Face Localization \\
FaceXAPI & 100 & 0 & 0 & 100 & Face Tools Retrieval & Face Tools Use \\
\bottomrule
\end{tabular}
}
\caption{Question distribution of FaceXBench across datasets}
\label{tab:dataset_stats}
\end{table*}

\subsection{Images used in the dataset}
\begin{figure*}
    \centering
    \resizebox{0.97\textwidth}{!}{
        \includegraphics{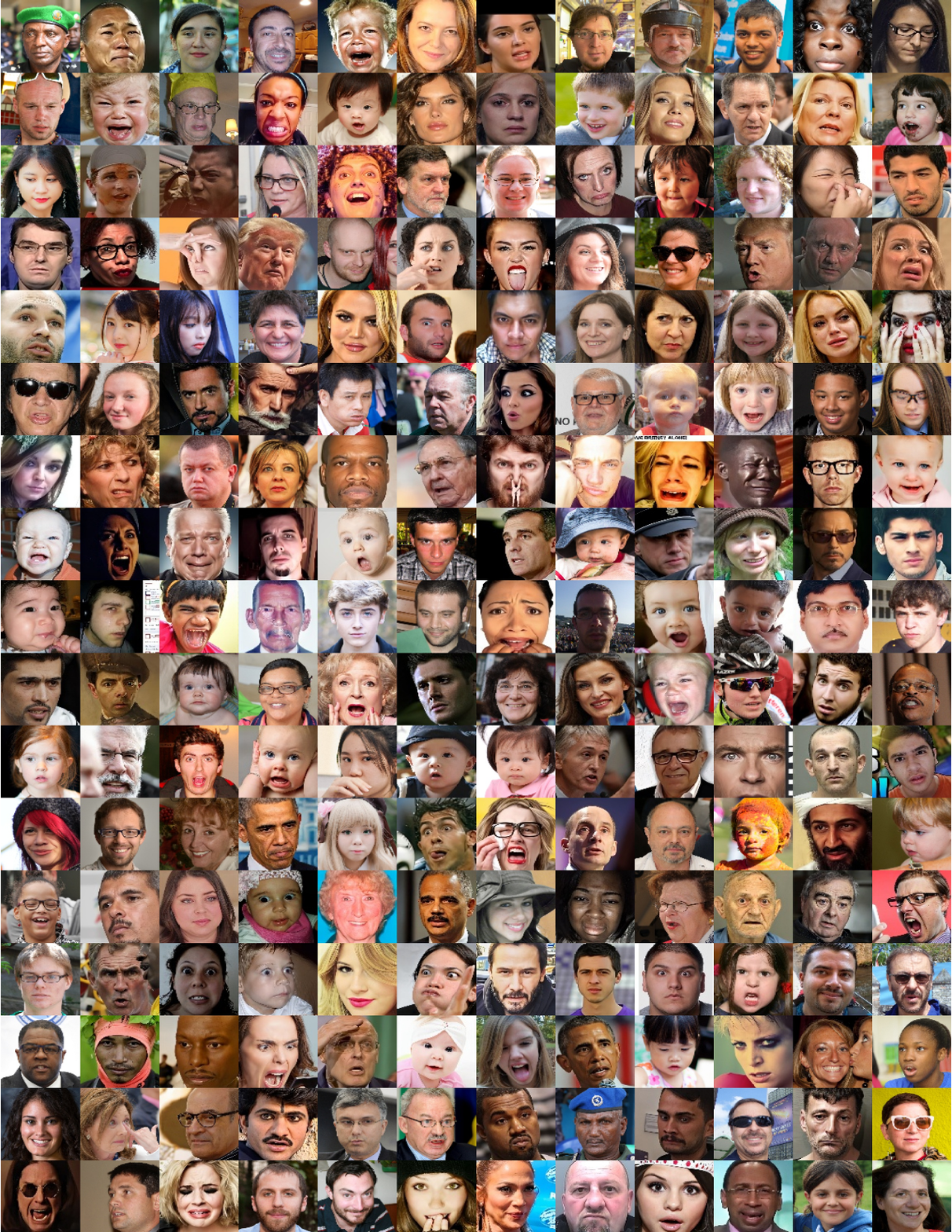}
    }
    \caption{Collage of a subset of images from the dataset, showcasing the diversity of images used in FaceXBench.}
    \label{fig:collage}
\end{figure*}

Figure~\ref{fig:collage} displays a subset of the facial images used in the dataset, highlighting the diversity of faces included in the benchmark. The benchmark consists of face images with varying backgrounds, resolutions, and head pose orientations, as well as a wide range of facial expressions. It includes individuals from different age groups, genders, and racial backgrounds, with each face characterized by a versatile set of attributes.

%% file: appendix/X_dataset.tex
\section{Dataset Collection}
In this section, we provide additional implementation details on the dataset curation process, including information about question templates, distractor options, the evaluation strategy, and the dataset format.

\subsection{Question Templates}
\label{appendix:qs_template}
We provide a selection of question templates for each task used in creating the dataset. The options for the questions may vary, and the examples provided below are intended as samples.
\input{appendix/X_question_templates}

\subsection{Generating Distractor Options}
\label{appendix:distractor_options}
We strategically design the distractor options to encourage MLLMs to carefully choose an option as the prediction. The following methods are employed for different types of tasks:

\begin{itemize}
    \item \textbf{Tasks with numerical answers (crowd counting, age estimation, headpose estimation, number of attributes):} Distractor options are chosen from a numerical range close to the actual ground truth value. For example, if the ground truth crowd count value is 7, we randomly choose distractor options from the range [4, 10], excluding the ground truth value (7).

    \item \textbf{Celebrity identification and facial expression recognition:} Distractor options are selected randomly.

    \item \textbf{Attribute prediction:} Attributes are randomly selected as distractors, ensuring that only one option is correct.

    \item \textbf{Tools retrieval:} The options are generated using GPT and then manually filtered to ensure one correct answer, with the distractor options tweaked to differ by 1 to 2 function calls.

    \item {\textbf{Tasks with options as permutations of image numbers:} Depending on the number of images in the question, four options are randomly selected, ensuring that one of the options is correct. A sample superset of options with three images is shown below: 
        \begin{itemize}
            \item Image 1
            \item Image 2
            \item Image 3
            \item None of the above
            \item Image 1 and Image 2
            \item Image 2 and Image 3
            \item Image 1 and Image 3
            \item Image 1, Image 2, and Image 3
        \end{itemize}}
\end{itemize}

\subsection{FaceXAPI dataset prompt}
\label{appendix:dataset_prompt}
To generate the FaceXAPI questions, we design a detailed prompt incorporating 13 API calls and a total of 32 functions. Additional guidelines are provided to ensure scenario realism, functional complexity, and diversity in the questions. We also include instructions for generating distractor options, emphasizing logical plausibility and ensuring they are sufficiently close yet distinct from the correct answer. A total of 150 questions are generated, out of which 100 are carefully selected to maintain diversity. To further enhance the quality, the options are manually reviewed to ensure the presence of one correct answer alongside logically plausible distractors. The detailed prompt for question generation is provided below.

\clearpage

\begin{tcolorbox}[colback=gray!10, colframe=gray!50, boxrule=0.5mm, rounded corners, title={\textbf{\textcolor{black}{Prompt for generating FaceXAPI questions}}}]
\label{api_prompt}
You are an AI tasked with generating complex, real-world scenario questions to assess a model's ability to select the correct API and function calls to accomplish nuanced tasks. Use the list of APIs and functions provided below.
\begin{enumerate}
    \item \textbf{Age Estimation:}
    \begin{description}
        \item[api\_name:] \texttt{api\_1}
        \small{
        \begin{itemize}
            \item \texttt{predict\_age:}
                \begin{description}
                    \item[Description:] Predicts the age of the person in the input face image.
                    \item[Input:] \texttt{np.ndarray} or \texttt{str} - The input face image.
                    \item[Output:] \texttt{int} - The estimated age.
                \end{description}
            \item \texttt{age\_confidence\_score:}
                \begin{description}
                    \item[Description:] Provides a confidence score for the age estimation.
                    \item[Input:] \texttt{dict} - Output from the age estimation model.
                    \item[Output:] \texttt{float} - Confidence score of the age prediction.
                \end{description}
        \end{itemize}
        }
    \end{description}
    
\item \textbf{Gender Prediction:}
\begin{description}
    \item[api\_name:] \texttt{api\_2}
    \small{
    \begin{itemize}
        \item \texttt{classify\_gender:}
            \begin{description}
                \item[Description:] Classifies the gender of the person in the face image.
                \item[Input:] \texttt{np.ndarray} or \texttt{str} - The input face image.
                \item[Output:] \texttt{str} - The predicted gender ('male' or 'female').
            \end{description}
        \item \texttt{get\_gender\_probabilities:}
            \begin{description}
                \item[Description:] Returns probabilities for each gender class.
                \item[Input:] \texttt{np.ndarray} - The input face image.
                \item[Output:] \texttt{dict} - Probabilities for 'male' and 'female' classes.
            \end{description}
    \end{itemize}
    }
\end{description}

\item \textbf{Race Detection:}
\begin{description}
    \item[api\_name:] \texttt{api\_3}
    \small{
    \begin{itemize}
        \item \texttt{predict\_race:}
            \begin{description}
                \item[Description:] Predicts the race of the individual(s) in the input image(s).
                \item[Input:] \texttt{str}, \texttt{bytes}, \texttt{np.ndarray}, or \texttt{list} - The input image or batch of images.
                \item[Output:] \texttt{list} - Predicted races for each detected face, including race labels and probabilities.
            \end{description}
        \item \texttt{get\_race\_probabilities:}
            \begin{description}
                \item[Description:] Returns probability distribution over races for detected faces.
                \item[Input:] \texttt{str}, \texttt{bytes}, \texttt{np.ndarray}, or \texttt{list} - The input image or batch of images.
                \item[Output:] \texttt{list} - For each detected face, a dictionary of race probabilities.
            \end{description}
        \item \texttt{race\_confidence\_score:}
            \begin{description}
                \item[Description:] Provides a confidence score for race prediction for each detected face.
                \item[Input:] \texttt{str}, \texttt{bytes}, \texttt{np.ndarray}, or \texttt{list} - The input image(s).
                \item[Output:] \texttt{list} - Confidence scores for race predictions.
            \end{description}
    \end{itemize}
    }
\end{description}

\item \textbf{Face Anti-Spoofing:}
\begin{description}
    \item[api\_name:] \texttt{api\_4}
    \small{
    \begin{itemize}
        \item \texttt{detect\_spoofing:}
            \begin{description}
                \item[Description:] Detects if the face in the image is real or a spoof.
                \item[Input:] \texttt{np.ndarray} or \texttt{str} - The input face image.
                \item[Output:] \texttt{bool} - True if spoof detected, False if face is real.
            \end{description}
        \item \texttt{spoof\_confidence\_score:}
            \begin{description}
                \item[Description:] Provides a confidence score indicating the likelihood of spoofing.
                \item[Input:] \texttt{dict} - Output from the anti-spoofing model.
                \item[Output:] \texttt{float} - Spoofing confidence score.
            \end{description}
    \end{itemize}
    }
\end{description}

\end{enumerate}
\end{tcolorbox}

\begin{tcolorbox}[colback=gray!10, colframe=gray!50, boxrule=0.5mm, rounded corners, title={\textbf{\textcolor{black}{Prompt for generating FaceXAPI questions}}}]
\begin{enumerate}
\setcounter{enumi}{4}
    \item \textbf{Deepfake Detection:}
    \begin{description}
        \item[api\_name:] \texttt{api\_5}
        \small{
        \begin{itemize}
            \item \texttt{detect\_deepfake:}
                \begin{description}
                    \item[Description:] Detects whether a video or image contains deepfake content.
                    \item[Input:] \texttt{str} or \texttt{np.ndarray} - Path to video/image or image array.
                    \item[Output:] \texttt{bool} - True if deepfake is detected, False otherwise.
                \end{description}
            \item \texttt{deepfake\_confidence\_score:}
                \begin{description}
                    \item[Description:] Provides a confidence score for deepfake detection.
                    \item[Input:] \texttt{dict} - Output from the deepfake detection model.
                    \item[Output:] \texttt{float} - Confidence score for deepfake presence.
                \end{description}
        \end{itemize}
        }
    \end{description}
\item \textbf{Low-Resolution Face Recognition:}
\begin{description}
    \item[api\_name:] \texttt{api\_6}
    \small{
    \begin{itemize}
        \item \texttt{extract\_low\_res\_embedding:}
            \begin{description}
                \item[Description:] Extracts facial embeddings from low-resolution images for recognition.
                \item[Input:] \texttt{np.ndarray} or \texttt{str} - The input low-res face image.
                \item[Output:] \texttt{np.ndarray} - Embedding vector for low-resolution face.
            \end{description}
        \item \texttt{compare\_low\_res\_embeddings:}
            \begin{description}
                \item[Description:] Compares two low-resolution face embeddings for a match.
                \item[Input:] Two \texttt{np.ndarray} embeddings - The embeddings of two faces.
                \item[Output:] \texttt{bool} - True if faces match, False otherwise.
            \end{description}
        \item \texttt{identify\_low\_res\_face:}
            \begin{description}
                \item[Description:] Identifies a face by comparing a low-resolution embedding to a database.
                \item[Input:] \texttt{np.ndarray} embedding and \texttt{dict} database - Embedding to identify and known embeddings.
                \item[Output:] \texttt{str} or \texttt{None} - Label of the identified face, or None if no match.
            \end{description}
    \end{itemize}
    }
\end{description}

\item \textbf{High-Resolution Face Recognition:}
\begin{description}
    \item[api\_name:] \texttt{api\_7}
    \small{
    \begin{itemize}
        \item \texttt{extract\_high\_res\_embedding:}
            \begin{description}
                \item[Description:] Extracts facial embeddings from high-resolution images for recognition.
                \item[Input:] \texttt{np.ndarray} or \texttt{str} - The input high-res face image.
                \item[Output:] \texttt{np.ndarray} - Embedding vector for high-resolution face.
            \end{description}
        \item \texttt{compare\_high\_res\_embeddings:}
            \begin{description}
                \item[Description:] Compares two high-resolution face embeddings for a match.
                \item[Input:] Two \texttt{np.ndarray} embeddings - The embeddings of two faces.
                \item[Output:] \texttt{bool} - True if faces match, False otherwise.
            \end{description}
        \item \texttt{identify\_high\_res\_face:}
            \begin{description}
                \item[Description:] Identifies a face by comparing a high-resolution embedding to a database.
                \item[Input:] \texttt{np.ndarray} embedding and \texttt{dict} database - Embedding to identify and known embeddings.
                \item[Output:] \texttt{str} or \texttt{None} - Label of the identified face, or None if no match.
            \end{description}
    \end{itemize}
    }
\end{description}

\item \textbf{Celebrity Identification:}
\begin{description}
    \item[api\_name:] \texttt{api\_8}
    \small{
    \begin{itemize}
        \item \texttt{identify\_celebrity:}
            \begin{description}
                \item[Description:] Identifies the celebrity in the input image by matching features against a database.
                \item[Input:] \texttt{str}, \texttt{bytes}, \texttt{np.ndarray}, or \texttt{list} - The input image(s).
                \item[Output:] \texttt{list} - Identified celebrities with names and scores.
            \end{description}
        \item \texttt{celebrity\_confidence\_score:}
            \begin{description}
                \item[Description:] Computes confidence score for the celebrity identification for each detected face.
                \item[Input:] \texttt{str}, \texttt{bytes}, \texttt{np.ndarray}, or \texttt{list} - The input image(s).
                \item[Output:] \texttt{list} - Confidence scores for identified celebrities.
            \end{description}
    \end{itemize}
    }
\end{description}
    
\end{enumerate}
\end{tcolorbox}

\begin{tcolorbox}[colback=gray!10, colframe=gray!50, boxrule=0.5mm, rounded corners, title={\textbf{\textcolor{black}{Prompt for generating FaceXAPI questions}}}]
\begin{enumerate}
\setcounter{enumi}{8}
\item \textbf{Attributes Prediction:}
\begin{description}
    \item[api\_name:] \texttt{api\_9}
    \small{
    \begin{itemize}
        \item \texttt{detect\_attributes:}
            \begin{description}
                \item[Description:] Detects various attributes of the face in the image.
                \item[Input:] \texttt{np.ndarray} or \texttt{str} - The input face image.
                \item[Output:] \texttt{dict} - Detected attributes and their values.
            \end{description}
        \item \texttt{list\_face\_attributes:}
            \begin{description}
                \item[Description:] Lists all possible face attributes that can be predicted.
                \item[Input:] \texttt{None}
                \item[Output:] \texttt{list} - List of attribute names.
            \end{description}
        \item \texttt{attribute\_confidence\_score:}
            \begin{description}
                \item[Description:] Provides confidence scores for each predicted attribute.
                \item[Input:] \texttt{dict} - Output from the attribute detection model.
                \item[Output:] \texttt{dict} - Confidence scores for each attribute.
            \end{description}
    \end{itemize}
    }
\end{description}

\item \textbf{Facial Expression Recognition:}
\begin{description}
    \item[api\_name:] \texttt{api\_10}
    \small{
    \begin{itemize}
        \item \texttt{detect\_expression:}
            \begin{description}
                \item[Description:] Detects facial expressions in the input image.
                \item[Input:] \texttt{np.ndarray} or \texttt{str} - The input face image.
                \item[Output:] \texttt{str} - The detected expression label.
            \end{description}
        \item \texttt{get\_emotion\_probabilities:}
            \begin{description}
                \item[Description:] Provides probabilities for each emotion class.
                \item[Input:] \texttt{np.ndarray} - The input face image.
                \item[Output:] \texttt{dict} - Probabilities for each emotion class.
            \end{description}
        \item \texttt{track\_expression\_over\_time:}
            \begin{description}
                \item[Description:] Tracks facial expressions over a sequence of frames.
                \item[Input:] \texttt{list} of \texttt{np.ndarray} - List of frames from a video.
                \item[Output:] \texttt{list} - Sequence of detected expressions.
            \end{description}
    \end{itemize}
    }
\end{description}

\item \textbf{Headpose Estimation:}
\begin{description}
    \item[api\_name:] \texttt{api\_11}
    \small{
    \begin{itemize}
        \item \texttt{estimate\_head\_pose:}
            \begin{description}
                \item[Description:] Estimates the head pose angles (yaw, pitch, roll) from the face image.
                \item[Input:] \texttt{np.ndarray} or \texttt{str} - The input face image.
                \item[Output:] \texttt{tuple} - Estimated angles (yaw, pitch, roll).
            \end{description}
        \item \texttt{pose\_confidence\_score:}
            \begin{description}
                \item[Description:] Provides a confidence score for the estimated head pose.
                \item[Input:] \texttt{dict} - Output from the head pose estimation model.
                \item[Output:] \texttt{float} - Confidence score for the head pose estimation.
            \end{description}
    \end{itemize}
    }
\end{description}

\item \textbf{Crowd Counting:}
\begin{description}
    \item[api\_name:] \texttt{api\_12}
    \small{
    \begin{itemize}
        \item \texttt{estimate\_crowd\_size:}
            \begin{description}
                \item[Description:] Estimates the size of the crowd based on face detection.
                \item[Input:] \texttt{np.ndarray} - The input image.
                \item[Output:] \texttt{int} - Estimated number of people in the crowd.
            \end{description}
        \item \texttt{aggregate\_counting\_data:}
            \begin{description}
                \item[Description:] Aggregates counting data over multiple images or frames.
                \item[Input:] \texttt{list} of \texttt{np.ndarray} - List of images.
                \item[Output:] \texttt{dict} - Aggregated counting results.
            \end{description}
    \end{itemize}
    }
\end{description}
    
\end{enumerate}
\end{tcolorbox}

\begin{tcolorbox}[colback=gray!10, colframe=gray!50, boxrule=0.5mm, rounded corners, title={\textbf{\textcolor{black}{Prompt for generating FaceXAPI questions}}}]
\begin{enumerate}
\setcounter{enumi}{12}
\item \textbf{Face Segmentation:}
\begin{description}
    \item[api\_name:] \texttt{api\_13}
    \small{
    \begin{itemize}
        \item \texttt{segment\_face\_regions:}
            \begin{description}
                \item[Description:] Segments different regions of the face in the image.
                \item[Input:] \texttt{np.ndarray} - The input face image.
                \item[Output:] \texttt{np.ndarray} - Segmentation mask of the face regions.
            \end{description}
        \item \texttt{classify\_face\_parts:}
            \begin{description}
                \item[Description:] Classifies different parts of the face into categories.
                \item[Input:] \texttt{np.ndarray} - The input face image.
                \item[Output:] \texttt{dict} - Dictionary of face parts and their classifications.
            \end{description}
        \item \texttt{get\_segmentation\_mask\_part:}
            \begin{description}
                \item[Description:] Generates a segmentation mask for a particular face part.
                \item[Input:] \texttt{np.ndarray} - The input face image.
                \item[Output:] \texttt{np.ndarray} - Binary mask of the segmented face part.
            \end{description}
    \end{itemize}
    }
\end{description}

\end{enumerate}

Each scenario should require the model to accurately retrieve and execute 3 to 5 function calls across multiple APIs, simulating the complexity and sequential decision-making needed in real-world applications.

\subsection*{Guidelines for Generating Questions:}
\begin{enumerate}
    \item \textbf{Scenario Realism:} Design questions that reflect realistic application scenarios where multiple APIs must be used in sequence or combined to achieve the correct outcome. Each question should require 3 to 5 function calls.
    \item \textbf{Functional Complexity:} Ensure that each question involves varied functions across multiple APIs without relying on the same set of functions every time. Aim to cover a diverse set of tasks and applications but stay within the list of APIs and functions.
    \item \textbf{Logical Flow:} Each question should suggest a sequence that logically flows with the task requirements. Clarify steps needed for functions that build upon each other to reach the final answer. Think in a step-by-step and logical manner to design the questions.
\end{enumerate}

\subsection*{Guidelines for Generating Options:}
\begin{itemize}
    \item \textbf{Complete API Chains:} Provide four option chains, each specifying a complete sequence of API function calls in the correct order necessary to solve the task. One of these sequences should be the correct answer, while the other three should be close but logically incorrect.
    \item \textbf{Logical Plausibility of Distractors:} Ensure distractors are constructed logically and appear plausible. Avoid overly obvious incorrect answers; the distractors should require reasoning to eliminate, demanding attention to function descriptions, and careful thought to get to the correct answer.
    \item \textbf{Randomized Answer Positioning:} Shuffle the options to ensure the correct answer appears randomly in position A, B, C, or D.
\end{itemize}

We provide a example question and output template, which needs to be strictly followed when providing output.
\subsection*{Example Question:}
In an airport security system, faces are checked for deepfakes, head poses are verified, and age is estimated. Age is analyzed only if head pose confidence is high. Which API sequence should be applied?
\begin{enumerate}[label=\textbf{\Alph*.}]
{\footnotesize
    \item \texttt{api\_5-detect\_deepfake}, \texttt{api\_1-predict\_age}, \texttt{api\_11-pose\_confidence\_score}, \texttt{api\_11-estimate\_head\_pose}
    \item \texttt{api\_11-estimate\_head\_pose}, \texttt{api\_11-pose\_confidence\_score}, \texttt{api\_5-detect\_deepfake}, \texttt{api\_1-predict\_age}
    \item \texttt{api\_5-detect\_deepfake}, \texttt{api\_11-estimate\_head\_pose}, \texttt{api\_11-pose\_confidence\_score}, \texttt{api\_1-predict\_age}
    \item \texttt{api\_11-pose\_confidence\_score}, \texttt{api\_5-detect\_deepfake}, \texttt{api\_11-estimate\_head\_pose}, \texttt{api\_1-predict\_age}}
\end{enumerate}
{\footnotesize
\textbf{Correct Answer:} \textbf{C.} \texttt{api\_5-detect\_deepfake}, \texttt{api\_11-estimate\_head\_pose}, \texttt{api\_11-pose\_confidence\_score}, \texttt{api\_1-predict\_age}}\\

\normalsize{Provide 150 questions. Remember to follow the JSON structure and guidelines strictly. Ensure each question and answer chain is unique and fits the described requirements.}

\end{tcolorbox}

\newpage

\subsection{Option Extraction from Prediction}
\label{sec:appendix_evaluation_function}
\begin{center}
\hrule
\vspace{0.5em}
\captionof*{listing}{\normalsize\textbf{Function used to extract option label from the predicted model output}}
\begin{small}
\begin{lstlisting}[mathescape, mathescape]
import re

def extract_option_label(model_output, option_labels, options):
    # Remove any leading/trailing whitespace
    model_output_clean = model_output.strip()
    
    # Define regex patterns to match option labels
    label_patterns = [
        r'\b\(?([A-Za-z])\)?\b',         # Matches 'A', '(A)', etc.
        r'\bOption\s+([A-Za-z])\b',      # Matches 'Option A', etc.
        r'\b([A-Za-z])\.',               # Matches 'A.', etc.
        r'\b([A-Za-z]):',                # Matches 'A:', etc.
        r'\b([A-Za-z])\s+-',             # Matches 'A -', etc.
        r'\b\(([A-Za-z])\)',             # Matches '(A)', etc.
    ]
    
    model_output_start = model_output_clean
    for pattern in label_patterns:
        match = re.match(pattern, model_output_start)
        if match:
            label = match.group(1)
            if label in option_labels:
                return label
    
    # If not found at the beginning, search the entire text
    for pattern in label_patterns:
        matches = re.findall(pattern, model_output_clean)
        for label in matches:
            if label in option_labels:
                return label

    # If no label found, try to match the output to the option texts exactly
    for idx, option_text in enumerate(options):
        option_text_stripped = str(option_text).strip()
        if option_text_stripped == model_output_clean.strip():
            return option_labels[idx]
        elif option_text_stripped in model_output_clean:
            return option_labels[idx]
    
    return None
\end{lstlisting}
\end{small}
\vspace{0.5em}
\hrule
\end{center}

We design \texttt{extract\_option\_label} funtion to robustly extract an option label from a given \texttt{model\_output}, utilizing a three-step fallback mechanism to ensure accurate identification. The first step involves cleaning the \texttt{model\_output} by removing leading and trailing whitespace. A series of regular expression (regex) patterns are then applied to match common formats of option labels, such as \texttt{A}, \texttt{(A)}, \texttt{A)}, \texttt{Option A}, \texttt{A:}, \texttt{A-}, specifically at the start of the text. If a match is found, the label is validated against the provided \texttt{option\_labels} list and returned if valid. If no valid label is identified at the start, the second fallback step searches the entire \texttt{model\_output} for matches to these regex patterns and again validates the extracted labels against \texttt{option\_labels}, allowing for flexibility in label positioning. If both regex-based approaches fail, the third and final fallback step compares the \texttt{model\_output} directly with the exact text of the provided \texttt{options}. It checks whether the entire \texttt{model\_output} or any substring within it matches an option text and maps it to the corresponding label. If no label is found after these steps, the function returns \texttt{None}. This layered process ensures minimal false positives by progressively narrowing down matches using robust regex and exact comparisons.

\subsection{Dataset Format}
\label{sec:appendix_dataset}

The benchmark consists of multiple JSON files, with each dataset represented in a JSON file named in the format \texttt{<dataset\_name>\_<multiple/single>.json}. These files include detailed metadata and questions for evaluation, enabling easy analysis and comprehensive testing. The JSON dataset format is designed to enable the evaluation of MLLMs on specific tasks or subsets of tasks. A sample structure of the JSON file is provided below.

{\footnotesize
\begin{verbatim}
{
    "category": "attributes_expression",
    "sub-category": "expression",
    "dataset": "affectnet",
    "question_type": "MCQ",
    "num_images": "multiple",
    "prepend_text": "Facial expression recognition involves identifying the emotional state of a 
        person based on their facial expressions in an image. You have to perform the task of 
        facial expression recognition.",
    "postpend_text": "Please answer the question and provide only the correct option 
        letter, e.g., A, B, C, D.",
    "questions": {
        "1": {
            "image_paths": [
                "image_1.png",
                "image_2.jpg"
            ],
            "question_text": "Which images have a person with the expression 'anger'?",
            "options": [
                "Image 2",
                "All of the images",
                "None of the images",
                "Image 1"
            ],
            "correct_answer_option": "D",
            "answer": "Image 1"
        },
        "2": {
            "image_paths": [
                "image_1.png",
                "image_2.png"
            ],
            "question_text": "Which images have a person with the expression 'contempt'?",
            "options": [
                "All of the images",
                "None of the images",
                "Image 2",
                "Image 1"
            ],
            "correct_answer_option": "D",
            "answer": "Image 1"
        }
        ...
    }
}
\end{verbatim}
}

\noindent The components in the JSON file are as follows:

\begin{itemize}
    \item \textbf{category:} Represents the primary category of the task, e.g., \texttt{attributes\_expression}.
    \item \textbf{sub-category:} Specifies the task within the main category, e.g., \texttt{expression}.
    \item \textbf{dataset:} Indicates the name of the dataset used for the task, e.g., \texttt{affectnet}.
    \item \textbf{question\_type:} Defines the type of question, e.g., \texttt{MCQ} (Multiple Choice Question).
    \item \textbf{num\_images:} Specifies whether the question involves \texttt{multiple} or \texttt{single} images.
    \item \textbf{prepend\_text:} A task description that is prepended to the question in the in-context task description evaluation setting.
    \item \textbf{postpend\_text:} Text appended to the question to encourage standardized answer formats (e.g., answering with option letters such as A, B, C, D).
    \item \textbf{questions:} A dictionary containing individual questions, with the following components:
    \begin{itemize}
        \item \textbf{image\_paths:} A list of paths to the images required for the question. Each image is named sequentially (e.g., \texttt{image\_1}, \texttt{image\_2}, etc.).
        \item \textbf{question\_text:} The text of the question.
        \item \textbf{options:} A list of possible answer choices.
        \item \textbf{correct\_answer\_option:} The correct answer in the format of an option letter (e.g., \texttt{A}, \texttt{B}, etc.).
        \item \textbf{answer:} The correct answer in textual format.
    \end{itemize}
\end{itemize}

\noindent This JSON format allows the community to evaluate MLLMs on a subset of categories or tasks, providing flexibility and consistency in evaluation. The inclusion of both \texttt{correct\_answer\_option} and \texttt{answer} supports diverse evaluation strategies.

\subsection{Other Evaluation Settings}
\label{sec:appendix_evaluation}
In the \texttt{\textbf{in-context evaluation}} setting, we prepend the task-description before the question to provide relevant context to the model. The prepend text for each task is as follows:

\begin{itemize}
    \item \textbf{Age Estimation:} Age estimation involves predicting the age of a person based on their facial features in an image. For this task, the following age groups are defined: 0--2, 3--9, 10--19, 20--29, 30--39, 40--49, 50--59, 60--69, and more than 70 years.

    \item \textbf{Gender Prediction:} Gender prediction involves determining the gender of a person based on their facial features in an image.

    \item \textbf{Race Estimation:} Race prediction involves determining the race or ethnicity of a person based on their facial features in an image. The possible race groups are White, Black, East Asian, Southeast Asian, Middle Eastern, Indian, Latino/Hispanic, and Other.

    \item \textbf{High-resolution Face Recognition:} High-resolution face recognition involves identifying a person based on their facial features in images of good quality or resolution.

    \item \textbf{Low-resolution Face Recognition:} Low-resolution face recognition involves identifying a person based on their facial features in images with reduced quality or resolution.

    \item \textbf{Celebrity Identification:} Celebrity identification involves recognizing and determining the identity of a celebrity based on their facial features in an image.

    \item \textbf{Face Anti-spoofing:} Face anti-spoofing involves detecting whether a presented face is real or a spoof attempt using methods such as photos, videos, or masks. Common attack types include `Glasses', `Print', `Replay', `Fake Head', `Rigid Mask', `Flexible Mask', `Wig', and `Paper Mask'.

    \item \textbf{Deepfake Detection:} Deepfake detection involves identifying manipulated images where faces are swapped or altered and are considered fake.

    \item \textbf{Attributes Prediction:} Face attributes recognition involves identifying specific characteristics of a person, such as beard, moustache, and facial features, based on their facial appearance in an image.

    \item \textbf{Facial Expression Recognition:} Facial expression recognition involves identifying the emotional state of a person based on their facial expressions in an image.

    \item \textbf{Headpose Estimation:} Headpose estimation involves predicting the orientation of a person's head in an image by estimating the yaw, pitch, and roll angles. For this task, the Euler angles are divided into the following bins: 
    \begin{itemize}
        \item $-100$ to $-90$, $-90$ to $-80$, $-80$ to $-70$, $-70$ to $-60$, $-60$ to $-50$, $-50$ to $-40$, $-40$ to $-30$, $-30$ to $-20$, $-20$ to $-10$, $-10$ to $0$, $0$ to $10$, $10$ to $20$, $20$ to $30$, $30$ to $40$, $40$ to $50$, $50$ to $60$, $60$ to $70$, $70$ to $80$, $80$ to $90$, and $90$ to $100$.
    \end{itemize}

    \item \textbf{Face Parsing:} Face parsing involves dividing a facial image into distinct regions such as eyes, nose, mouth, and other facial features.

    \item \textbf{Crowd Counting:} Crowd counting involves estimating the number of people present in an image, typically in a crowded scene.

    \item \textbf{Tools Retrieval:} This task involves selecting the correct sequence of API function calls from a set of multiple-choice options based on a given question. Each question describes a specific use case scenario that demands a tailored sequence of API functions to achieve desired results, such as identity verification, emotion detection, or demographic analysis.
\end{itemize}

We append each question with the instruction, `\textit{Please answer the question and provide only the correct option letter, e.g., A, B, C, D.}' to ensure standardized outputs, making it easier to extract the predicted option. We append the same instruction for zero-shot inference as well.

For the \texttt{\textbf{chain-of-thought evaluation}} setting, we append each question with the instruction, `\textit{Please think step by step and provide your reasoning before giving the final answer. Include the final correct answer option at the end of your answer.}', to encourage the model to reason before making its final prediction.

%% file: appendix/X_question_templates.tex
\begin{tcolorbox}[colback=gray!10, colframe=gray!50, boxrule=0.5mm, rounded corners, title={\textbf{\textcolor{black}{Age Estimation \textbar{} Multiple Images}}}]
\textbf{Which image shows a person in \texttt{\{age\_group\}} age group?}
\begin{enumerate}[label=\textbf{\Alph*.}]
    \item Image 1
    \item Image 2
    \item None of the above
    \item Both Images
\end{enumerate}
\end{tcolorbox}

\begin{tcolorbox}[colback=gray!10, colframe=gray!50, boxrule=0.5mm, rounded corners, title={\textbf{\textcolor{black}{Age Estimation \textbar{} Multiple Images}}}]
\textbf{Who among these two appears to be older?}
\begin{enumerate}[label=\textbf{\Alph*.}]
    \item Image 1
    \item Image 2
    \item Both are of the same age group
    \item None of the above
\end{enumerate}
\end{tcolorbox}

\begin{tcolorbox}[colback=gray!10, colframe=gray!50, boxrule=0.5mm, rounded corners, title={\textbf{\textcolor{black}{Age Estimation \textbar{} Multiple Images}}}]
\textbf{Arrange the following images in ascending order of age.}
\begin{enumerate}[label=\textbf{\Alph*.}]
    \item Image 1, Image 2, Image 3
    \item Image 2, Image 1, Image 3
    \item Image 3, Image 2, Image 1
    \item Image 3, Image 1, Image 2
\end{enumerate}
\end{tcolorbox}

\begin{tcolorbox}[colback=gray!10, colframe=gray!50, boxrule=0.5mm, rounded corners, title={\textbf{\textcolor{black}{Age Estimation \textbar{} Multiple Images}}}]
\textbf{Arrange the following images in descending order of age.}
\begin{enumerate}[label=\textbf{\Alph*.}]
    \item Image 2, Image 3, Image 1
    \item Image 2, Image 1, Image 3
    \item Image 3, Image 2, Image 1
    \item Image 1, Image 2, Image 3
\end{enumerate}
\end{tcolorbox}

\begin{tcolorbox}[colback=gray!10, colframe=gray!50, boxrule=0.5mm, rounded corners, title={\textbf{\textcolor{black}{Age Estimation \textbar{} Multiple Images}}}]
\textbf{How many images have a person of age between \texttt{\{age\_start\}} and \texttt{\{age\_end\}}?}
\begin{enumerate}[label=\textbf{\Alph*.}]
    \item 0
    \item 1
    \item 2
    \item 3
\end{enumerate}
\end{tcolorbox}

\begin{tcolorbox}[colback=gray!10, colframe=gray!50, boxrule=0.5mm, rounded corners, title={\textbf{\textcolor{black}{Age Estimation \textbar{} Multiple Images}}}]
\textbf{Among these images, which image shows the oldest person?}
\begin{enumerate}[label=\textbf{\Alph*.}]
    \item Image 1
    \item Image 2
    \item Image 3
    \item All are of same age group
\end{enumerate}
\end{tcolorbox}

\begin{tcolorbox}[colback=gray!10, colframe=gray!50, boxrule=0.5mm, rounded corners, title={\textbf{\textcolor{black}{Age Estimation \textbar{} Single Image}}}] \textbf{What is the most appropriate age group for the person in this image?}
\begin{enumerate}[label=\textbf{\Alph*.}] 
    \item \textless \textit{correct\_option} \textgreater
    \item \textless \textit{distractor\_option} \textgreater
    \item \textless \textit{distractor\_option} \textgreater
    \item \textless \textit{distractor\_option} \textgreater
\end{enumerate} \end{tcolorbox}

\begin{tcolorbox}[colback=gray!10, colframe=gray!50, boxrule=0.5mm, rounded corners, title={\textbf{\textcolor{black}{Age Estimation \textbar{} Single Image}}}] \textbf{Approximately how old is the person in this image?}
\begin{enumerate}[label=\textbf{\Alph*.}] 
    \item \textless \textit{correct\_option} \textgreater
    \item \textless \textit{distractor\_option} \textgreater
    \item \textless \textit{distractor\_option} \textgreater
    \item \textless \textit{distractor\_option} \textgreater
\end{enumerate} \end{tcolorbox}

\begin{tcolorbox}[colback=gray!10, colframe=gray!50, boxrule=0.5mm, rounded corners, title={\textbf{\textcolor{black}{Age Estimation \textbar{} Single Image}}}] \textbf{Select the age group that best describes the person in this image.}
\begin{enumerate}[label=\textbf{\Alph*.}] 
    \item \textless \textit{correct\_option} \textgreater
    \item \textless \textit{distractor\_option} \textgreater
    \item \textless \textit{distractor\_option} \textgreater
    \item \textless \textit{distractor\_option} \textgreater
\end{enumerate} \end{tcolorbox}

\begin{tcolorbox}[colback=gray!10, colframe=gray!50, boxrule=0.5mm, rounded corners, title={\textbf{\textcolor{black}{Age Estimation \textbar{} Single Image}}}] \textbf{Estimate the age group of the person in this image.}
\begin{enumerate}[label=\textbf{\Alph*.}] 
    \item \textless \textit{correct\_option} \textgreater
    \item \textless \textit{distractor\_option} \textgreater
    \item \textless \textit{distractor\_option} \textgreater
    \item \textless \textit{distractor\_option} \textgreater
\end{enumerate} \end{tcolorbox}

\begin{tcolorbox}[colback=gray!10, colframe=gray!50, boxrule=0.5mm, rounded corners, title={\textbf{\textcolor{black}{Gender Prediction \textbar{} Multiple Images}}}]
\textbf{Which image shows a male person?}
\begin{enumerate}[label=\textbf{\Alph*.}]
    \item Image 1
    \item Image 2
    \item None of the above
    \item Both Images
\end{enumerate}
\end{tcolorbox}

\begin{tcolorbox}[colback=gray!10, colframe=gray!50, boxrule=0.5mm, rounded corners, title={\textbf{\textcolor{black}{Gender Prediction \textbar{} Multiple Images}}}]
\textbf{Which image shows a female person?}
\begin{enumerate}[label=\textbf{\Alph*.}]
    \item Image 1
    \item Image 2
    \item None of the above
    \item Both Images
\end{enumerate}
\end{tcolorbox}

\begin{tcolorbox}[colback=gray!10, colframe=gray!50, boxrule=0.5mm, rounded corners, title={\textbf{\textcolor{black}{Gender Prediction \textbar{} Multiple Images}}}]
\textbf{Identify which image shows a person whose gender is male.}
\begin{enumerate}[label=\textbf{\Alph*.}]
    \item Image 1
    \item Image 2
    \item Both Images
    \item None of the above
\end{enumerate}
\end{tcolorbox}

\begin{tcolorbox}[colback=gray!10, colframe=gray!50, boxrule=0.5mm, rounded corners, title={\textbf{\textcolor{black}{Gender Prediction \textbar{} Multiple Images}}}]
\textbf{Which images appear to have a person with male gender?}
\begin{enumerate}[label=\textbf{\Alph*.}]
    \item Image 1, Image 2
    \item Image 3, Image 1
    \item Image 1
    \item None of the above
\end{enumerate}
\end{tcolorbox}

\begin{tcolorbox}[colback=gray!10, colframe=gray!50, boxrule=0.5mm, rounded corners, title={\textbf{\textcolor{black}{Gender Prediction \textbar{} Multiple Images}}}]
\textbf{How many images show female individuals?}
\begin{enumerate}[label=\textbf{\Alph*.}]
    \item 0
    \item 1
    \item 2
    \item 3
\end{enumerate}
\end{tcolorbox}

\begin{tcolorbox}[colback=gray!10, colframe=gray!50, boxrule=0.5mm, rounded corners, title={\textbf{\textcolor{black}{Gender Prediction \textbar{} Single Image}}}]
\textbf{What is the gender of the person in this image?}
\begin{enumerate}[label=\textbf{\Alph*.}]
    \item Male
    \item Female
    \item Both
    \item None of the above
\end{enumerate}
\end{tcolorbox}

\begin{tcolorbox}[colback=gray!10, colframe=gray!50, boxrule=0.5mm, rounded corners, title={\textbf{\textcolor{black}{Gender Prediction \textbar{} Single Image}}}]
\textbf{Identify the gender of the person in this image.}
\begin{enumerate}[label=\textbf{\Alph*.}]
    \item Male
    \item Female
    \item Both
    \item None of the above
\end{enumerate}
\end{tcolorbox}

\begin{tcolorbox}[colback=gray!10, colframe=gray!50, boxrule=0.5mm, rounded corners, title={\textbf{\textcolor{black}{Gender Prediction \textbar{} Single Image}}}]
\textbf{Which gender category best describes the person in this image?}
\begin{enumerate}[label=\textbf{\Alph*.}]
    \item Male
    \item Female
    \item Both
    \item None of the above
\end{enumerate}
\end{tcolorbox}

\begin{tcolorbox}[colback=gray!10, colframe=gray!50, boxrule=0.5mm, rounded corners, title={\textbf{\textcolor{black}{Gender Prediction \textbar{} Single Image}}}]
\textbf{Select the most appropriate gender for the person in this image.}
\begin{enumerate}[label=\textbf{\Alph*.}]
    \item Male
    \item Female
    \item Both
    \item None of the above
\end{enumerate}
\end{tcolorbox}

\begin{tcolorbox}[colback=gray!10, colframe=gray!50, boxrule=0.5mm, rounded corners, title={\textbf{\textcolor{black}{Gender Prediction \textbar{} Single Image}}}]
\textbf{Determine the gender of the person shown in this image.}
\begin{enumerate}[label=\textbf{\Alph*.}]
    \item Male
    \item Female
    \item Both
    \item None of the above
\end{enumerate}
\end{tcolorbox}

\begin{tcolorbox}[colback=gray!10, colframe=gray!50, boxrule=0.5mm, rounded corners, title={\textbf{\textcolor{black}{Race Estimation \textbar{} Multiple Images}}}]
\textbf{Between the two images, who seems more likely to belong to the Black race?}
\begin{enumerate}[label=\textbf{\Alph*.}]
    \item Image 1
    \item Image 2
    \item Both Images
    \item None of the above
\end{enumerate}
\end{tcolorbox}

\begin{tcolorbox}[colback=gray!10, colframe=gray!50, boxrule=0.5mm, rounded corners, title={\textbf{\textcolor{black}{Race Estimation \textbar{} Multiple Images}}}]
\textbf{Which individual appears to be of Latino Hispanic origin?}
\begin{enumerate}[label=\textbf{\Alph*.}]
    \item Image 1
    \item Image 2
    \item Both Images
    \item None of the above
\end{enumerate}
\end{tcolorbox}

\begin{tcolorbox}[colback=gray!10, colframe=gray!50, boxrule=0.5mm, rounded corners, title={\textbf{\textcolor{black}{Race Estimation \textbar{} Multiple Images}}}]
\textbf{Which images depict individuals of Indian origin?}
\begin{enumerate}[label=\textbf{\Alph*.}]
    \item Image 1, Image 2
    \item Image 3
    \item None of the above
    \item Image 2, Image 3
\end{enumerate}
\end{tcolorbox}

\begin{tcolorbox}[colback=gray!10, colframe=gray!50, boxrule=0.5mm, rounded corners, title={\textbf{\textcolor{black}{Race Estimation \textbar{} Multiple Images}}}]
\textbf{Which images appear to have individuals of Southeast Asian descent?}

\begin{enumerate}[label=\textbf{\Alph*.}]
    \item Image 1, Image 2
    \item Image 1
    \item None of the above
    \item Image 2, Image 3
\end{enumerate}
\end{tcolorbox}

\begin{tcolorbox}[colback=gray!10, colframe=gray!50, boxrule=0.5mm, rounded corners, title={\textbf{\textcolor{black}{Race Estimation \textbar{} Multiple Images}}}]
\textbf{How many images show people of Black race?}
\begin{enumerate}[label=\textbf{\Alph*.}]
    \item 0
    \item 1
    \item 2
    \item 3
\end{enumerate}
\end{tcolorbox}

\begin{tcolorbox}[colback=gray!10, colframe=gray!50, boxrule=0.5mm, rounded corners, title={\textbf{\textcolor{black}{Race Estimation \textbar{} Single Image}}}]
\textbf{What is the race of the person in this image?}
\begin{enumerate}[label=\textbf{\Alph*.}]
    \item White
    \item East Asian
    \item Latino Hispanic
    \item Other
\end{enumerate}
\end{tcolorbox}

\begin{tcolorbox}[colback=gray!10, colframe=gray!50, boxrule=0.5mm, rounded corners, title={\textbf{\textcolor{black}{Race Estimation \textbar{} Single Image}}}]
\textbf{Identify the race of the person in this image.}
\begin{enumerate}[label=\textbf{\Alph*.}]
    \item Black
    \item Indian
    \item Southeast Asian
    \item Middle Eastern
\end{enumerate}
\end{tcolorbox}

\begin{tcolorbox}[colback=gray!10, colframe=gray!50, boxrule=0.5mm, rounded corners, title={\textbf{\textcolor{black}{Race Estimation \textbar{} Single Image}}}]
\textbf{Which race category best describes the person in this image?}
\begin{enumerate}[label=\textbf{\Alph*.}]
    \item White
    \item Black
    \item Indian
    \item Southeast Asian
\end{enumerate}
\end{tcolorbox}

\begin{tcolorbox}[colback=gray!10, colframe=gray!50, boxrule=0.5mm, rounded corners, title={\textbf{\textcolor{black}{Race Estimation \textbar{} Single Image}}}]
\textbf{Select the most appropriate race for the person in this image.}
\begin{enumerate}[label=\textbf{\Alph*.}]
    \item Latino Hispanic
    \item East Asian
    \item Black
    \item Other
\end{enumerate}
\end{tcolorbox}

\begin{tcolorbox}[colback=gray!10, colframe=gray!50, boxrule=0.5mm, rounded corners, title={\textbf{\textcolor{black}{Race Estimation \textbar{} Single Image}}}]
\textbf{Determine the race of the person shown in this image.}
\begin{enumerate}[label=\textbf{\Alph*.}]
    \item Middle Eastern
    \item Indian
    \item White
    \item Southeast Asian
\end{enumerate}
\end{tcolorbox}

\begin{tcolorbox}[colback=gray!10, colframe=gray!50, boxrule=0.5mm, rounded corners, title={\textbf{\textcolor{black}{High-Resolution Face Recognition/Low-Resolution Face Recognition \textbar{} Multiple Images}}}]
\textbf{The first image is of a person A. The same person A is present in which of the other images?}
\begin{enumerate}[label=\textbf{\Alph*.}]
    \item Image 2
    \item Image 3
    \item Image 4
    \item None of the above
\end{enumerate}
\end{tcolorbox}

\begin{tcolorbox}[colback=gray!10, colframe=gray!50, boxrule=0.5mm, rounded corners, title={\textbf{\textcolor{black}{High-Resolution Face Recognition/Low-Resolution Face Recognition \textbar{} Multiple Images}}}]
\textbf{The first image is of a person A. The same person A is present in how many of the remaining images?}
\begin{enumerate}[label=\textbf{\Alph*.}]
    \item 1
    \item 2
    \item 3
    \item 4
\end{enumerate}
\end{tcolorbox}

\begin{tcolorbox}[colback=gray!10, colframe=gray!50, boxrule=0.5mm, rounded corners, title={\textbf{\textcolor{black}{High-Resolution Face Recognition/Low-Resolution Face Recognition \textbar{} Multiple Images}}}]
\textbf{How many unique identities are present in these images?}
\begin{enumerate}[label=\textbf{\Alph*.}]
    \item 1
    \item 2
    \item 0
    \item 4
\end{enumerate}
\end{tcolorbox}

\begin{tcolorbox}[colback=gray!10, colframe=gray!50, boxrule=0.5mm, rounded corners, title={\textbf{\textcolor{black}{High-Resolution Face Recognition/Low-Resolution Face Recognition \textbar{} Multiple Images}}}]
\textbf{The first image is of a person A. Which of the other images show person A?}
\begin{enumerate}[label=\textbf{\Alph*.}]
    \item Image 2, Image 3
    \item Image 2, Image 4
    \item Image 3, Image 4
    \item None of the above
\end{enumerate}
\end{tcolorbox}

\begin{tcolorbox}[colback=gray!10, colframe=gray!50, boxrule=0.5mm, rounded corners, title={\textbf{\textcolor{black}{High-Resolution Face Recognition/Low-Resolution Face Recognition \textbar{} Multiple Images}}}]
\textbf{Which two images show the same person?}
\begin{enumerate}[label=\textbf{\Alph*.}]
    \item Image 1, Image 2
    \item Image 3, Image 4
    \item Image 2, Image 5
    \item None of the above
\end{enumerate}
\end{tcolorbox}

\begin{tcolorbox}[colback=gray!10, colframe=gray!50, boxrule=0.5mm, rounded corners, title={\textbf{\textcolor{black}{High-Resolution Face Recognition/Low-Resolution Face Recognition \textbar{} Multiple Images}}}]
\textbf{The first image is of person A. Which images are not person A?}
\begin{enumerate}[label=\textbf{\Alph*.}]
    \item Image 2
    \item Image 3
    \item Image 1
    \item None of the above
\end{enumerate}
\end{tcolorbox}

\begin{tcolorbox}[colback=gray!10, colframe=gray!50, boxrule=0.5mm, rounded corners, title={\textbf{\textcolor{black}{High-Resolution Face Recognition/Low-Resolution Face Recognition \textbar{} Multiple Images}}}]
\textbf{There are two images of the same person. Which images are of people different from that person?}
\begin{enumerate}[label=\textbf{\Alph*.}]
    \item Image 1, Image 2
    \item Image 2, Image 3
    \item Image 3, Image 4
    \item None of the above
\end{enumerate}
\end{tcolorbox}

\begin{tcolorbox}[colback=gray!10, colframe=gray!50, boxrule=0.5mm, rounded corners, title={\textbf{\textcolor{black}{High-Resolution Face Recognition/Low-Resolution Face Recognition \textbar{} Multiple Images}}}]
\textbf{How many pairs of images show the same person?}
\begin{enumerate}[label=\textbf{\Alph*.}]
    \item 1
    \item 2
    \item 3
    \item 0
\end{enumerate}
\end{tcolorbox}

\begin{tcolorbox}[colback=gray!10, colframe=gray!50, boxrule=0.5mm, rounded corners, title={\textbf{\textcolor{black}{Celebrity Identification \textbar{} Multiple Images}}}]
\textbf{Which images is of the celebrity \texttt{\{celebrity\_name\}}?}
\begin{enumerate}[label=\textbf{\Alph*.}]
    \item Image 1
    \item Image 2
    \item Image 3
    \item Image 4
\end{enumerate}
\end{tcolorbox}

\begin{tcolorbox}[colback=gray!10, colframe=gray!50, boxrule=0.5mm, rounded corners, title={\textbf{\textcolor{black}{Celebrity Identification \textbar{} Multiple Images}}}]
\textbf{How many images have the celebrity \texttt{\{celebrity\_name\}}?}
\begin{enumerate}[label=\textbf{\Alph*.}]
    \item 1
    \item 2
    \item 3
    \item 0
\end{enumerate}
\end{tcolorbox}

\begin{tcolorbox}[colback=gray!10, colframe=gray!50, boxrule=0.5mm, rounded corners, title={\textbf{\textcolor{black}{Celebrity Identification \textbar{} Multiple Images}}}]
\textbf{Which images have the celebrity \texttt{\{celebrity\_name\}}?}

\begin{enumerate}[label=\textbf{\Alph*.}]
    \item Image 1, Image 2
    \item Image 3, Image 2
    \item Image 1
    \item None of the above
\end{enumerate}
\end{tcolorbox}

\begin{tcolorbox}[colback=gray!10, colframe=gray!50, boxrule=0.5mm, rounded corners, title={\textbf{\textcolor{black}{Celebrity Identification \textbar{} Multiple Images}}}]
\textbf{Which images do not have the celebrity \texttt{\{celebrity\_name\}}?}
\begin{enumerate}[label=\textbf{\Alph*.}]
    \item Image 1, Image 2
    \item Image 3
    \item Image 1
    \item None of the above
\end{enumerate}
\end{tcolorbox}

\begin{tcolorbox}[colback=gray!10, colframe=gray!50, boxrule=0.5mm, rounded corners, title={\textbf{\textcolor{black}{Celebrity Identification \textbar{} Multiple Images}}}]
\textbf{Which pair of images share the same celebrity?}
\begin{enumerate}[label=\textbf{\Alph*.}]
    \item Image 1, Image 2
    \item Image 3, Image 4
    \item Image 2, Image 3
    \item None of the above
\end{enumerate}
\end{tcolorbox}

\begin{tcolorbox}[colback=gray!10, colframe=gray!50, boxrule=0.5mm, rounded corners, title={\textbf{\textcolor{black}{Celebrity Identification \textbar{} Multiple Images}}}]
\textbf{How many unique celebrities are present in these images?}
\begin{enumerate}[label=\textbf{\Alph*.}]
    \item 1
    \item 2
    \item 3
    \item 0
\end{enumerate}
\end{tcolorbox}

\begin{tcolorbox}[colback=gray!10, colframe=gray!50, boxrule=0.5mm, rounded corners, title={\textbf{\textcolor{black}{Celebrity Identification \textbar{} Multiple Images}}}]
\textbf{What is the name of the most frequently occurring celebrity in these images?}

\begin{enumerate}[label=\textbf{\Alph*.}]
    \item \textless \textit{celebrity\_A} \textgreater
    \item \textless \textit{celebrity\_B} \textgreater
    \item \textless \textit{celebrity\_C} \textgreater
    \item \textless \textit{celebrity\_D} \textgreater
\end{enumerate}
\end{tcolorbox}

\begin{tcolorbox}[colback=gray!10, colframe=gray!50, boxrule=0.5mm, rounded corners, title={\textbf{\textcolor{black}{Celebrity Identification \textbar{} Single Image}}}]
\textbf{What is the name of this celebrity?}
\begin{enumerate}[label=\textbf{\Alph*.}]
    \item \textless \textit{celebrity\_A} \textgreater
    \item \textless \textit{celebrity\_B} \textgreater
    \item \textless \textit{celebrity\_C} \textgreater
    \item \textless \textit{celebrity\_D} \textgreater
\end{enumerate}
\end{tcolorbox}

\begin{tcolorbox}[colback=gray!10, colframe=gray!50, boxrule=0.5mm, rounded corners, title={\textbf{\textcolor{black}{Celebrity Identification \textbar{} Single Image}}}]
\textbf{Who is this person?}
\begin{enumerate}[label=\textbf{\Alph*.}]
    \item \textless \textit{celebrity\_A} \textgreater
    \item \textless \textit{celebrity\_B} \textgreater
    \item \textless \textit{celebrity\_C} \textgreater
    \item \textless \textit{celebrity\_D} \textgreater
\end{enumerate}
\end{tcolorbox}

\begin{tcolorbox}[colback=gray!10, colframe=gray!50, boxrule=0.5mm, rounded corners, title={\textbf{\textcolor{black}{Celebrity Identification \textbar{} Single Image}}}]
\textbf{Name the celebrity shown in the image.}
\begin{enumerate}[label=\textbf{\Alph*.}]
    \item \textless \textit{celebrity\_A} \textgreater
    \item \textless \textit{celebrity\_B} \textgreater
    \item \textless \textit{celebrity\_C} \textgreater
    \item \textless \textit{celebrity\_D} \textgreater
\end{enumerate}
\end{tcolorbox}

\begin{tcolorbox}[colback=gray!10, colframe=gray!50, boxrule=0.5mm, rounded corners, title={\textbf{\textcolor{black}{Face Anti-spoofing \textbar{} Multiple Images}}}]
\textbf{Which image shows a bonafide person?}
\begin{enumerate}[label=\textbf{\Alph*.}]
    \item Image 1
    \item Image 2
    \item Both Images
    \item None of the above
\end{enumerate}
\end{tcolorbox}

\begin{tcolorbox}[colback=gray!10, colframe=gray!50, boxrule=0.5mm, rounded corners, title={\textbf{\textcolor{black}{Face Anti-spoofing \textbar{} Multiple Images}}}]
\textbf{Which image is spoof attacked?}
\begin{enumerate}[label=\textbf{\Alph*.}]
    \item Image 1
    \item Image 2
    \item Both Images
    \item None of the above
\end{enumerate}
\end{tcolorbox}

\begin{tcolorbox}[colback=gray!10, colframe=gray!50, boxrule=0.5mm, rounded corners, title={\textbf{\textcolor{black}{Face Anti-spoofing \textbar{} Multiple Images}}}]
\textbf{Which statement best describes the images?}
\begin{enumerate}[label=\textbf{\Alph*.}]
    \item Both images are bonafide
    \item Both images are attacks
    \item Image 1 is bonafide and Image 2 is attack
    \item Image 1 is attack and Image 2 is bonafide
\end{enumerate}
\end{tcolorbox}

\begin{tcolorbox}[colback=gray!10, colframe=gray!50, boxrule=0.5mm, rounded corners, title={\textbf{\textcolor{black}{Face Anti-spoofing \textbar{} Multiple Images}}}]
\textbf{How many images are bonafide?}
\begin{enumerate}[label=\textbf{\Alph*.}]
    \item 0
    \item 1
    \item 2
    \item 3
\end{enumerate}
\end{tcolorbox}

\begin{tcolorbox}[colback=gray!10, colframe=gray!50, boxrule=0.5mm, rounded corners, title={\textbf{\textcolor{black}{Face Anti-spoofing \textbar{} Multiple Images}}}]
\textbf{How many images are spoof attacked?}
\begin{enumerate}[label=\textbf{\Alph*.}]
    \item 0
    \item 1
    \item 2
    \item 3
\end{enumerate}
\end{tcolorbox}

\begin{tcolorbox}[colback=gray!10, colframe=gray!50, boxrule=0.5mm, rounded corners, title={\textbf{\textcolor{black}{Face Anti-spoofing \textbar{} Multiple Images}}}]
\textbf{Which images are not bonafide?}
\begin{enumerate}[label=\textbf{\Alph*.}]
    \item Image 1 and Image 2
    \item Image 2 and Image 3
    \item Image 3
    \item None of the above
\end{enumerate}
\end{tcolorbox}

\begin{tcolorbox}[colback=gray!10, colframe=gray!50, boxrule=0.5mm, rounded corners, title={\textbf{\textcolor{black}{Face Anti-spoofing \textbar{} Single Image}}}]
\textbf{Is this image a bonafide person or it's a spoof attack?}
\begin{enumerate}[label=\textbf{\Alph*.}]
    \item Bonafide
    \item Attack
\end{enumerate}
\end{tcolorbox}

\begin{tcolorbox}[colback=gray!10, colframe=gray!50, boxrule=0.5mm, rounded corners, title={\textbf{\textcolor{black}{Face Anti-spoofing \textbar{} Single Image}}}]
\textbf{Is this image an attack of type \texttt{\{attack\_type\}}?}
\begin{enumerate}[label=\textbf{\Alph*.}]
    \item Yes
    \item No
\end{enumerate}
\end{tcolorbox}

\begin{tcolorbox}[colback=gray!10, colframe=gray!50, boxrule=0.5mm, rounded corners, title={\textbf{\textcolor{black}{Face Anti-spoofing \textbar{} Single Image}}}]
\textbf{Is this image an example of a bonafide person?}
\begin{enumerate}[label=\textbf{\Alph*.}]
    \item Yes
    \item No
\end{enumerate}
\end{tcolorbox}

\begin{tcolorbox}[colback=gray!10, colframe=gray!50, boxrule=0.5mm, rounded corners, title={\textbf{\textcolor{black}{Face Anti-spoofing \textbar{} Single Image}}}]
\textbf{Is this image depicting a spoof attack?}
\begin{enumerate}[label=\textbf{\Alph*.}]
    \item Yes
    \item No
\end{enumerate}
\end{tcolorbox}

\begin{tcolorbox}[colback=gray!10, colframe=gray!50, boxrule=0.5mm, rounded corners, title={\textbf{\textcolor{black}{Deepfake Detection \textbar{} Multiple Images}}}]
\textbf{Which deepfakes belong to the same identity?}
\begin{enumerate}[label=\textbf{\Alph*.}]
    \item Images 1 and Image 2
    \item Images 1 and Image 3
    \item Images 2 and Image 4
    \item None of the above
\end{enumerate}
\end{tcolorbox}

\begin{tcolorbox}[colback=gray!10, colframe=gray!50, boxrule=0.5mm, rounded corners, title={\textbf{\textcolor{black}{Deepfake Detection \textbar{} Multiple Images}}}]
\textbf{How many deepfakes belong to the same identity?}
\begin{enumerate}[label=\textbf{\Alph*.}]
    \item 0
    \item 1
    \item 2
    \item 3
\end{enumerate}
\end{tcolorbox}

\begin{tcolorbox}[colback=gray!10, colframe=gray!50, boxrule=0.5mm, rounded corners, title={\textbf{\textcolor{black}{Deepfake Detection \textbar{} Multiple Images}}}]
\textbf{Which among these images are fake?}
\begin{enumerate}[label=\textbf{\Alph*.}]
    \item Image 1 and Image 2
    \item Image 2 and Image 3
    \item Image 1
    \item None of the above
\end{enumerate}
\end{tcolorbox}

\begin{tcolorbox}[colback=gray!10, colframe=gray!50, boxrule=0.5mm, rounded corners, title={\textbf{\textcolor{black}{Deepfake Detection \textbar{} Multiple Images}}}]
\textbf{How many images are real?}
\begin{enumerate}[label=\textbf{\Alph*.}]
    \item 0
    \item 1
    \item 2
    \item 3
\end{enumerate}
\end{tcolorbox}

\begin{tcolorbox}[colback=gray!10, colframe=gray!50, boxrule=0.5mm, rounded corners, title={\textbf{\textcolor{black}{Deepfake Detection \textbar{} Multiple Images}}}]
\textbf{Which images are deepfakes?}
\begin{enumerate}[label=\textbf{\Alph*.}]
    \item Image 1 and Image 2
    \item Image 2 and Image 3
    \item Image 1, Image 2, and Image 3
    \item None of the above
\end{enumerate}
\end{tcolorbox}

\begin{tcolorbox}[colback=gray!10, colframe=gray!50, boxrule=0.5mm, rounded corners, title={\textbf{\textcolor{black}{Deepfake Detection \textbar{} Multiple Images}}}]
\textbf{How many images are deepfakes?}
\begin{enumerate}[label=\textbf{\Alph*.}]
    \item 0
    \item 1
    \item 2
    \item 3
\end{enumerate}
\end{tcolorbox}

\begin{tcolorbox}[colback=gray!10, colframe=gray!50, boxrule=0.5mm, rounded corners, title={\textbf{\textcolor{black}{Deepfake Detection \textbar{} Multiple Images}}}]
\textbf{Which image is real?}
\begin{enumerate}[label=\textbf{\Alph*.}]
    \item Image 1
    \item Image 2
    \item Both
    \item None
\end{enumerate}
\end{tcolorbox}

\begin{tcolorbox}[colback=gray!10, colframe=gray!50, boxrule=0.5mm, rounded corners, title={\textbf{\textcolor{black}{Deepfake Detection \textbar{} Multiple Images}}}]
\textbf{Are these images real or fake?}
\begin{enumerate}[label=\textbf{\Alph*.}]
    \item Both are real
    \item Both are fake
    \item Image 1 is real, Image 2 is fake
    \item Image 1 is fake, Image 2 is real
\end{enumerate}
\end{tcolorbox}

\begin{tcolorbox}[colback=gray!10, colframe=gray!50, boxrule=0.5mm, rounded corners, title={\textbf{\textcolor{black}{Deepfake Detection \textbar{} Multiple Images}}}]
\textbf{Which images are real and which are fake?}
\begin{enumerate}[label=\textbf{\Alph*.}]
    \item Images 1 and 2 are real, Image 3 is fake
    \item Images 2 and 3 are real, Image 1 is fake
    \item Images 1 and 3 are real, Image 2 is fake
    \item All images are fake
\end{enumerate}
\end{tcolorbox}

\begin{tcolorbox}[colback=gray!10, colframe=gray!50, boxrule=0.5mm, rounded corners, title={\textbf{\textcolor{black}{Deepfake Detection \textbar{} Multiple Images}}}]
\textbf{Which images are not deepfakes?}
\begin{enumerate}[label=\textbf{\Alph*.}]
    \item Image 1 and Image 2
    \item Image 2 and Image 3
    \item Image 1
    \item None of the above
\end{enumerate}
\end{tcolorbox}

\begin{tcolorbox}[colback=gray!10, colframe=gray!50, boxrule=0.5mm, rounded corners, title={\textbf{\textcolor{black}{Attribute Prediction \textbar{} Single Image}}}]
\textbf{Identify which of the following attributes is present in the image.}
\begin{enumerate}[label=\textbf{\Alph*.}]
    \item 5o Clock Shadow
    \item Blond Hair
    \item Eyeglasses
    \item None of the above
\end{enumerate}
\end{tcolorbox}

\begin{tcolorbox}[colback=gray!10, colframe=gray!50, boxrule=0.5mm, rounded corners, title={\textbf{\textcolor{black}{Attribute Prediction \textbar{} Single Image}}}]
\textbf{Which of the following attributes is NOT present in the image?}
\begin{enumerate}[label=\textbf{\Alph*.}]
    \item Heavy Makeup
    \item Bald
    \item Black Hair
    \item None of the above
\end{enumerate}
\end{tcolorbox}

\begin{tcolorbox}[colback=gray!10, colframe=gray!50, boxrule=0.5mm, rounded corners, title={\textbf{\textcolor{black}{Attribute Prediction \textbar{} Single Image}}}]
\textbf{Which attribute is the most prominent in the image?}
\begin{enumerate}[label=\textbf{\Alph*.}]
    \item Rosy Cheeks
    \item Smiling
    \item Big Nose
    \item None of the above
\end{enumerate}
\end{tcolorbox}

\begin{tcolorbox}[colback=gray!10, colframe=gray!50, boxrule=0.5mm, rounded corners, title={\textbf{\textcolor{black}{Attribute Prediction \textbar{} Multiple Images}}}]
\textbf{Which of the following attributes do both images have in common?}
\begin{enumerate}[label=\textbf{\Alph*.}]
    \item Bald
    \item Bushy Eyebrows
    \item Smiling
    \item None of the above
\end{enumerate}
\end{tcolorbox}

\begin{tcolorbox}[colback=gray!10, colframe=gray!50, boxrule=0.5mm, rounded corners, title={\textbf{\textcolor{black}{Attribute Prediction \textbar{} Multiple Images}}}]
\textbf{How many images have a person with the attribute 'Smiling'?}
\begin{enumerate}[label=\textbf{\Alph*.}]
    \item 1
    \item 2
    \item 3
    \item 4
\end{enumerate}
\end{tcolorbox}

\begin{tcolorbox}[colback=gray!10, colframe=gray!50, boxrule=0.5mm, rounded corners, title={\textbf{\textcolor{black}{Attribute Prediction \textbar{} Multiple Images}}}]
\textbf{Which images have a person with the attribute \texttt{\{attribute\_type\}}?}

\begin{enumerate}[label=\textbf{\Alph*.}]
    \item Image 1 and Image 2
    \item Image 2 and Image 3
    \item Image 1
    \item None of the above
\end{enumerate}
\end{tcolorbox}

\begin{tcolorbox}[colback=gray!10, colframe=gray!50, boxrule=0.5mm, rounded corners, title={\textbf{\textcolor{black}{Attribute Prediction \textbar{} Multiple Images}}}]
\textbf{Which of the following attributes are common to all images?}
\begin{enumerate}[label=\textbf{\Alph*.}]
    \item Bald
    \item Bushy Eyebrows
    \item Black Hair
    \item None of the above
\end{enumerate}
\end{tcolorbox}

\begin{tcolorbox}[colback=gray!10, colframe=gray!50, boxrule=0.5mm, rounded corners, title={\textbf{\textcolor{black}{Attribute Prediction \textbar{} Multiple Images}}}]
\textbf{Which attribute is unique to one of the images?}
\begin{enumerate}[label=\textbf{\Alph*.}]
    \item Bald
    \item Smiling
    \item Black Hair
    \item None of the above
\end{enumerate}
\end{tcolorbox}

\begin{tcolorbox}[colback=gray!10, colframe=gray!50, boxrule=0.5mm, rounded corners, title={\textbf{\textcolor{black}{Attribute Prediction \textbar{} Multiple Images}}}]
\textbf{Which pair of images shares the most attributes?}
\begin{enumerate}[label=\textbf{\Alph*.}]
    \item Image 1 and Image 2
    \item Image 2 and Image 3
    \item Image 1 and Image 3
    \item None of the above
\end{enumerate}
\end{tcolorbox}

\begin{tcolorbox}[colback=gray!10, colframe=gray!50, boxrule=0.5mm, rounded corners, title={\textbf{\textcolor{black}{Attribute Prediction \textbar{} Multiple Images}}}]
\textbf{Which attribute is most frequent among the images?}
\begin{enumerate}[label=\textbf{\Alph*.}]
    \item Bald
    \item Smiling
    \item Black Hair
    \item None of the above
\end{enumerate}
\end{tcolorbox}

\begin{tcolorbox}[colback=gray!10, colframe=gray!50, boxrule=0.5mm, rounded corners, title={\textbf{\textcolor{black}{Facial  Expression Recognition \textbar{} Multiple Images}}}]
\textbf{How many images have a person with the expression \texttt{\{expression\_type\}}?}
\begin{enumerate}[label=\textbf{\Alph*.}]
    \item 1
    \item 2
    \item 3
    \item None of the above
\end{enumerate}
\end{tcolorbox}

\begin{tcolorbox}[colback=gray!10, colframe=gray!50, boxrule=0.5mm, rounded corners, title={\textbf{\textcolor{black}{Facial  Expression Recognition \textbar{} Multiple Images}}}]
\textbf{Which images have a person with the expression \texttt{\{expression\_type\}}?}
\begin{enumerate}[label=\textbf{\Alph*.}]
    \item Image 1
    \item Image 2
    \item Both Images
    \item None of the above
\end{enumerate}
\end{tcolorbox}

\begin{tcolorbox}[colback=gray!10, colframe=gray!50, boxrule=0.5mm, rounded corners, title={\textbf{\textcolor{black}{Facial  Expression Recognition \textbar{} Multiple Images}}}]
\textbf{Which images do not have a person with the expression \texttt{\{expression\_type\}}?}
\begin{enumerate}[label=\textbf{\Alph*.}]
    \item Image 1 and Image 2
    \item Image 3 and Image 4
    \item Image 1
    \item None of the above
\end{enumerate}
\end{tcolorbox}

\begin{tcolorbox}[colback=gray!10, colframe=gray!50, boxrule=0.5mm, rounded corners, title={\textbf{\textcolor{black}{Facial  Expression Recognition \textbar{} Multiple Images}}}]
\textbf{Which pair of images share the same expression?}
\begin{enumerate}[label=\textbf{\Alph*.}]
    \item Image 1 and Image 2
    \item Image 2 and Image 3
    \item Image 1 and Image 3
    \item None of the above
\end{enumerate}
\end{tcolorbox}

\begin{tcolorbox}[colback=gray!10, colframe=gray!50, boxrule=0.5mm, rounded corners, title={\textbf{\textcolor{black}{Facial  Expression Recognition \textbar{} Multiple Images}}}]
\textbf{Which image has a person with the expression \texttt{\{expression\_type\}}?}
\begin{enumerate}[label=\textbf{\Alph*.}]
    \item Image 1
    \item Image 2
    \item Both Images
    \item None of the above
\end{enumerate}
\end{tcolorbox}

\begin{tcolorbox}[colback=gray!10, colframe=gray!50, boxrule=0.5mm, rounded corners, title={\textbf{\textcolor{black}{Facial Expression Recognition \textbar{} Single Image}}}]
\textbf{Which expression is the person showing in the image?}
\begin{enumerate}[label=\textbf{\Alph*.}]
    \item Surprise
    \item Happy
    \item Sad
    \item Neutral
\end{enumerate}
\end{tcolorbox}

\begin{tcolorbox}[colback=gray!10, colframe=gray!50, boxrule=0.5mm, rounded corners, title={\textbf{\textcolor{black}{Facial Expression Recognition \textbar{} Single Image}}}]
\textbf{Identify the primary expression shown by the person in the image.}
\begin{enumerate}[label=\textbf{\Alph*.}]
    \item Fear
    \item Disgust
    \item Anger
    \item Neutral
\end{enumerate}
\end{tcolorbox}

\begin{tcolorbox}[colback=gray!10, colframe=gray!50, boxrule=0.5mm, rounded corners, title={\textbf{\textcolor{black}{Facial Expression Recognition \textbar{} Single Image}}}]
\textbf{What is the main expression of the person in the image?}
\begin{enumerate}[label=\textbf{\Alph*.}]
    \item Happy
    \item Sad
    \item Surprise
    \item Anger
\end{enumerate}
\end{tcolorbox}

\begin{tcolorbox}[colback=gray!10, colframe=gray!50, boxrule=0.5mm, rounded corners, title={\textbf{\textcolor{black}{Headpose Estimation \textbar{} Multiple Images}}}]
\textbf{How many images have a person with the yaw angle of headpose orientation in range \texttt{\{yaw\_range\}}?}
\begin{enumerate}[label=\textbf{\Alph*.}]
    \item 1
    \item 2
    \item 3
    \item None of the above
\end{enumerate}
\end{tcolorbox}

\begin{tcolorbox}[colback=gray!10, colframe=gray!50, boxrule=0.5mm, rounded corners, title={\textbf{\textcolor{black}{Headpose Estimation \textbar{} Multiple Images}}}]
\textbf{Which images have a person with the pitch angle of headpose orientation in range \texttt{\{pitch\_range\}}?}
\begin{enumerate}[label=\textbf{\Alph*.}]
    \item Image 1 and Image 2
    \item Image 3 and Image 4
    \item Image 1
    \item None of the above
\end{enumerate}
\end{tcolorbox}

\begin{tcolorbox}[colback=gray!10, colframe=gray!50, boxrule=0.5mm, rounded corners, title={\textbf{\textcolor{black}{Headpose Estimation \textbar{} Multiple Images}}}]
\textbf{Which images have a person with the roll angle of headpose orientation in range \texttt{\{roll\_range\}}?}
\begin{enumerate}[label=\textbf{\Alph*.}]
    \item Image 2
    \item Image 3
    \item Image 4
    \item None of the above
\end{enumerate}
\end{tcolorbox}

\begin{tcolorbox}[colback=gray!10, colframe=gray!50, boxrule=0.5mm, rounded corners, title={\textbf{\textcolor{black}{Headpose Estimation \textbar{} Multiple Images}}}]
\textbf{Which pair of images have a person that share the same yaw angle bin index of headpose orientation, given that the total yaw angle range is from -100 to 100 degrees, with bins of 10 degrees each?}
\begin{enumerate}[label=\textbf{\Alph*.}]
    \item Image 1 and Image 2
    \item Image 2 and Image 3
    \item Image 3 and Image 4
    \item None of the above
\end{enumerate}
\end{tcolorbox}

\begin{tcolorbox}[colback=gray!10, colframe=gray!50, boxrule=0.5mm, rounded corners, title={\textbf{\textcolor{black}{Headpose Estimation \textbar{} Multiple Images}}}]
\textbf{Which images have a person with the yaw, pitch, and roll angles of headpose orientation between \texttt{\{yaw\_range\}}, \texttt{\{pitch\_range\}}, and \texttt{\{roll\_range\}} degrees respectively?}
\begin{enumerate}[label=\textbf{\Alph*.}]
    \item Image 1
    \item Image 2
    \item Image 3
    \item None of the above
\end{enumerate}
\end{tcolorbox}

\begin{tcolorbox}[colback=gray!10, colframe=gray!50, boxrule=0.5mm, rounded corners, title={\textbf{\textcolor{black}{Headpose Estimation \textbar{} Multiple Images}}}]
\textbf{Which images have a person with the yaw and pitch angles of headpose orientation between \texttt{\{yaw\_range\}} and \texttt{\{pitch\_range\}} degrees respectively?}
\begin{enumerate}[label=\textbf{\Alph*.}]
    \item Image 1 and Image 2
    \item Image 3 and Image 4
    \item Image 2
    \item None of the above
\end{enumerate}
\end{tcolorbox}

\begin{tcolorbox}[colback=gray!10, colframe=gray!50, boxrule=0.5mm, rounded corners, title={\textbf{\textcolor{black}{Headpose Estimation \textbar{} Multiple Images}}}]
\textbf{Which images have a person with the pitch and roll angles of headpose orientation between \texttt{\{pitch\_range\}} and \texttt{\{roll\_range\}} degrees respectively?}

\begin{enumerate}[label=\textbf{\Alph*.}]
    \item Image 1
    \item Image 2 and Image 3
    \item Image 4
    \item None of the above
\end{enumerate}
\end{tcolorbox}

\begin{tcolorbox}[colback=gray!10, colframe=gray!50, boxrule=0.5mm, rounded corners, title={\textbf{\textcolor{black}{Headpose Estimation \textbar{} Single Image}}}]
\textbf{What is the yaw angle range of headpose orientation for the person in this image?}

\begin{enumerate}[label=\textbf{\Alph*.}]
    \item -30 to -20 degrees
    \item -10 to 10 degrees
    \item 20 to 30 degrees
    \item None of the above
\end{enumerate}
\end{tcolorbox}

\begin{tcolorbox}[colback=gray!10, colframe=gray!50, boxrule=0.5mm, rounded corners, title={\textbf{\textcolor{black}{Headpose Estimation \textbar{} Single Image}}}]
\textbf{What is the pitch angle range of headpose orientation for the person in this image?}

\begin{enumerate}[label=\textbf{\Alph*.}]
    \item -15 to -5 degrees
    \item 0 to 10 degrees
    \item 15 to 25 degrees
    \item None of the above
\end{enumerate}
\end{tcolorbox}

\begin{tcolorbox}[colback=gray!10, colframe=gray!50, boxrule=0.5mm, rounded corners, title={\textbf{\textcolor{black}{Headpose Estimation \textbar{} Single Image}}}]
\textbf{What is the roll angle range of headpose orientation for the person in this image?}

\begin{enumerate}[label=\textbf{\Alph*.}]
    \item -25 to -15 degrees
    \item -5 to 5 degrees
    \item 10 to 20 degrees
    \item None of the above
\end{enumerate}
\end{tcolorbox}

\begin{tcolorbox}[colback=gray!10, colframe=gray!50, boxrule=0.5mm, rounded corners, title={\textbf{\textcolor{black}{Crowd Counting \textbar{} Single Image}}}]
\textbf{How many people are present in this image?}

\begin{enumerate}[label=\textbf{\Alph*.}]
    \item \textless \textit{correct\_option} \textgreater
    \item \textless \textit{distractor\_option} \textgreater
    \item \textless \textit{distractor\_option} \textgreater
    \item \textless \textit{distractor\_option} \textgreater
\end{enumerate}
\end{tcolorbox}

\begin{tcolorbox}[colback=gray!10, colframe=gray!50, boxrule=0.5mm, rounded corners, title={\textbf{\textcolor{black}{Crowd Counting \textbar{} Single Image}}}]
\textbf{What is the number of individuals shown in this image?}

\begin{enumerate}[label=\textbf{\Alph*.}]
    \item \textless \textit{correct\_option} \textgreater
    \item \textless \textit{distractor\_option} \textgreater
    \item \textless \textit{distractor\_option} \textgreater
    \item \textless \textit{distractor\_option} \textgreater
\end{enumerate}
\end{tcolorbox}

\begin{tcolorbox}[colback=gray!10, colframe=gray!50, boxrule=0.5mm, rounded corners, title={\textbf{\textcolor{black}{Crowd Counting \textbar{} Single Image}}}]
\textbf{Determine the number of people in this picture.}

\begin{enumerate}[label=\textbf{\Alph*.}]
    \item \textless \textit{correct\_option} \textgreater
    \item \textless \textit{distractor\_option} \textgreater
    \item \textless \textit{distractor\_option} \textgreater
    \item \textless \textit{distractor\_option} \textgreater
\end{enumerate}
\end{tcolorbox}

\begin{tcolorbox}[colback=gray!10, colframe=gray!50, boxrule=0.5mm, rounded corners, title={\textbf{\textcolor{black}{Crowd Counting \textbar{} Single Image}}}]
\textbf{Estimate the count of people present in this image.}

\begin{enumerate}[label=\textbf{\Alph*.}]
    \item \textless \textit{correct\_option} \textgreater
    \item \textless \textit{distractor\_option} \textgreater
    \item \textless \textit{distractor\_option} \textgreater
    \item \textless \textit{distractor\_option} \textgreater
\end{enumerate}
\end{tcolorbox}

\begin{tcolorbox}[colback=gray!10, colframe=gray!50, boxrule=0.5mm, rounded corners, title={\textbf{\textcolor{black}{Crowd Counting \textbar{} Single Image}}}]
\textbf{Please identify the number of individuals in this image.}

\begin{enumerate}[label=\textbf{\Alph*.}]
    \item \textless \textit{correct\_option} \textgreater
    \item \textless \textit{distractor\_option} \textgreater
    \item \textless \textit{distractor\_option} \textgreater
    \item \textless \textit{distractor\_option} \textgreater
\end{enumerate}
\end{tcolorbox}

\begin{tcolorbox}[colback=gray!10, colframe=gray!50, boxrule=0.5mm, rounded corners, title={\textbf{\textcolor{black}{Face Parsing \textbar{} Single Image}}}]
\textbf{Which of the following regions is not present or is segmented out with white color?}
\begin{enumerate}[label=\textbf{\Alph*.}]
    \item left eyebrow
    \item glasses
    \item right eye
    \item left ear
\end{enumerate}
\end{tcolorbox}

\begin{tcolorbox}[colback=gray!10, colframe=gray!50, boxrule=0.5mm, rounded corners, title={\textbf{\textcolor{black}{Face Parsing \textbar{} Single Image}}}]
\textbf{Which region is segmented out with white color?}
\begin{enumerate}[label=\textbf{\Alph*.}]
    \item None of the above
    \item inner mouth
    \item lower lip
    \item upper lip
\end{enumerate}
\end{tcolorbox}

\begin{tcolorbox}[colback=gray!10, colframe=gray!50, boxrule=0.5mm, rounded corners, title={\textbf{\textcolor{black}{Tools Retrieval \textbar{} Text Only}}}]
\textbf{A video conferencing platform needs to ensure user engagement by tracking head pose, verifying expressions over time, and confirming that the detected face is real and not a spoof. Expression tracking should continue only if head pose confidence is high. What is the correct sequence of API calls?}

\begin{enumerate}[label=\textbf{\Alph*.}]
{\small
    \item api\_4-detect\_spoofing, api\_11-estimate\_head\_pose, api\_11-pose\_confidence\_score, api\_10-track\_expression\_over\_time
    \item api\_11-estimate\_head\_pose, api\_11-pose\_confidence\_score, api\_10-track\_expression\_over\_time, api\_4-detect\_spoofing
    \item api\_11-estimate\_head\_pose, api\_11-pose\_confidence\_score, api\_10-track\_expression\_over\_time, api\_4-detect\_spoofing
    \item api\_4-detect\_spoofing, api\_10-track\_expression\_over\_time, api\_11-estimate\_head\_pose, api\_11-pose\_confidence\_score
    }
\end{enumerate}
\end{tcolorbox}

\begin{tcolorbox}[colback=gray!10, colframe=gray!50, boxrule=0.5mm, rounded corners, title={\textbf{\textcolor{black}{Tools Retrieval \textbar{} Text Only}}}]
\textbf{An AR app segments users' faces into regions, detects head pose, and applies filters based on gender and expressions. Expression analysis is performed only if gender is classified with high confidence. Which sequence of API calls is appropriate?}

\begin{enumerate}[label=\textbf{\Alph*.}]
{\small
    \item api\_13-segment\_face\_regions, api\_10-detect\_expression, api\_11-estimate\_head\_pose, api\_2-classify\_gender, api\_2-get\_gender\_probabilities
    \item api\_2-classify\_gender, api\_10-track\_expression\_over\_time, api\_11-estimate\_head\_pose, api\_2-get\_gender\_probabilities
    \item api\_13-segment\_face\_regions, api\_11-estimate\_head\_pose, api\_2-classify\_gender, api\_2-get\_gender\_probabilities, api\_10-detect\_expression
    \item api\_13-segment\_face\_regions, api\_2-classify\_gender, api\_10-detect\_expression, api\_2-get\_gender\_probabilities, api\_11-estimate\_head\_pose
}
\end{enumerate}
\end{tcolorbox}

\begin{tcolorbox}[colback=gray!10, colframe=gray!50, boxrule=0.5mm, rounded corners, title={\textbf{\textcolor{black}{Tools Retrieval \textbar{} Text Only}}}]
\textbf{A high-security facility verifies individuals' identities, estimates age, and monitors expressions. Age estimation is only done if expression confidence is above a threshold. What is the appropriate API function sequence?}
\begin{enumerate}[label=\textbf{\Alph*.}]
{\small
    \item api\_7-identify\_high\_res\_face, api\_1-predict\_age, api\_10-detect\_expression, api\_10-get\_emotion\_probabilities
    \item api\_7-identify\_high\_res\_face, api\_10-get\_emotion\_probabilities, api\_10-detect\_expression, api\_1-predict\_age
    \item api\_7-identify\_high\_res\_face, api\_10-detect\_expression, api\_10-get\_emotion\_probabilities, api\_1-predict\_age
    \item api\_10-detect\_expression, api\_7-identify\_high\_res\_face, api\_1-predict\_age, api\_10-get\_emotion\_probabilities
}
\end{enumerate}
\end{tcolorbox}

\begin{tcolorbox}[colback=gray!10, colframe=gray!50, boxrule=0.5mm, rounded corners, title={\textbf{\textcolor{black}{Tools Retrieval \textbar{} Text Only}}}]
\textbf{For a security checkpoint, the system detects deepfakes, checks for spoofing, and estimates head pose. Spoof detection is mandatory before head pose estimation if deepfake confidence is high. What is the correct sequence?}
\begin{enumerate}[label=\textbf{\Alph*.}]
{\small
    \item api\_11-estimate\_head\_pose, api\_5-detect\_deepfake, api\_4-detect\_spoofing
    \item api\_5-detect\_deepfake, api\_11-estimate\_head\_pose, api\_4-spoof\_confidence\_score
    \item api\_5-detect\_deepfake, api\_4-detect\_spoofing, api\_11-estimate\_head\_pose
    \item api\_4-detect\_spoofing, api\_11-estimate\_head\_pose, api\_5-detect\_deepfake
}
\end{enumerate}
\end{tcolorbox}

\begin{tcolorbox}[colback=gray!10, colframe=gray!50, boxrule=0.5mm, rounded corners, title={\textbf{\textcolor{black}{Tools Retrieval \textbar{} Text Only}}}]
\textbf{An interactive museum exhibit uses head pose tracking and demographic analysis (age, gender, race) to tailor content for visitors. Race prediction is skipped if age confidence falls below a certain threshold. Which API sequence would you use?}
\begin{enumerate}[label=\textbf{\Alph*.}]
{\small
    \item api\_11-estimate\_head\_pose, api\_1-predict\_age, api\_1-age\_confidence\_score, api\_3-predict\_race, api\_2-get\_gender\_probabilities
    \item api\_11-estimate\_head\_pose, api\_1-predict\_age, api\_1-age\_confidence\_score, api\_2-classify\_gender, api\_3-predict\_race
    \item api\_1-predict\_age, api\_11-estimate\_head\_pose, api\_1-age\_confidence\_score, api\_2-classify\_gender, api\_3-get\_race\_probabilities
    \item api\_11-estimate\_head\_pose, api\_1-predict\_age, api\_2-classify\_gender, api\_1-age\_confidence\_score, api\_3-predict\_race
}
\end{enumerate}
\end{tcolorbox}

%% file: appendix/X_results.tex
\section{Results}
In this section, we provide additional results as an extension of the main paper. We also showcase several failure cases and include more details on the baselines to facilitate easier reproducibility.

\subsection{Baselines} 
\label{appendix:baselines}
We detail the sources of the pretrained models we use in the paper for evaluation.
\begin{itemize}
    \item \textbf{GPT-4o}: The API key can be found at~{\small\url{https://platform.openai.com/docs/api-reference/introduction}}.
    \item \textbf{Gemini 1.5 Pro}: The API key can be found at~{\small\url{https://ai.google.dev/gemini-api/docs?gad_source=1&gclid=Cj0KCQiA6Ou5BhCrARIsAPoTxrBNiYImqyaX_VkK97FSSAtzQ2nAfN__Ksk_hJut5SdzpVxrWvVOsBQaAqpVEALw_wcB}}.
    \item \textbf{PaliGemma 3b Mix 448}: The model is available at~{\small\url{https://huggingface.co/google/paligemma-3b-mix-448}}.
    \item \textbf{LLaVA-OneVision Qwen2 0.5b}: The model can be accessed here~{\small\url{https://huggingface.co/llava-hf/llava-onevision-qwen2-0.5b-ov-hf}}.
    \item \textbf{VILA1.5-3b}: The model can be found here~{\small\url{https://huggingface.co/Efficient-Large-Model/VILA1.5-3b}}.
    \item \textbf{Chameleon 7b}: The model is provided by Meta and available at~{\small\url{https://huggingface.co/facebook/chameleon-7b}}.
    \item \textbf{Eagle-X4-8b-Plus}: The model is accessible at~{\small\url{https://hf.rst.im/NVEagle/Eagle-X4-8B-Plus}}.
    \item \textbf{Idefics-9b-Instruct}: The pretrained model is available at~{\small\url{https://huggingface.co/HuggingFaceM4/idefics-9b-instruct}}.
    \item \textbf{LLaVA v1.5 7b}: The model can be found here~{\small\url{https://huggingface.co/liuhaotian/llava-v1.5-7b}}.
    \item \textbf{Monkey-Chat}: We obtained the pretrained model released by its author~{\small\url{https://huggingface.co/echo840/Monkey-Chat}}.
    \item \textbf{MiniCPM-Llama3-V-2.5}: The pretrained model can be found here~{\small\url{https://huggingface.co/openbmb/MiniCPM-Llama3-V-2_5}}.
    \item \textbf{LLaVA-NeXT-Interleave-7b}: The pretrained model is available at~{\small\url{https://huggingface.co/lmms-lab/llava-next-interleave-qwen-7b}}.
    \item \textbf{LLaVA-OneVision Qwen2 7b SI}: The pretrained model is available at~{\small\url{https://huggingface.co/lmms-lab/llava-onevision-qwen2-7b-si}}.
    \item \textbf{Idefics2-8b}: The pretrained model is available at~{\small\url{https://huggingface.co/HuggingFaceM4/idefics2-8b}}.
    \item \textbf{Mantis-SIGLIP-8b}: The pretrained model is available at~{\small\url{https://huggingface.co/TIGER-Lab/Mantis-8B-siglip-llama3}}.
    \item \textbf{Phi-3.5 Vision}: The model is provided by Microsoft and available at~{\small\url{https://huggingface.co/microsoft/Phi-3.5-vision-instruct}}.
    \item \textbf{LLaVA-OneVision Qwen2 7b OV}: The pretrained model is available at~{\small\url{https://huggingface.co/lmms-lab/llava-onevision-qwen2-7b-ov}}.
    \item \textbf{Qwen2-VL-7b-Instruct}: The pretrained model is available at~{\small\url{https://huggingface.co/Qwen/Qwen2-VL-7B-Instruct}}.
    \item \textbf{InternVL2-8b}: The pretrained model is available at~{\small\url{https://huggingface.co/OpenGVLab/InternVL2-8B}}.
    \item \textbf{Idefics-80b-Instruct}: The pretrained model is available at~{\small\url{https://huggingface.co/HuggingFaceM4/idefics-80b-instruct}}.
    \item \textbf{LLaVA v1.5 13b}: We obtained the pretrained model released by its author~{\small\url{https://huggingface.co/liuhaotian/llava-v1.5-13b}}.
    \item \textbf{VILA1.5-13b}: The model is accessible at~{\small\url{https://huggingface.co/Efficient-Large-Model/VILA1.5-13b}}.
    \item \textbf{CogVLM2-19b}: The pretrained model is accessible at~{\small\url{https://huggingface.co/THUDM/cogvlm2-llama3-chat-19B}}.
    \item \textbf{InternVL-Chat-v1.5}: The model is accessible at~{\small\url{https://huggingface.co/OpenGVLab/InternVL-Chat-V1-5}}.
    \item \textbf{VILA1.5-40b}: The pretrained model is provided at~{\small\url{https://huggingface.co/Efficient-Large-Model/VILA1.5-40b}}.
    \item \textbf{LLaVA-OneVision Qwen2 72b OV}: The pretrained model can be accessed here~{\small\url{https://huggingface.co/lmms-lab/llava-onevision-qwen2-72b-ov}}.
    \item \textbf{InternVL2-76b}: The pretrained model is available at~{\small\url{https://huggingface.co/OpenGVLab/InternVL2-Llama3-76B}}.
    \item \textbf{Qwen2-VL-72b-Instruct}: The pretrained model is available at~{\small\url{https://huggingface.co/Qwen/Qwen2-VL-72B-Instruct}}.
\end{itemize}

\begin{table}[htbp]
\centering
\begin{tabular}{lccc}
\toprule
\textbf{Model} & \textbf{Vision Encoder} & \textbf{LLM} & \textbf{LLM Size} \\ \hline
PaliGemma~\cite{beyer2024paligemma} & SigLIP-So400m & Gemma-2B & 2B \\ \hline
LLaVA-OneVision-0.5b-OV~\cite{li2024llava} & SigLIP-So400m & Qwen2-0.5b & 0.5B \\ \hline
VILA 1.5-3b~\cite{lin2024vila} & SigLIP-So400m & Sheared LLaMA-2.7b & 2.7B \\ \hline
Eagle-X4-8B-Plus~\cite{shi2024eagle} & Mixture of Vision Encoder & LLama3-8b-instruct & 8B \\ \hline
Idefics2-8b~\cite{laurençon2024matters} & SigLIP-So400m & Mistral-7B-v0.1 & 7B \\ \hline
Idefics-9b-instruct~\cite{laurencon2023obelics} & CLIP-ViT-H-14-laion2B-s32B-b79K & LLama-7b & 7B \\ \hline
LLaVA-v1.5-7b~\cite{liu2024visual} & CLIP ViT-L-14 & Vicuna-7b & 7B \\ \hline
Monkey-Chat~\cite{li2024monkey} & ViT-BigG-2b & Qwen-7b & 7B \\ \hline
MiniCPM-Llama3-v-2.5~\cite{yao2024minicpmv} & SigLIP-So400m & Llama3-8B-Instruct & 8B \\ \hline
LLaVA-OneVision-7b-SI~\cite{li2024llava} & SigLIP-So400m & Qwen-7b & 7B \\ \hline
LLaVA-NeXT-Interleave-7b~\cite{li2024llava_next} & SigLIP-So400m & Qwen1.5-7b & 7B \\ \hline
Phi-3.5-Vision~\cite{abdin2024phi} & CLIP ViT-L/14 & phi-3.5-mini & 3.8B \\ \hline
LLaVA-OneVision-7b-OV~\cite{li2024llava} & SigLIP-So400m & Llama3-8B-Instruct & 8B \\ \hline
Qwen2-VL-7b-Instruct~\cite{Qwen2VL} & ViT-H & Qwen2 & 7B \\ \hline
InternVL2-8B~\cite{chen2024far} & InternViT-300M-448px & Internlm2\_5-7b-chat & 7B \\ \hline
CogVLM2-19b~\cite{hong2024cogvlm2} & EVA-CLIP & Llama-3-8B-Instruct & 8B \\ \hline
Idefics-80b-instruct~\cite{laurencon2023obelics} & CLIP-ViT-H-14-laion2B-s32B-b79K & LLama-65b & 65B \\ \hline
LLaVA-v1.5-13b~\cite{liu2024visual} & CLIP ViT-L-14 & Vicuna-13b & 13B \\ \hline
VILA 1.5-13b~\cite{lin2024vila} & SigLIP-So400m & Vicuna-13b & 13B \\ \hline
Mantis-SIGLIP-8b~\cite{jiang2024mantis} & SigLIP-So400m & LLama-8b & 8B \\ \hline
InternVL-Chat-v1.5~\cite{chen2024far} & InternViT-6B-448px-V1-5 & InternLM2-Chat-20B & 20B \\ \hline
LLaVA-OneVision-72b-OV~\cite{li2024llava} & SigLIP-So400m & Qwen-72B & 72B \\ \hline
Qwen2-VL-72b-Instruct~\cite{Qwen2VL} & ViT-H & Qwen2 & 72B \\ \hline
InternVL2-76B~\cite{chen2024far} & InternViT-6B-448px-V1-5 & Hermes-2-Theta-Llama-3-70B & 70B \\ \bottomrule
\end{tabular}
\caption{Vision encoder, LLM and LLM size of the baseline models}
\label{tab:model_comparison}
\end{table}

\subsection{Results across different evaluation settings}
\label{sec:appendix_results}
To provide additional insights, we present the results of selected models under the "in-context" and "chain-of-thought" evaluation settings in  Table~\ref{tab:in-context} and Table~\ref{tab:chain-of-thought}, respectively. We observe a consistent drop in performance, indicating that the models are not capable of effectively processing in-context information and fail to reason in tasks related to face understanding.

\begin{table}
    \centering
    \rotatebox{90}{
    \parbox{0.95\textheight}{
        \centering
        \begin{minipage}{0.95\textheight}
            \centering
            \resizebox{0.95\textheight}{!}{
                \begin{tabular}{lccccccccccccccc}
                    \toprule
                    \textbf{Model} & \textbf{Overall} & \textbf{\begin{tabular}[c]{@{}c@{}}Expression \\ Recognition\end{tabular}} & \textbf{\begin{tabular}[c]{@{}c@{}}Age \\ Estimation\end{tabular}} & \textbf{\begin{tabular}[c]{@{}c@{}}Race \\ Estimation\end{tabular}} & \textbf{\begin{tabular}[c]{@{}c@{}}Crowd \\ Counting\end{tabular}} & \textbf{\begin{tabular}[c]{@{}c@{}}Celebrity \\ Identification\end{tabular}} & \textbf{HR FR} & \textbf{LR FR} & \textbf{\begin{tabular}[c]{@{}c@{}}Face \\ Anti-spoofing\end{tabular}} & \textbf{\begin{tabular}[c]{@{}c@{}}Tools \\ Retrieval\end{tabular}} & \textbf{\begin{tabular}[c]{@{}c@{}}Attributes \\ Prediction\end{tabular}} & \textbf{\begin{tabular}[c]{@{}c@{}}Gender \\ Prediction\end{tabular}} & \textbf{\begin{tabular}[c]{@{}c@{}}Headpose \\ Estimation\end{tabular}} & \textbf{\begin{tabular}[c]{@{}c@{}}Face \\ Parsing\end{tabular}} & \textbf{\begin{tabular}[c]{@{}c@{}}Deepfake \\ Detection\end{tabular}} \\
                    \midrule
                    Phi-3.5-Vision & 41.48 & 53.00 & 33.20 & 46.40 & 30.33 & 51.33 & 38.75 & 30.00 & 43.00 & 39.00 & 42.75 & 62.80 & 24.25 & 40.00 & 27.00 \\
                    Qwen2-VL-7B-Instruct & 50.32 & 51.75 & 38.00 & 54.40 & 22.33 & 55.67 & 60.50 & 34.00 & 45.00 & 42.00 & \textbf{\textcolor{blue}{58.50}} & 72.80 & 27.25 & 78.75 & 31.00 \\
                    InternVL2-8B & \textbf{\textcolor{blue}{52.14}} & \textbf{\textcolor{blue}{58.00}} & \textbf{\textcolor{blue}{40.60}} & \textbf{\textcolor{blue}{63.20}} & \textbf{\textcolor{blue}{27.67}} & \textbf{\textcolor{blue}{57.67}} & \textbf{\textcolor{blue}{69.50}} & \textbf{\textcolor{blue}{39.00}} & \textbf{\textcolor{blue}{48.50}} & \textbf{\textcolor{blue}{48.00}} & 48.75 & \textbf{\textcolor{blue}{74.00}} & \textbf{\textcolor{blue}{28.75}} & 66.25 & \textbf{\textcolor{blue}{32.00}} \\
                    \bottomrule
                \end{tabular}
            }
        \caption{(a) Performance of models in in-context evaluation setting.}
        \label{tab:in-context}
    \end{minipage}
    
    \vspace{1cm} 

    \begin{minipage}{0.95\textheight}
        \centering
        \resizebox{0.95\textheight}{!}{
        \begin{tabular}{lccccccccccccccc}
            \toprule
            \textbf{Model} & \textbf{Overall} & \textbf{\begin{tabular}[c]{@{}c@{}}Expression \\ Recognition\end{tabular}} & \textbf{\begin{tabular}[c]{@{}c@{}}Age \\ Estimation\end{tabular}} & \textbf{\begin{tabular}[c]{@{}c@{}}Race \\ Estimation\end{tabular}} & \textbf{\begin{tabular}[c]{@{}c@{}}Crowd \\ Counting\end{tabular}} & \textbf{\begin{tabular}[c]{@{}c@{}}Celebrity \\ Identification\end{tabular}} & \textbf{HR FR} & \textbf{LR FR} & \textbf{\begin{tabular}[c]{@{}c@{}}Face \\ Anti-spoofing\end{tabular}} & \textbf{\begin{tabular}[c]{@{}c@{}}Tools \\ Retrieval\end{tabular}} & \textbf{\begin{tabular}[c]{@{}c@{}}Attributes \\ Prediction\end{tabular}} & \textbf{\begin{tabular}[c]{@{}c@{}}Gender \\ Prediction\end{tabular}} & \textbf{\begin{tabular}[c]{@{}c@{}}Headpose \\ Estimation\end{tabular}} & \textbf{\begin{tabular}[c]{@{}c@{}}Face \\ Parsing\end{tabular}} & \textbf{\begin{tabular}[c]{@{}c@{}}Deepfake \\ Detection\end{tabular}} \\
            \midrule
            Phi-3.5-Vision & 33.36 & 45.00 & 26.60 & 32.40 & 30.67 & 24.00 & 35.50 & 28.00 & \textbf{\textcolor{blue}{42.25}} & 28.00 & 30.00 & 32.40 & 25.50 & 43.25 & 26.00 \\
            Qwen2-VL-7B-Instruct & 39.62 & 50.00 & 34.60 & 48.40 & 31.00 & 49.00 & 37.00 & 34.00 & 29.50 & 38.00 & 40.75 & 48.20 & 21.75 & \textbf{\textcolor{blue}{56.25}} & 24.00 \\
            InternVL2-8B & \textbf{\textcolor{blue}{48.88}} & \textbf{\textcolor{blue}{59.00}} & \textbf{\textcolor{blue}{41.00}} & \textbf{\textcolor{blue}{56.00}} & \textbf{\textcolor{blue}{37.00}} & \textbf{\textcolor{blue}{55.00}} & \textbf{\textcolor{blue}{64.00}} & \textbf{\textcolor{blue}{43.00}} & 35.00 & \textbf{\textcolor{blue}{45.00}} & \textbf{\textcolor{blue}{40.75}} & \textbf{\textcolor{blue}{81.20}} & \textbf{\textcolor{blue}{31.50}} & 48.50 & \textbf{\textcolor{blue}{24.67}} \\
            \bottomrule
        \end{tabular}
        }
        \caption{(b) Performance of models in chain-of-thought evaluation setting.}
        \label{tab:chain-of-thought}
    \end{minipage}
    }
    }
\end{table}

\subsection{Human and SOTA Vision models evaluation details}
For human evaluation, we conducted assessments with five graduate student participants and reported the average results. For evaluating Vision SOTA models, certain tasks, such as tool retrieval and celebrity identification, do not have a specific benchmark. In these cases, we considered Gemini Pro 1.5~\cite{team5gemini} as the SOTA model. For tasks such as expression recognition, age, gender, and race estimation, attribute prediction, and head pose estimation, we use FaceXFormer~\cite{narayan2024facexformer} as the SOTA model. For high-resolution face recognition, we employ ArcFace~\cite{deng2019arcface}, while for low-resolution face recognition, we use PETALface~\cite{narayan2024petalface}. Additionally, we utilize SegFace~\cite{narayan2024segface} for face parsing, FLIP~\cite{le2024gradient} for face anti-spoofing, and XceptionNet~\cite{chollet2017xception} for deepfake detection.

As can be seen from Table~\ref{tab:results_tasks}, there are several tasks where human participants perform significantly worse than SOTA models, primarily due to the nature of the tasks. These include deepfake detection, head pose estimation, celebrity identification, crowd counting, and age estimation. In deepfake detection, some manipulations are more pronounced in the spectral domain but remain imperceptible to the human eye. Similarly, age estimation requires high precision, as the distractor options are designed to be very close in value, meaning even a small human error results in an incorrect choice, making the benchmark more challenging and forcing the model to be precise. For head pose estimation, humans struggle to visualize fine angular changes in yaw, pitch, and roll, leading to poor performance. Additionally, the low human performance in celebrity identification is primarily due to the participants' limited familiarity with a wide range of celebrities.

\subsection{Failure Cases}
We showed one example in the main paper. We have included examples for the other tasks below. Our observations reveal that these models struggle to answer basic questions related to face understanding, highlighting the need for a dedicated benchmark to monitor the progress of MLLMs in this domain.

\clearpage

\input{appendix/X_failure_cases}

%% file: appendix/X_failure_cases.tex
\begin{tcolorbox}[colback=white, colframe=gray!50, boxrule=0.5mm, rounded corners, title={\textbf{\textcolor{black}{Age Estimation}}}]
\begin{tcolorbox}[colback=orange!10, boxrule=0mm]
    \textbf{Question:} \texttt{<image 1><image 2>} Identify the person who is most likely in 20-29 age group.

    \textbf{Option:}
    \begin{enumerate}[label=(\Alph*)]
        \item None of the Above
        \item Image 1
        \item Image 2
        \item Both Images
    \end{enumerate}
\end{tcolorbox}
    
\begin{tabular}{c@{\hspace{2em}}c}
    \includegraphics[width=0.2\textwidth]{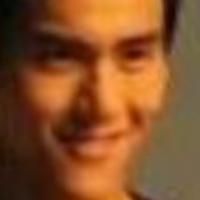} &
    \includegraphics[width=0.2\textwidth]{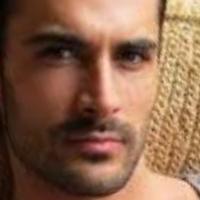} \\
    \texttt{<image 1>} & \texttt{<image 2>}
\end{tabular}

\begin{tcolorbox}[colback=blue!10, boxrule=0mm]
    \textbf{GPT-4o:} (A) None of the above \\
    \textbf{Prediction:} \ding{55}
\end{tcolorbox}

\begin{tcolorbox}[colback=purple!10, boxrule=0mm]
    \textbf{Qwen2-VL:} C \\
    \textbf{Prediction:} \ding{55} 
\end{tcolorbox}

\begin{tcolorbox}[colback=yellow!20, boxrule=0mm]
    \textbf{InternVL2:} (B) Image 1 \\
    \textbf{Prediction:} \ding{55} 
\end{tcolorbox}

\begin{tcolorbox}[colback=green!10, boxrule=0mm]
    \textbf{Ground Truth:} (\textbf{D}) \textbf{Both Images}
\end{tcolorbox}

\end{tcolorbox}

\begin{tcolorbox}[colback=white, colframe=gray!50, boxrule=0.5mm, rounded corners, title={\textbf{\textcolor{black}{Gender Prediction}}}]
\begin{tcolorbox}[colback=orange!10, boxrule=0mm]
    \textbf{Question:} \texttt{<image 1><image 2><image 3>} Which images depict female individuals?

    \textbf{Option:}
    \begin{enumerate}[label=(\Alph*)]
        \item Image 3
        \item Image 1, Image 2
        \item Image 3, Image 1
        \item Image 1
    \end{enumerate}
\end{tcolorbox}
    
\begin{tabular}{c@{\hspace{2em}}c@{\hspace{2em}}c}
    \includegraphics[width=0.2\textwidth]{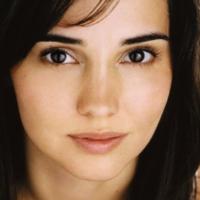} &
    \includegraphics[width=0.2\textwidth]{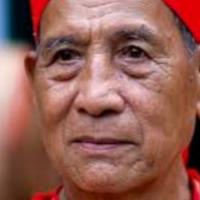} &
    \includegraphics[width=0.2\textwidth]{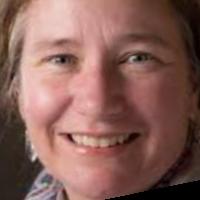} \\
    \texttt{<image 1>} & \texttt{<image 2>} & \texttt{<image 3>}
\end{tabular}

\begin{tcolorbox}[colback=blue!10, boxrule=0mm]
    \textbf{GPT-4o:} (D) Image 1 \\
    \textbf{Prediction:} \ding{55}
\end{tcolorbox}

\begin{tcolorbox}[colback=purple!10, boxrule=0mm]
    \textbf{Qwen2-VL:} C \\
    \textbf{Prediction:} \ding{51} 
\end{tcolorbox}

\begin{tcolorbox}[colback=yellow!20, boxrule=0mm]
    \textbf{InternVL2:} (C) Image 3, Image 1 \\
    \textbf{Prediction:} \ding{51} 
\end{tcolorbox}

\begin{tcolorbox}[colback=green!10, boxrule=0mm]
    \textbf{Ground Truth:} (\textbf{C}) \textbf{Image 3, Image 1}
\end{tcolorbox}

\end{tcolorbox}

\begin{tcolorbox}[colback=white, colframe=gray!50, boxrule=0.5mm, rounded corners, title={\textbf{\textcolor{black}{Race Estimation}}}]
\begin{tcolorbox}[colback=orange!10, boxrule=0mm]
    \textbf{Question:} \texttt{<image 1><image 2><image 3>} Which images depict people of the same race?

    \textbf{Option:}
    \begin{enumerate}[label=(\Alph*)]
        \item Image 2, Image 3
        \item Image 1, Image 2
        \item None of the above
        \item Image 1, Image 2, Image 3
    \end{enumerate}
\end{tcolorbox}
    
\begin{tabular}{c@{\hspace{2em}}c@{\hspace{2em}}c}
    \includegraphics[width=0.2\textwidth]{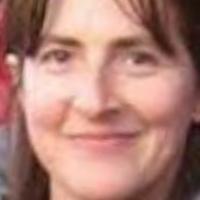} &
    \includegraphics[width=0.2\textwidth]{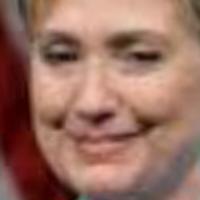} &
    \includegraphics[width=0.2\textwidth]{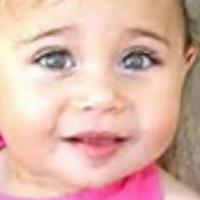} \\
    \texttt{<image 1>} & \texttt{<image 2>} & \texttt{<image 3>}
\end{tabular}

\begin{tcolorbox}[colback=blue!10, boxrule=0mm]
    \textbf{GPT-4o:} C) None of the above \\
    \textbf{Prediction:} \ding{51}
\end{tcolorbox}

\begin{tcolorbox}[colback=purple!10, boxrule=0mm]
    \textbf{Qwen2-VL:} B \\
    \textbf{Prediction:} \ding{55}
\end{tcolorbox}

\begin{tcolorbox}[colback=yellow!20, boxrule=0mm]
    \textbf{InternVL2:} (B) Image 1, Image 2 \\
    \textbf{Prediction:} \ding{55}
\end{tcolorbox}

\begin{tcolorbox}[colback=green!10, boxrule=0mm]
    \textbf{Ground Truth:} (\textbf{C}) \textbf{None of the above}
\end{tcolorbox}

\end{tcolorbox}

\begin{tcolorbox}[colback=white, colframe=gray!50, boxrule=0.5mm, rounded corners, title={\textbf{\textcolor{black}{High-Resolution Face Recognition}}}]
\begin{tcolorbox}[colback=orange!10, boxrule=0mm]
    \textbf{Question:} \texttt{<image 1><image 2><image 3><image 4>} How many unique identities are present in these images?

    \textbf{Option:}
    \begin{enumerate}[label=(\Alph*)]
        \item 3
        \item 2
        \item 4
        \item 1
    \end{enumerate}
\end{tcolorbox}
    
\begin{tabular}{c@{\hspace{2em}}c@{\hspace{2em}}c@{\hspace{2em}}c}
    \includegraphics[width=0.2\textwidth]{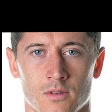} &
    \includegraphics[width=0.2\textwidth]{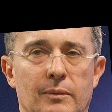} &
    \includegraphics[width=0.2\textwidth]{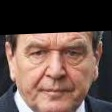} & 
    \includegraphics[width=0.2\textwidth]{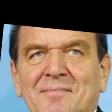} \\
    \texttt{<image 1>} & \texttt{<image 2>} & \texttt{<image 3>} & \texttt{<image 4>}
\end{tabular}

\begin{tcolorbox}[colback=blue!10, boxrule=0mm]
    \textbf{GPT-4o:} B \\
    \textbf{Prediction:} \ding{55}
\end{tcolorbox}

\begin{tcolorbox}[colback=purple!10, boxrule=0mm]
    \textbf{Qwen2-VL:} C \\
    \textbf{Prediction:} \ding{55}
\end{tcolorbox}

\begin{tcolorbox}[colback=yellow!20, boxrule=0mm]
    \textbf{InternVL2:} (C) 4 \\
    \textbf{Prediction:} \ding{55}
\end{tcolorbox}

\begin{tcolorbox}[colback=green!10, boxrule=0mm]
    \textbf{Ground Truth:} (\textbf{A}) \textbf{3}
\end{tcolorbox}

\end{tcolorbox}

\begin{tcolorbox}[colback=white, colframe=gray!50, boxrule=0.5mm, rounded corners, title={\textbf{\textcolor{black}{Low-Resolution Face Recognition}}}]
\begin{tcolorbox}[colback=orange!10, boxrule=0mm]
    \textbf{Question:} \texttt{<image 1><image 2><image 3><image 4><image 5>} The first image is of person A. The same person A is present in which of these images?

    \textbf{Option:}
    \begin{enumerate}[label=(\Alph*)]
        \item Image 3 and Image 5
        \item Image 2 and Image 3 and Image 4
        \item Image 4 and Image 5
        \item Image 2 and Image 5
    \end{enumerate}
\end{tcolorbox}
    
\begin{tabular}{c@{\hspace{2em}}c@{\hspace{2em}}c@{\hspace{2em}}c@{\hspace{2em}}c}
    \includegraphics[width=60px,height=60px]{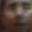} &
    \includegraphics[width=60px,height=60px]{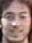} &
    \includegraphics[width=60px,height=60px]{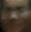} & 
    \includegraphics[width=60px,height=60px]{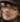} &
    \includegraphics[width=60px,height=60px]{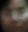} \\
    \texttt{<image 1>} & \texttt{<image 2>} & \texttt{<image 3>} & \texttt{<image 4>} & \texttt{<image 5>}
\end{tabular}

\begin{tcolorbox}[colback=blue!10, boxrule=0mm]
    \textbf{GPT-4o:} I can't determine if the same person is in multiple images. \\
    \textbf{Prediction:} \ding{55}
\end{tcolorbox}

\begin{tcolorbox}[colback=purple!10, boxrule=0mm]
    \textbf{Qwen2-VL:} D \\
    \textbf{Prediction:} \ding{55}
\end{tcolorbox}

\begin{tcolorbox}[colback=yellow!20, boxrule=0mm]
    \textbf{InternVL2:} (D) Image 2 and Image 5 \\
    \textbf{Prediction:} \ding{55}
\end{tcolorbox}

\begin{tcolorbox}[colback=green!10, boxrule=0mm]
    \textbf{Ground Truth:} (\textbf{A}) \textbf{Image 3 and Image 5}
\end{tcolorbox}

\end{tcolorbox}

\begin{tcolorbox}[colback=white, colframe=gray!50, boxrule=0.5mm, rounded corners, title={\textbf{\textcolor{black}{Celebrity Identification}}}]
\begin{tcolorbox}[colback=orange!10, boxrule=0mm]
    \textbf{Question:} \texttt{<image 1>} What is the name of this celebrity?

    \textbf{Option:}
    \begin{enumerate}[label=(\Alph*)]
        \item Alison Brie
        \item Callan Mulvey
        \item Paul Soter
        \item Ashley Eckstein
    \end{enumerate}
\end{tcolorbox}
    
\begin{tabular}{c}
    \includegraphics[width=0.2\textwidth]{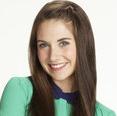} \\
    \texttt{<image 1>}
\end{tabular}

\begin{tcolorbox}[colback=blue!10, boxrule=0mm]
    \textbf{GPT-4o:} I don't know who this person is. \\
    \textbf{Prediction:} \ding{55}
\end{tcolorbox}

\begin{tcolorbox}[colback=purple!10, boxrule=0mm]
    \textbf{Qwen2-VL:} D \\
    \textbf{Prediction:} \ding{55}
\end{tcolorbox}

\begin{tcolorbox}[colback=yellow!20, boxrule=0mm]
    \textbf{InternVL2:} (D) Ashley Eckstein \\
    \textbf{Prediction:} \ding{55}
\end{tcolorbox}

\begin{tcolorbox}[colback=green!10, boxrule=0mm]
    \textbf{Ground Truth:} (\textbf{A}) \textbf{Alison Brie}
\end{tcolorbox}

\end{tcolorbox}

\begin{tcolorbox}[colback=white, colframe=gray!50, boxrule=0.5mm, rounded corners, title={\textbf{\textcolor{black}{Deepfake Detection}}}]
\begin{tcolorbox}[colback=orange!10, boxrule=0mm]
    \textbf{Question:} \texttt{<image 1><image 2><image 3>} How many images are real?

    \textbf{Option:}
    \begin{enumerate}[label=(\Alph*)]
        \item 2
        \item 1
        \item 3
        \item 0
    \end{enumerate}
\end{tcolorbox}
    
\begin{tabular}{c@{\hspace{2em}}c@{\hspace{2em}}c}
    \includegraphics[width=110px, height=70px]{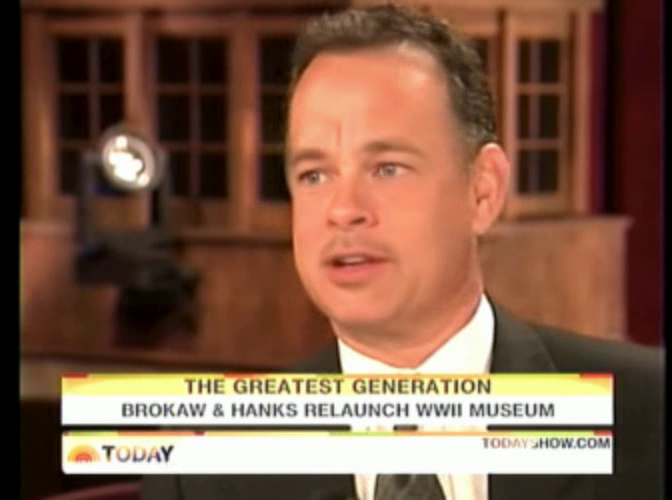} &
    \includegraphics[width=110px, height=70px]{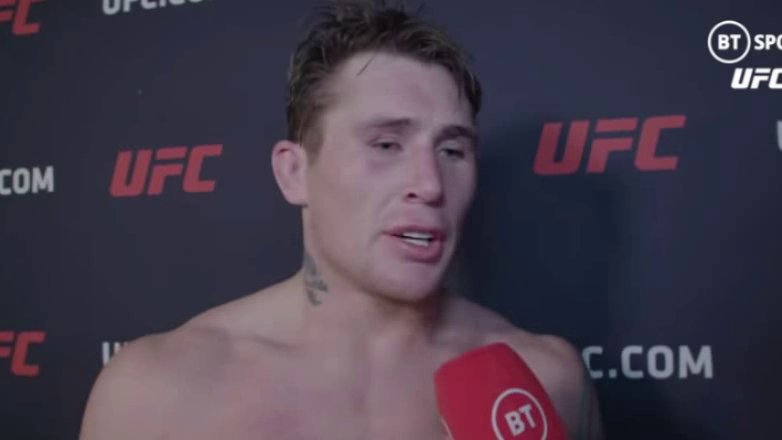} &
    \includegraphics[width=110px, height=70px]{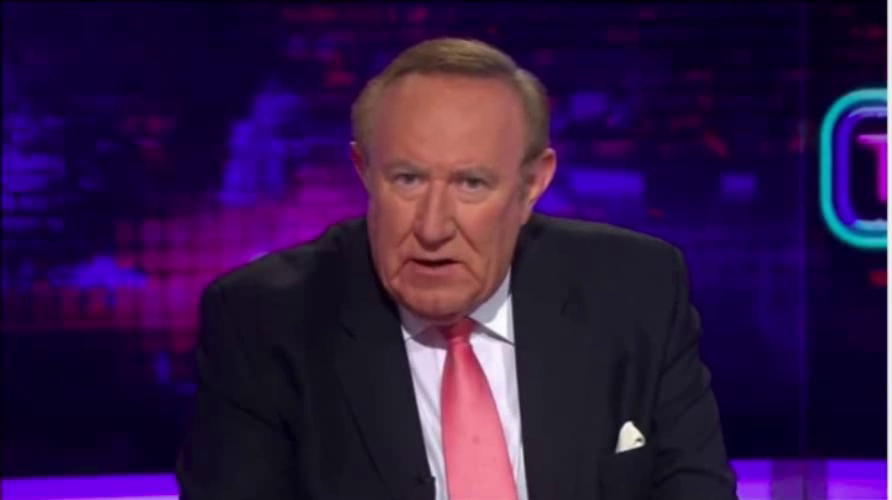} \\
    \texttt{<image 1>} & \texttt{<image 2>} & \texttt{<image 3>}
\end{tabular}

\begin{tcolorbox}[colback=blue!10, boxrule=0mm]
    \textbf{GPT-4o:} (A) 2 \\
    \textbf{Prediction:} \ding{51}
\end{tcolorbox}

\begin{tcolorbox}[colback=purple!10, boxrule=0mm]
    \textbf{Qwen2-VL:} C \\
    \textbf{Prediction:} \ding{55}
\end{tcolorbox}

\begin{tcolorbox}[colback=yellow!20, boxrule=0mm]
    \textbf{InternVL2:} (C) 3 \\
    \textbf{Prediction:} \ding{55} 
\end{tcolorbox}

\begin{tcolorbox}[colback=green!10, boxrule=0mm]
    \textbf{Ground Truth:} (\textbf{A}) \textbf{2}
\end{tcolorbox}

\end{tcolorbox}

\begin{tcolorbox}[colback=white, colframe=gray!50, boxrule=0.5mm, rounded corners, title={\textbf{\textcolor{black}{Attributes Prediction}}}]
\begin{tcolorbox}[colback=orange!10, boxrule=0mm]
    \textbf{Question:} \texttt{<image 1><image 2><image 3>} Which images have a person with the attribute 'Pointy Nose'?

    \textbf{Option:}
    \begin{enumerate}[label=(\Alph*)]
        \item Image 2, Image 3
        \item Image 1, Image 3
        \item Image 1
        \item Image 1, Image 2
    \end{enumerate}
\end{tcolorbox}
    
\begin{tabular}{c@{\hspace{2em}}c@{\hspace{2em}}c}
    \includegraphics[width=0.2\textwidth]{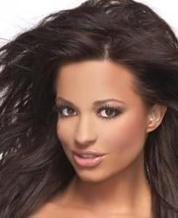} &
    \includegraphics[width=0.2\textwidth]{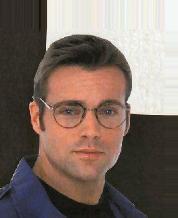} &
    \includegraphics[width=0.2\textwidth]{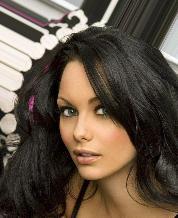} \\
    \texttt{<image 1>} & \texttt{<image 2>} & \texttt{<image 3>}
\end{tabular}

\begin{tcolorbox}[colback=blue!10, boxrule=0mm]
    \textbf{GPT-4o:} (B) Image 1, Image 3 \\
    \textbf{Prediction:} \ding{55}
\end{tcolorbox}

\begin{tcolorbox}[colback=purple!10, boxrule=0mm]
    \textbf{Qwen2-VL:} B \\
    \textbf{Prediction:} \ding{55} 
\end{tcolorbox}

\begin{tcolorbox}[colback=yellow!20, boxrule=0mm]
    \textbf{InternVL2:} (C) Image 1 \\
    \textbf{Prediction:} \ding{51}
\end{tcolorbox}

\begin{tcolorbox}[colback=green!10, boxrule=0mm]
    \textbf{Ground Truth:} (\textbf{C}) \textbf{Image 1}
\end{tcolorbox}

\end{tcolorbox}

\begin{tcolorbox}[colback=white, colframe=gray!50, boxrule=0.5mm, rounded corners, title={\textbf{\textcolor{black}{Facial Expression Recognition}}}]
\begin{tcolorbox}[colback=orange!10, boxrule=0mm]
    \textbf{Question:} \texttt{<image 1><image 2>} Which image has a person with the expression 'sad'?

    \textbf{Option:}
    \begin{enumerate}[label=(\Alph*)]
        \item Neither
        \item Image 1
        \item Both
        \item Image 2
    \end{enumerate}
\end{tcolorbox}
    
\begin{tabular}{c@{\hspace{2em}}c@{\hspace{2em}}c}
    \includegraphics[width=0.2\textwidth]{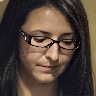} &
    \includegraphics[width=0.2\textwidth]{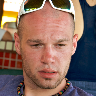} \\
    \texttt{<image 1>} & \texttt{<image 2>}
\end{tabular}

\begin{tcolorbox}[colback=blue!10, boxrule=0mm]
    \textbf{GPT-4o:} A \\
    \textbf{Prediction:} \ding{55}
\end{tcolorbox}

\begin{tcolorbox}[colback=purple!10, boxrule=0mm]
    \textbf{Qwen2-VL:} A \\
    \textbf{Prediction:} \ding{55} 
\end{tcolorbox}

\begin{tcolorbox}[colback=yellow!20, boxrule=0mm]
    \textbf{InternVL2:} (D) Image 2 \\
    \textbf{Prediction:} \ding{51} 
\end{tcolorbox}

\begin{tcolorbox}[colback=green!10, boxrule=0mm]
    \textbf{Ground Truth:} (\textbf{D}) \textbf{Image 2}
\end{tcolorbox}

\end{tcolorbox}

\begin{tcolorbox}[colback=white, colframe=gray!50, boxrule=0.5mm, rounded corners, title={\textbf{\textcolor{black}{Headpose Estimation}}}]
\begin{tcolorbox}[colback=orange!10, boxrule=0mm]
    \textbf{Question:} \texttt{<image 1><image 2><image 3><image 4>} How many images have a person with the yaw angle of headpose orientation in range 0 to 10?

    \textbf{Option:}
    \begin{enumerate}[label=(\Alph*)]
        \item 3
        \item 1
        \item 0
        \item 2
    \end{enumerate}
\end{tcolorbox}
    
\begin{tabular}{c@{\hspace{2em}}c@{\hspace{2em}}c@{\hspace{2em}}c}
    \includegraphics[width=0.2\textwidth]{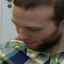} &
    \includegraphics[width=0.2\textwidth]{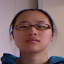} &
    \includegraphics[width=0.2\textwidth]{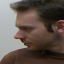} &
    \includegraphics[width=0.2\textwidth]{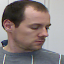} \\
    \texttt{<image 1>} & \texttt{<image 2>} & \texttt{<image 3>} & \texttt{<image 4>}
\end{tabular}

\begin{tcolorbox}[colback=blue!10, boxrule=0mm]
    \textbf{GPT-4o:} D) 2 \\
    \textbf{Prediction:} \ding{55}
\end{tcolorbox}

\begin{tcolorbox}[colback=purple!10, boxrule=0mm]
    \textbf{Qwen2-VL:} B \\
    \textbf{Prediction:} \ding{51} 
\end{tcolorbox}

\begin{tcolorbox}[colback=yellow!20, boxrule=0mm]
    \textbf{InternVL2:} (B) 1 \\
    \textbf{Prediction:} \ding{51} 
\end{tcolorbox}

\begin{tcolorbox}[colback=green!10, boxrule=0mm]
    \textbf{Ground Truth:} (\textbf{B}) \textbf{1}
\end{tcolorbox}

\end{tcolorbox}

\begin{tcolorbox}[colback=white, colframe=gray!50, boxrule=0.5mm, rounded corners, title={\textbf{\textcolor{black}{Face Parsing}}}]
\begin{tcolorbox}[colback=orange!10, boxrule=0mm]
    \textbf{Question:} \texttt{<image 1>} Which of the following regions is not present or is segmented out with white color?

    \textbf{Option:}
    \begin{enumerate}[label=(\Alph*)]
        \item nose
        \item left eye
        \item hair
        \item face
    \end{enumerate}
\end{tcolorbox}
    
\begin{tabular}{c}
    \includegraphics[width=0.2\textwidth]{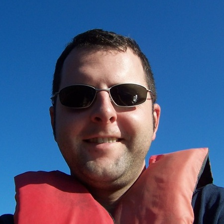} \\
    \texttt{<image 1>}
\end{tabular}

\begin{tcolorbox}[colback=blue!10, boxrule=0mm]
    \textbf{GPT-4o:} A \\
    \textbf{Prediction:} \ding{55}
\end{tcolorbox}

\begin{tcolorbox}[colback=purple!10, boxrule=0mm]
    \textbf{Qwen2-VL:} B \\
    \textbf{Prediction:} \ding{51} 
\end{tcolorbox}

\begin{tcolorbox}[colback=yellow!20, boxrule=0mm]
    \textbf{InternVL2:} (C) hair \\
    \textbf{Prediction:} \ding{55} 
\end{tcolorbox}

\begin{tcolorbox}[colback=green!10, boxrule=0mm]
    \textbf{Ground Truth:} (\textbf{B}) \textbf{left eye}
\end{tcolorbox}

\end{tcolorbox}

\begin{tcolorbox}[colback=white, colframe=gray!50, boxrule=0.5mm, rounded corners, title={\textbf{\textcolor{black}{Crowd Counting}}}]
\begin{tcolorbox}[colback=orange!10, boxrule=0mm]
    \textbf{Question:} \texttt{<image 1>} Estimate the count of people present in this image.

    \textbf{Option:}
    \begin{enumerate}[label=(\Alph*)]
        \item 20 to 24
        \item 25 to 29
        \item 35 to 39
        \item 30 to 34
    \end{enumerate}
\end{tcolorbox}
    
\begin{tabular}{c}
    \includegraphics[width=0.4\textwidth]{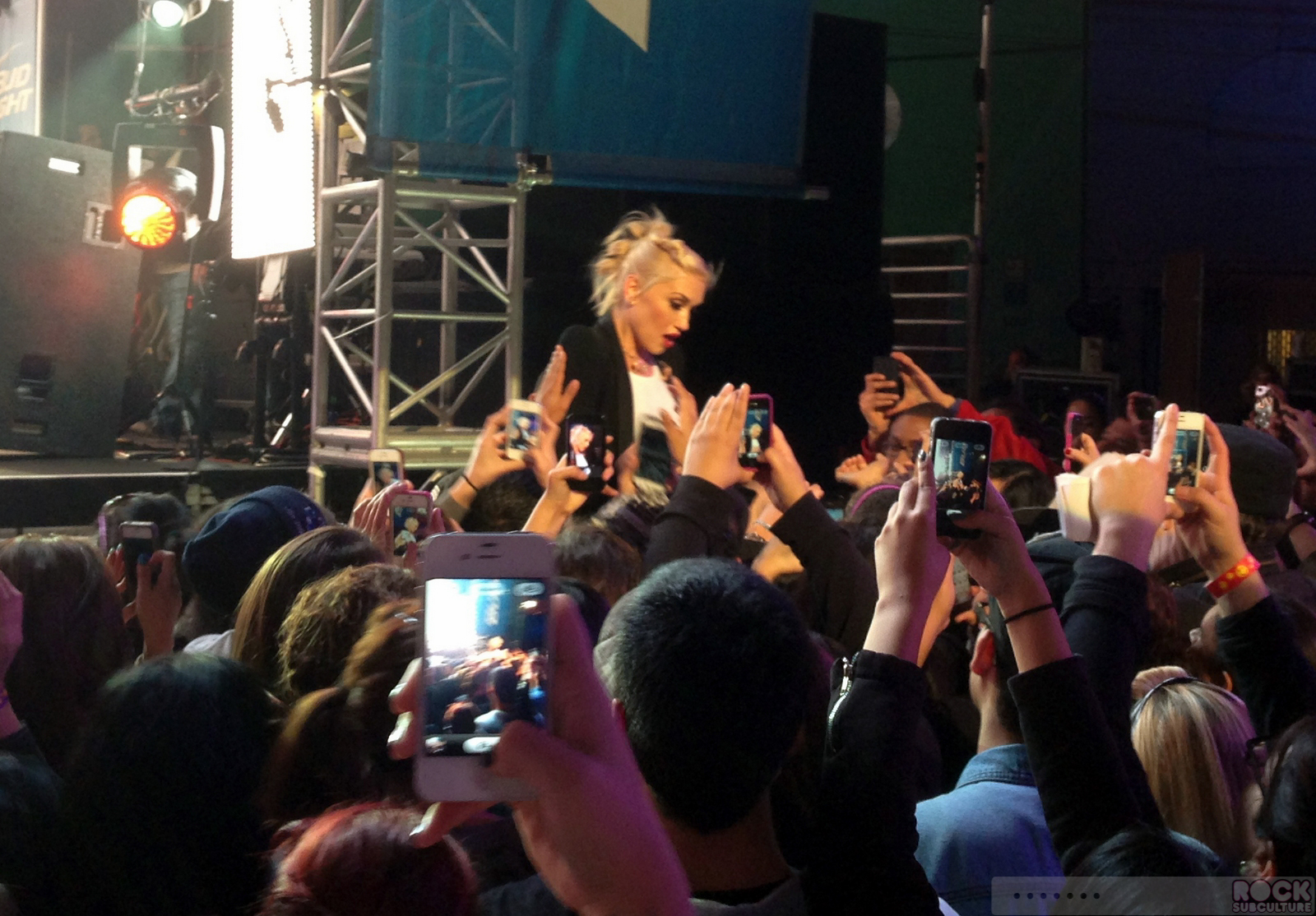} \\
    \texttt{<image 1>}
\end{tabular}

\begin{tcolorbox}[colback=blue!10, boxrule=0mm]
    \textbf{GPT-4o:} C) 35 to 39 \\
    \textbf{Prediction:} \ding{55}
\end{tcolorbox}

\begin{tcolorbox}[colback=purple!10, boxrule=0mm]
    \textbf{Qwen2-VL:} D \\
    \textbf{Prediction:} \ding{55} 
\end{tcolorbox}

\begin{tcolorbox}[colback=yellow!20, boxrule=0mm]
    \textbf{InternVL2:} (B) 25 to 29 \\
    \textbf{Prediction:} \ding{51} 
\end{tcolorbox}

\begin{tcolorbox}[colback=green!10, boxrule=0mm]
    \textbf{Ground Truth:} (\textbf{B}) \textbf{25 to 29}
\end{tcolorbox}

\end{tcolorbox}

\begin{tcolorbox}[colback=white, colframe=gray!50, boxrule=0.5mm, rounded corners, title={\textbf{\textcolor{black}{Tools Retrieval}}}]
\begin{tcolorbox}[colback=orange!10, boxrule=0mm]
    \textbf{Question:} At a large gathering, the system identifies celebrities and tracks their expressions. Expressions are ignored if a spoof attempt is detected. Which sequence of API function calls is suitable?

    \textbf{Option:}
    \begin{enumerate}[label=(\Alph*)]
        \item api\_8-identify\_celebrity, api\_10-track\_expression\_over\_time, api\_4-detect\_spoofing
        \item api\_8-identify\_celebrity, api\_10-detect\_spoofing, api\_4-track\_expression\_over\_time
        \item api\_8-identify\_celebrity, api\_4-spoof\_confidence\_score, api\_10-track\_expression\_over\_time
        \item api\_8-identify\_celebrity, api\_4-spoof\_confidence\_score, api\_10-detect\_expression
    \end{enumerate}
\end{tcolorbox}

\begin{tcolorbox}[colback=blue!10, boxrule=0mm]
    \textbf{GPT-4o:} A \\
    \textbf{Prediction:} \ding{55}
\end{tcolorbox}

\begin{tcolorbox}[colback=purple!10, boxrule=0mm]
    \textbf{Qwen2-VL:} A \\
    \textbf{Prediction:} \ding{55} 
\end{tcolorbox}

\begin{tcolorbox}[colback=yellow!20, boxrule=0mm]
    \textbf{InternVL2:} A \\
    \textbf{Prediction:} \ding{55} 
\end{tcolorbox}

\begin{tcolorbox}[colback=green!10, boxrule=0mm]
    \textbf{Ground Truth:} (\textbf{B}) \textbf{api\_8-identify\_celebrity, api\_10-detect\_spoofing, api\_4-track\_expression\_over\_time}
\end{tcolorbox}
\end{tcolorbox}

%% file: appendix/X_ethics.tex
\section{Ethical Considerations}
In this work, we have ensured that all datasets were obtained exclusively from their official repositories to maintain their integrity, authenticity, and alignment with the original intent of the data creators. This practice mitigates risks associated with data tampering or unauthorized use. Wherever necessary, we have reviewed, acknowledged, and signed the appropriate license agreements to fully comply with the terms and conditions specified by the data and model providers. By doing so, we aim to respect intellectual property rights and uphold the highest ethical standards. Additionally, to promote transparency and reproducibility, we provide source links to all open-source models utilized in our study.